\documentstyle[jair,twoside,11pt,theapa,graphicx,amssymb]{article}

\input epsf %% at top of file

\jairheading{24}{2005}{519-579}{12/04}{10/05}

\ShortHeadings{The Deterministic Part of IPC-4: An Overview} {Hoffmann \& Edelkamp}
\firstpageno{519}

\def\IN{\hbox{{I\kern-.14em N}}}
\def\IR{\hbox{{I\kern-.14em R}}}

\def\oom{$^{\tt +}$}
\def\zom{$^*$}
\def\bump{\hspace{1cm}}
\def\req#1{$^{\tt #1}$}
\newenvironment{tabtt}{\begin{tt} \begin{small}
\begin{tabbing}}{\end{tabbing} \end{small} \end{tt}}
\def\noteme#1{}%{[[#1]]}
\def\notecoauth#1{\ }

\title{The Deterministic Part of IPC-4: An Overview}

\author{\name J\"org Hoffmann
  \email hoffmann@mpi-sb.mpg.de\\
  \addr Max-Planck-Institut f\"ur Informatik,\\
  Saarbr\"ucken, Germany\\
\name Stefan Edelkamp
  \email stefan.edelkamp@cs.uni-dortmund.de\\
  \addr Fachbereich Informatik,\\
  Universit\"at Dortmund, Germany}

\begin{document}
\maketitle

\begin{abstract}
  We provide an overview of the organization and results of the
  deterministic part of the 4th International Planning Competition,
  i.e., of the part concerned with evaluating systems doing
  deterministic planning. IPC-4 attracted even more competing systems
  than its already large predecessors, and the competition event was
  revised in several important respects. After giving an introduction
  to the IPC, we briefly explain the main differences between the
  deterministic part of IPC-4 and its predecessors. We then introduce
  formally the language used, called PDDL2.2 that extends PDDL2.1 by
  \emph{derived predicates} and \emph{timed initial literals}. We list
  the competing systems and overview the results of the competition.
  The entire set of data is far too large to be presented in full.  We
  provide a detailed summary; the complete data is available in an
  online appendix. We explain how we awarded the competition prizes.
\end{abstract}

%% \vfill
%% \pagebreak

%% %\tableofcontents
%% \vfill
%% \pagebreak

\section{Introduction}
\label{introduction}

In the application of Artificial Intelligence technology to the
real-world, time and space resources are usually limited. This has led
to a performance-oriented interpretation of AI in many of its research
branches. Competition events have been established in automated
theorem proving, in satisfiability testing, and, in particular, in AI
Planning. A competition provides a large-scale evaluation platform.
Due to the broadness and neutrality of that platform, a competition is
far better at assessing the state-of-the-art in a research branch than
the experiments ran by individual authors: more systems are compared,
and the benchmarks are chosen by the competition organizers, rather
than by the system authors themselves. Moreover, a competition can
serve to establish a common representation formalism, and a common
core set of benchmarks, marking the edge of current system
capabilities.

The International Planning Competition ({IPC}) is a biennial
event, hosted at the international conferences on AI Planning and
Scheduling. The IPC began in 1998 when Drew McDermott and a committee
created a common specification language (PDDL) and a collection of
problems forming a first benchmark~\cite{pddl-handbook}.  PDDL is a
Lisp-like input language description format that includes planning
formalisms like STRIPS~\cite{fikes:nilsson:ai-71}.  Five systems
participated in the first international planning competition, IPC-1
for short, hosted at AIPS~1998 in Pittsburgh,
Pennsylvania~\cite{mcdermott:aim-00}.

In the year 2000, Fahiem Bacchus continued this work, and the IPC-2
event attracted 16 competitors~\cite{bacchus:aim-01}. The event --
hosted at AIPS~2000 in Breckenridge, Colorado -- was extended to
include both fully automatic and hand-tailored planning systems. As
hand-tailored planners were allowed to use some additional
domain-dependent information to the PDDL input in order to improve
their performance, they participated in an additional, separate,
track. Both STRIPS and ADL~\cite{pednault:kr-89} domains were used but
no further extensions were made to the
language~\cite{pddl-2000-subset}.

The 3rd International Planning Competition, IPC-3, was run by Derek
Long and Maria Fox and was hosted at AIPS~2002, Toulouse, France.  The
competition attracted 14 competitors \cite{long:fox:jair-03}, and
focussed on planning in temporal and metric domains. For that purpose,
Fox and Long developed the PDDL2.1 language~\cite{fox:long:jair-03},
of which the first three {\em levels} were used in IPC-3. Level 1 was
STRIPS and ADL planning as before, Level 2 added numeric variables,
Level 3 added durational constructs.

\begin{sloppypar}
  The 4th International Planning Competition, IPC-4, was hosted at
  ICAPS-2004, Whistler, Canada. IPC-4 built on the previous efforts,
  in particular the language PDDL2.1. The competition event was
  extended and revised in several respects. In particular, IPC-4
  featured, for the first time, a competition for probabilistic
  planners, so that the overall competition was split into a {\em
    deterministic} part -- a continuation of the previous events -- as
  well as a {\em probabilistic} part.\footnote{While IPC-4 was
    running, the deterministic part was named ``classical'' part. We
    re-named it into ``deterministic'' part since that wording is less
    ambiguous.} In the latter part, co-organized by Michael Littman
  and H{\aa}kan Younes, the main objective of the event was to
  introduce a common representation language for probabilistic
  planners, and to establish some first benchmarks and results. For
  more information on the probabilistic part of IPC-4 see the work of
  \citeA{ProbabilisticPart}.
\end{sloppypar}

Herein, we provide an overview of the organization and results of the
deterministic part of IPC-4. With 19 competing systems (21 when
counting different system versions), the event was even a little
larger than its already large predecessors.  Several important
revisions were made to the event. We briefly explain the main
differences in Section~\ref{revisions}.  Afterwards,
Section~\ref{pddl} describes the input language used, named {\em
  PDDL2.2}: the first three levels of PDDL2.1, extended with
\emph{derived predicates} and \emph{timed initial literals}.
Section~\ref{systems} lists and briefly explains the competing
systems. Section~\ref{results} then presents, for each benchmark
domain, a selection of results plots, highlighting the most important
points. The entire set of data points is far too large to be presented
in detail. The full data, including plots for all results, is
available in an online appendix.  Section~\ref{awards} explains how we
awarded the competition prizes, Section~\ref{conclusion} closes the
paper with some concluding remarks. Appendix~\ref{bnf} gives a BNF
description of PDDL2.2.

%%% Local Variables: 
%%% mode: latex
%%% TeX-master: t
%%% End: 

\section{Main Revisions made in IPC-4}
\label{revisions}

The main revisions we made to the deterministic part of IPC-4, in
difference to its predecessors, were the following.

{\bf Competition Workshop.} We ran an international workshop on the
competition one year before the event itself
\cite{Own:WS-Proceedings}, providing the involved groups of people
(system developers, organizing committee, AI Planning researchers in
general) with an opportunity to express their views on issues related
to the IPC. Discussions, technical talks, and panels covered all
relevant topics ranging from the event's organizational structure and
its input language to the selection of benchmarks and the evaluation
of results. The workshop was especially useful for us as the
organizers, giving us direct feedback on the event's organization.

{\bf PDDL Extensions.} There was large agreement in the community that
PDDL2.1 still posed a significant challenge, and so the IPC-4 language
featured only relatively minor extensions. We added language features
for \emph{derived predicates} and \emph{timed initial literals}. The
resulting language is called PDDL2.2, and keeps the 3-leveled
structure of PDDL2.1. Both new language features are practically
motivated and were put to use in some of the IPC-4 domains. Derived
predicates add a form of domain axioms to PDDL2.1. A typical use is to
formulate indirect consequences of a planner's actions.  Timed initial
literals add to the temporal part of PDDL2.1 (level 3) 
a way of defining literals that will become true at a certain time point,
independently of the actions taken by the planner. A typical use is to
formulate time windows and/or goal deadlines.

{\bf Application-Oriented Benchmarks.} Our main effort in the
organization of IPC-4 was to devise a range of interesting benchmark
domains, oriented at (and as close as possible to) real-world
application domains. We collaborated with a number of people to
achieve this goal. The description of the application domains, and of
our PDDL2.2 adaptations, is long (50+ pages). It is submitted to this
same JAIR special track \cite{Benchmarks}. The application domains we
modelled were:
\begin{itemize}
\item \emph{Airport}, modelling airport ground traffic control
  \cite{hatzack:nebel:ecp-01,trueg:etal:ki-04}.
\item \emph{Pipesworld}, modelling oil derivative transportation in
  pipeline networks
  \cite{milidiu:et:al:icaps-ws-03,milidiu:liporace-hms-04}.
\item \emph{Promela}, modelling deadlock detection in communication
  protocols formulated in the Pro\-me\-la language \cite{edelkamp:spin-03,edelkamp:icaps-ws-03}.
\item \emph{PSR}, a deterministic variant of the Power Supply
  Restoration benchmark
  \cite{bertoli:etal:ecai-02,bonet:thiebaux:icaps-03}.
\item {\em UMTS}, modelling the task of scheduling the concurrent
  application setup in UMTS mobile devices~\cite{Englert:TP4,Roman:Journal}.
\end{itemize}
In addition, we re-used the \emph{Satellite} and \emph{Settlers}
domains from IPC-3. In the case of \emph{Satellite}, we added some
additional domain versions to model more advanced aspects of the
application (namely, the sending of the data to earth). Each domain is
described in a little more detail in the part (sub-section) of
Section~\ref{results} that contains the respective competition
results.

{\bf Domain Compilations.} For the first time in the IPC, we provided
domain formulations where problem constraints were compiled from the
more complex PDDL subsets into the less complex subsets. In many of
our domains the most natural domain formulation comes with complex
precondition formulas and conditional effects, i.e., in ADL. We
compiled these domain formulations to STRIPS. In previous IPCs, for
example in the {\em Elevator} domain used in IPC-2, the more
interesting problem constraints were dropped in the STRIPS domain
versions. By using the compilation approach, we hope and believe to
have created a structurally much more interesting range of STRIPS
benchmarks. The ADL and STRIPS encodings of the same instances were
posed to the competitors in an optional way, i.e., of every domain
{\em version} there could be several domain version {\em
  formulations}. The competitors were allowed to choose the
formulation they liked best. The results within a domain version were
evaluated together, in order to keep the number of separation lines in
the data at an acceptable level.

We also employed compilations encoding the new PDDL2.2 language
features in terms of artificial constructs with the old PDDL2.1
features. By this we intended to enable as wide participation in the
domains as possible. The compiled domains were offered as separate
domain {\em versions} rather than alternative formulations because, in
difference to the ADL/STRIPS case, we figured that the compiled
domains were too different from the original ones to allow joint
evaluation. Most importantly, when compiling derived predicates or
timed initial literals away, the plan length increases. Details about
the compilation methods and about the arrangement of the individual
domains are in the paper describing these domains \cite{Benchmarks}.

{\bf Optimal vs. Satisficing Planning.} We define {\em optimal}
planners as planners that prove a guarantee on the quality of the
found solution.  Opposed to that, {\em satisficing} planners are
planners that do {\em not} prove any guarantee other than the
correctness of the solution.\footnote{Satisficing planners were
  referred to as ``sub-optimal'' when IPC-4 was running. We decided to
  replace the term since it is a bit misleading. While not
  guaranteeing optimal solutions, a satisficing planner may well
  produce such solutions in some cases.} Previous IPCs did not
distinguish between optimal planners and satisficing planners. But
that is not fair since optimal planners are essentially solving a
different problem. The theoretical hardness of a domain can differ for
the different kinds of planners. In fact, it was recently proved that,
in most benchmark domains, satisficing planning is easy (polynomial)
while optimal planning is hard (NP-complete)
\cite{helmert:ecp-01,helmert:ai-03}. In practice, i.e., in most of the
commonly used benchmark domains, nowadays there is indeed a huge
performance gap between optimal and satisficing planners. In IPC-4, we
separated them into different tracks. The optimal track attracted
seven systems; for example, the planning as satisfiability approach,
that had disappeared in IPC-3, resurfaced.

{\bf Competition Booklet and Results Posters.} At previous
competitions, at conference time one could neither access the results
obtained in the competition, nor descriptions of the competing
planners. This is clearly a drawback, especially given the growing
complexity of the event. For ICAPS 2004, we assembled a booklet
containing extended abstracts describing the core aspects of all
competing systems \cite{Own:Comp-Proceedings}; the booklet was
distributed to all conference participants. The competition results,
i.e., those of the deterministic part, were made available in the form
of posters showing runtime and plan quality plots.

\begin{sloppypar}
{\bf IPC Web-page.} An important repository for any large-scale
competition event is a web page containing all the relevant information --
benchmark problems, language descriptions, result files, etc. For
IPC-4, we have set this page up at the permanent Internet address
\texttt{http://ipc.icaps-conference.org}. In the long run, this
address is intended to provide an entry point to {\em the IPC event as
  a whole}, thereby avoiding the need to look up the pages for the
different IPC editions at completely separate points in the web.
\end{sloppypar}

As a less positive change from IPCs 2 and 3 to IPC-4, the track for
{\em hand-tailored} planners disappeared. The reason for this is
simply that no such systems registered as competitors. Of course, this
is not a coincidence. There is a large agreement in the community that
there are several problems with the hand-tailored track as ran in
IPC-2 and IPC-3. The most important criticism is that, for
hand-tailored planners, ``performance'' is not just runtime and plan
quality results on a set of benchmarks. What's also important -- maybe
the most important aspect of all -- is how much effort was spent in
achieving these results: {\em how hard is it to tailor the planner?}
While the latter obviously is an important question, likewise
obviously there isn't an easy answer. We discussed the matter with a
lot of people (e.g., in the ICAPS'03 workshop), but no-one could come
up with an idea that seemed adequate and feasible. So basically we
offered the hand-tailored planners the opportunity to participate in a
track similar to the previous ones, maybe with some additional ad-hoc
measurements such as how many person hours were spent in tailoring the
planner to a domain. Apart from the evaluation shortcomings of such an
approach, another important reason why no hand-tailored planners
registered is that participating in that track is a huge amount of
work -- maintaining/developing the planner {\em plus} understanding
the domains and tailoring the planner. Understanding the domains would
have been particularly hard with the more complex domains used in
IPC-4. It is thus, obviously, difficult to find enough time to
participate in a hand-tailored IPC. Some more on the future of the
hand-tailored track is said in Section~\ref{conclusion}.

%%% Local Variables: 
%%% mode: latex
%%% TeX-master: t
%%% End: 

\section{PDDL 2.2}
\label{pddl}

As said, the IPC-4 competition language PDDL2.2 is an extension of the
first three levels of PDDL2.1 \cite{fox:long:jair-03}. PDDL2.2
inherits the separation into the levels. The language features added
on top of PDDL2.1 are {\em derived predicates} (into levels 1,2, and
3) and {\em timed initial literals} (into level 3 only). We now
discuss these two features in that order, describing their syntax and
semantics. A full BNF description of PDDL2.2 is in Appendix~\ref{bnf}.

\subsection{Derived Predicates}
\label{pddl:derived}

Derived predicates have been implemented in several planning systems
in the past, for example in UCPOP \cite{penberthy:weld:kr-92}. They
are predicates that are not affected by any of the actions available
to the planner. Instead, the predicate's truth values are derived by a
set of rules of the form {\bf if} $\phi(\overline{x})$ {\bf then}
$P(\overline{x})$.  The semantics, roughly, are that an instance of a
derived predicate (a derived predicate whose arguments are
instantiated with constants; a {\em fact}, for short) is TRUE if and
only if it can be derived using the available rules (more details
below). Under the name ``axioms'', derived predicates were a part of
the original PDDL language defined by McDermott \cite{pddl-handbook}
for the first planning competition, but they have never been put to
use in a competition benchmark (we use the name ``derived predicates''
instead of ``axioms'' in order to avoid confusion with safety
conditions).

Derived predicates combine several key aspects that made them a useful
language extension for IPC-4:

\begin{itemize}
\item They are practically motivated: in particular, they provide a
  concise and convenient means to express updates on the transitive
  closure of a relation. Such updates occur in domains that include
  structures such as paths or flows (electricity flows, chemical
  flows, etc.); in particular, the {\em PSR} domain includes this kind
  of structure.
\item They are also theoretically justified in that compiling them
  away can be infeasible. It was recently proved that, in the worst
  case, compiling derived predicates away results in an exponential
  blow up of either the problem description or the plan length
  \cite{thiebaux:etal:ijcai-03,thiebaux:etal:ai-05}.
\item Derived predicates do not cause a significant implementation
  overhead in, at least, the state transition routines used by forward
  search planners. When the world state -- the truth values of all
  non-derived, {\em basic}, predicates -- is known, computing the
  truth values of the derived predicates is trivial.\footnote{Note,
    though, that it may be much less trivial to adapt a heuristic
    function to handle derived predicates; this is, for example,
    discussed by \citeA{thiebaux:etal:ijcai-03,thiebaux:etal:ai-05}.}
\end{itemize}

In the IPC-4 benchmarks, derived predicates were used only in the
non-durational context, PDDL2.2 level 1.

\subsubsection{Syntax}
\label{pddl:derived:syntax}

The BNF definition of derived predicates involves just two small
modifications to the BNF definition of PDDL2.1:

\begin{center}
  {\tt {\small $\langle$structure-def$\rangle$ ::=\req{:derived-predicates}
      $\langle$derived-def$\rangle$}}
\end{center}

The domain file specifies a list of ``structures''. In PDDL2.1 these
were either actions or durational actions. Now we also allow
``derived'' definitions at these points.

\begin{center}
  {\tt {\small $\langle$derived-def$\rangle$ ::= (:derived
      $\langle$atomic formula(term)$\rangle$ $\langle$GD$\rangle$)}}
\end{center}

The ``derived'' definitions are the ``rules'' mentioned above. They
simply specify the predicate $P$ to be derived (with variable vector
$\overline{x}$), and the formula $\phi(\overline{x})$ from which
instances of $P$ can be concluded to be true. Syntactically, the
predicate and variables are given by the {\tt $\langle$atomic
  formula(term)$\rangle$} expression, and the formula is given by {\tt
  $\langle$GD$\rangle$} (a ``goal description'', i.e. a formula).

The BNF is more generous than what we actually allow in PDDL2.2,
respectively in IPC-4. We make a number of restrictions to ensure that
the definitions make sense and are easy to treat algorithmically. We
call a predicate $P$ {\em derived} if there is a rule that has a
predicate $P$ in its head; otherwise we call $P$ {\em basic}. The
restrictions we make are the following.

\begin{enumerate}
\item The actions available to the planner do not affect the derived
  predicates: no derived predicate occurs on any of the effect lists
  of the domain actions.
\item If a rule defines that $P(\overline{x})$ can be derived from
  $\phi(\overline{x})$, then the variables in $\overline{x}$ are
  pairwise different (and, as the notation suggests, the free
  variables of $\phi(\overline{x})$ are exactly the variables in
  $\overline{x}$).
\item If a rule defines that $P(\overline{x})$ can be derived from
  $\phi$, then the Negation Normal Form (NNF) of $\phi(\overline{x})$
  does not contain any derived predicates in negated form.
\end{enumerate}

The first restriction ensures that there is a separation between the
predicates that the planner can affect (the basic predicates) and
those (the derived predicates) whose truth values follow from the
basic predicates. The second restriction ensures that the rule right
hand sides match the rule left hand sides. Let us explain the third
restriction. The NNF of a formula is obtained by ``pushing the
negations downwards'', i.e. transforming $\neg \forall x: \phi$ into
$\exists x: (\neg \phi)$, $\neg \exists x: \phi$ into $\forall x:
(\neg \phi)$, $\neg \bigvee \phi_i$ into $\bigwedge (\neg \phi_i)$,
and $\neg \bigwedge \phi_i$ into $\bigvee (\neg \phi_i)$. Iterating
these transformation steps, one ends up with a formula where negations
occur only in front of atomic formulas -- predicates with variable
vectors, in our case. The formula contains a predicate $P$ {\em in
  negated form} if and only if there is an occurrence of $P$ that is negated. By
requiring that the formulas in the rules (that derive predicate
values) do not contain any derived predicates in negated form, we
ensure that there can not be any negative interactions between
applications of the rules (see the semantics below).

An example of a derived predicate is the ``above'' predicate in the
{\sl Blocksworld}, which is true between blocks $x$ and $y$ whenever
$x$ is transitively (possibly with some blocks in between) on $y$.
Using the derived predicates syntax, this predicate can be defined as
follows.

{\small
\begin{verbatim}
(:derived (above ?x ?y)
  (or (on ?x ?y) 
      (exists (?z) (and (on ?x ?z) 
                        (above ?z ?y)))))
\end{verbatim}}
  
  Note that formulating the truth value of ``above'' in terms of the
  effects of the normal {Blocksworld} actions is very awkward.
  Since the set of atoms affected by an action depends on the
  situation, one either needs artificial actions or complex
  conditional effects.

\subsubsection{Semantics}
\label{pddl:derived:semantics}

We now describe the updates that need to be made to the PDDL2.1
semantics definitions given by \citeA{fox:long:jair-03}. We introduce
formal notations to capture the semantics of derived predicates. We
then ``hook'' these semantics into the PDDL2.1 language by modifying
two of \citeA{fox:long:jair-03}'s definitions.

Say we are given the truth values of all (instances of the) basic
predicates, and want to compute the truth values of the (instances of
the) derived predicates from that. We are in this situation every time
we have applied an action, or parallel action set. In the durational
context, we are in this situation at the ``happenings'' in our current
plan, that is every time a durative action starts or finishes.
Formally, what we want to have is a function ${\mathcal D}$ that maps
a set of basic facts (instances of basic predicates) to the same set
but enriched with derived facts (the derivable instances of the
derived predicates). Assume we are given the set $R$ of rules for the
derived predicates, where the elements of $R$ have the form
$(P(\overline{x}), \phi(\overline{x}))$ -- {\bf if}
$\phi(\overline{x})$ {\bf then} $P(\overline{x})$. Then ${\mathcal
  D}(s)$, for a set of basic facts $s$, is defined as follows.
\begin{equation}\label{eq-derived}
  {\mathcal D}(s) := \bigcap \{ s' \mid s \subseteq s', \forall
  (P(\overline{x}), \phi(\overline{x})) \in R: \forall \overline{c},
  |\overline{c}| = |\overline{x}|: (s' \models \phi(\overline{c}) \Rightarrow P(\overline{c}) \in s') \}
\end{equation}
This definition uses the standard notations of the modelling relation
$\models$ between states (represented as sets of facts in our case)
and formulas, and of the substitution $\phi(\overline{c})$ of the free
variables in formula $\phi(\overline{x})$ with a constant vector
$\overline{c}$. In words, ${\mathcal D}(s)$ is the intersection of all
supersets of $s$ that are closed under application of the rules $R$.

Remember that we restrict the rules to not contain any derived
predicates in negated form. This implies that the order in which the
rules are applied to a state does not matter (we can not ``lose'' any
derived facts by deriving other facts first). This, in turn, implies
that ${\mathcal D}(s)$ is itself closed under application of the rules
$R$. In other words, ${\mathcal D}(s)$ is the least fixed point over
the possible applications of the rules $R$ to the state where all
derived facts are assumed to be FALSE (represented by their not being
contained in $s$). 

More constructively, ${\mathcal D}(s)$ can be computed by the
following simple process.

\begin{tabbing}
$s' := s$\\
{\bf do} \=\\
\> {\bf select} \= a rule $(P(\overline{x}), \phi(\overline{x}))$ and a
vector $\overline{c}$ of constants,\\
\> \> with $|\overline{c}| = |\overline{x}|$, such that $s' \models \phi(\overline{c})$ and $P(\overline{c}) \not \in s'$\\
\> let $s' := s' \cup \{ P(\overline{c}) \}$\\
{\bf until} no such rule and constant vector exist\\
let ${\mathcal D}(s) := s'$
\end{tabbing}

In words, apply the applicable rules in an arbitrary order until no
new facts can be derived anymore.

We can now specify what an executable plan is in PDDL2.1 with derived
predicates. All we need to do is to hook the function ${\mathcal D}$
into Definition 13, ``Happening Execution'', given by
\citeA{fox:long:jair-03}. By this definition, \citeA{fox:long:jair-03}
define the state transitions in a plan. The happenings in a (temporal
or non-temporal) plan are all time points at which at least one action
effect occurs. \citeA{fox:long:jair-03}'s definition is this:

\bigskip

\noindent
{\bf Definition 13 Happening Execution} \cite{fox:long:jair-03}
\\ {\em Given a state, $(t,s,{\bf x})$ and a
  happening, $H$, the {\em activity} for $H$ is the set of grounded
  actions 
\begin{tabbing}
$A_{H} = \{a |$ \= $\mbox{the name for $a$ is in $H$, $a$ is
    valid and}$\\
\> $Pre_{a} \,\, \mbox{is satisfied in}\,\, (t,s,{\bf
    x})\}$
\end{tabbing}
The {\em result of executing a happening}, $H$, associated with time
$t_{H}$, in a state $(t,s,{\bf x})$ is undefined if $|A_{H}| \not=
|H|$ or if any pair of actions in $A_{H}$ is mutex. Otherwise, it is
the state $(t_{H},s',{\bf x'})$ where 
\begin{equation}\label{eq-foxlong-result}
s' = (s \setminus \bigcup_{a \in A_{H}} Del_{a}) \cup \bigcup_{a \in
  A_{H}} Add_{a} \hspace{0.5cm}
\end{equation}
and ${\bf x'}$ is the result of applying the composition of the
functions $\{${\em NPF}$_{a} \, | \, a \in A_{H}\}$ to {\bf x}.}

\medskip

Note that the happenings consist of grounded actions, i.e., all
operator parameters are instantiated with constants. To introduce the
semantics of derived predicates, we now modify the result of executing
the happening. (We will also adapt the definition of mutex actions,
see below.) The result of executing the happening is now obtained by
applying the actions to $s$, then subtracting all derived facts from
this, then applying the function ${\mathcal D}$.  That is, in the
above definition we replace Equation~\ref{eq-foxlong-result} with the
following:
\begin{equation}\label{eq-result-derived}
s' = {\mathcal D}(((s \setminus \bigcup_{a \in A_{H}} Del_{a}) \cup
\bigcup_{a \in A_{H}} Add_{a}) \setminus D)
\end{equation}
where $D$ denotes the set of all derived facts. If there are no
derived predicates, $D$ is the empty set and ${\mathcal D}$ is the
identity function.

As an example, say we have a {\sl Blocksworld} instance where A is on
B is on C, $s = \{ clear(A)$, $on(A,B)$, $on(B,C)$, $ontable(C)$,
$above(A,B)$, $above(B,C)$, $above(A,C) \}$, and our happening applies
an action that moves A to the table. Then the happening execution
result will be computed by removing $on(A,B)$ from $s$, adding
$clear(B)$ and $ontable(A)$ into $s$, removing all of $above(A,B)$,
$above(B,C)$, and $above(A,C)$ from $s$, and applying ${\mathcal D}$
to this, which will re-introduce (only) $above(B,C)$.  So $s'$ will be
$s' = \{ clear(A),$ $ontable(A),$ $clear(B),$ $on(B,C),$ $ontable(C),$
$above(B,C)$ $\}$.

By the definition of happening execution, \citeA{fox:long:jair-03}
define the state transitions in a plan. The definitions of what an
executable plan is, and when a plan achieves the goal, are then
standard. The plan is {\em executable} if the result of all happenings
in the plan is defined. This means that all action preconditions have
to be fulfilled in the state of execution, and that no two pairs of
actions in a happening are {\em mutex}. The plan {\em achieves the
  goal} if the goal holds true in the state that results after the
execution of all actions in the plan.

With our above extension of the definition of happening executions,
the definitions of plan executability and goal achievement need not be
changed. We do, however, need to adapt the definition of when a pair
of actions is mutex. This is important if the happenings can contain
more than one action, i.e., if we consider parallel (Graphplan-style)
or concurrent (durational) planning. \citeA{fox:long:jair-03} give a
conservative definition that forbids the actions to interact in any
possible way.  The definition is the following.

\bigskip

\noindent
{\bf Definition 12 Mutex Actions} \cite{fox:long:jair-03}
\\ 
{\em Two grounded actions,
$a$ and $b$ are {\em non-interfering} if 
{\small
\begin{equation}\label{eq-foxlong-mutex}
  \begin{array}{c}
  GPre_{a} \cap (Add_{b} \cup Del_{b}) = GPre_{b} \cap (Add_{a} \cup Del_{a}) = \emptyset,\\
  Add_{a} \cap Del_{b} = Add_{b} \cap Del_{a} = \emptyset,\\
  L_{a} \cap R_{b} = R_{a} \cap L_{b} = \emptyset,\\
  L_{a} \cap L_{b} \subseteq L^{*}_{a} \cup L^{*}_{b}
\end{array}
\end{equation}
}
If two actions are not non-interfering they are {\em mutex}.}

\medskip

Note that the definition talks about grounded actions where all
operator parameters are instantiated with constants. $L_{a}$, $L_b$,
$R_a$, and $R_b$ refer to the left and right hand side of $a$'s and
$b$'s numeric effects. $Add_a$/$Add_b$ and $Del_a$/$Del_b$ are $a$'s
and $b$'s positive (add) respectively negative (delete) effects.
$GPre_a$/$Gpre_b$ denotes all (ground) facts that occur in $a$'s/$b$'s
precondition. If a precondition contains quantifiers then these are
grounded out ($\forall x$ transforms to $\bigwedge c_i$, $\exists x$
transforms to $\bigvee c_i$ where the $c_i$ are all objects in the
given instance), and $GPre$ is defined over the resulting
quantifier-free (and thus variable-free) formula. Note that this
definition of mutex actions is very conservative -- if, for example,
fact $F$ occurs only positively in $a$'s precondition, then it does
not matter if $F$ is among the add effects of $b$. The conservative
definition has the advantage that it makes it algorithmically very
easy to figure out if or if not $a$ and $b$ are mutex.

In the presence of derived predicates, the above definition needs to
be extended to exclude possible interactions that can arise indirectly
due to derived facts, in the precondition of the one action, whose
truth value depends on the truth value of (basic) facts affected by
the effects of the other action. In the same spirit in that
\citeA{fox:long:jair-03} forbid any possibility of direct interaction,
we now forbid any possibility of indirect interaction. Assume we
ground out all rules $(P(\overline{x}), \phi(\overline{x}))$ for the
derived predicates, i.e., we insert all possible vectors
$\overline{c}$ of constants; we also ground out the quantifiers in the
formulas $\phi(\overline{c})$, ending up with variable free rules. We
define a directed graph where the nodes are (ground) facts, and an
edge from fact $F$ to fact $F'$ is inserted iff there is a grounded
rule $(P(\overline{c}), \phi(\overline{c}))$ such that $F' =
P(\overline{c})$, and $F$ occurs in $\phi(\overline{c})$. Now say we
have an action $a$, where all ground facts occurring in $a$'s
precondition are, see above, denoted by $GPre_a$. By $DPre_a$ we
denote all ground facts that can possibly influence the truth values
of the derived facts in $GPre_a$:
\begin{equation}\label{eq-derivedP}
DPre_a := \{ F \mid \mbox{there is a path from $F$ to an $F' \in GPre_a$}
\}
\end{equation}
The definition of mutex actions is now updated simply by replacing
Equation~\ref{eq-foxlong-mutex} with: {\small
\begin{equation}\label{eq-derivedmutex}
\begin{array}{c}
  (DPre_a \cup GPre_{a}) \cap (Add_{b} \cup Del_{b}) =\\
 (DPre_b \cup GPre_b) \cap (Add_{a} \cup Del_{a}) = \emptyset,\\
  Add_{a} \cap Del_{b} = Add_{b} \cap Del_{a} = \emptyset,\\
  L_{a} \cap R_{b} = R_{a} \cap L_{b} = \emptyset,\\
  L_{a} \cap L_{b} \subseteq L^{*}_{a} \cup L^{*}_{b}
\end{array}
\end{equation}
}Note that the only thing that has changed is the first line,
regarding interference of propositional effects and preconditions. As
an example, reconsider the {\sl Blocksworld} and the ``above''
predicate.  Assume that the action that moves a block $A$ to the table
requires as an additional, derived, precondition, that $A$ is above
some third block. Then, in principle, two actions that move two
different blocks $A$ and $B$ to the table can be executed in parallel.
Which block $A$ ($B$) is on can influence the $above$ relations in
that $B$ ($A$) participates; however, this does not matter because if
$A$ and $B$ can be both moved then this implies that they are both
clear, which implies that they are on top of different stacks anyway.
We observe that the latter is a statement about the domain semantics
that either requires non-trivial reasoning, or access to the world
state in which the actions are executed. In order to avoid the need to
either do non-trivial reasoning about domain semantics, or resort to a
forward search, our definition is the conservative one given above.
The definition makes the actions moving $A$ and $B$ mutex on the
grounds that they can possibly influence each other's derived
preconditions.

The definition adaptations described above suffice to define the
semantics of derived predicates for the whole of PDDL2.2.
\citeA{fox:long:jair-03} reduce the temporal case to the case of
simple plans above, so by adapting the simple-plan definitions we have
automatically adapted the definitions of the more complex cases. In
the temporal setting, PDDL2.2 level 3, the derived predicates
semantics are that their values are computed anew at each happening in
the plan where an action effect occurs. As said, in IPC-4 we used
derived predicates only in the non-temporal setting. Some remarks on
limitations of the IPC-4 treatment of derived predicates, and on
future prospects, are in Section~\ref{conclusion}.

\subsection{Timed Initial Literals}
\label{pddl:timed}

Timed initial literals are a syntactically very simple way of
expressing a certain restricted form of exogenous events: facts that
will become TRUE or FALSE at time points that are known to the planner
in advance, independently of the actions that the planner chooses to
execute. Timed initial literals are thus deterministic unconditional
exogenous events. Syntactically, we simply allow the initial state to
specify -- beside the usual facts that are true at time point $0$ --
literals that will become true at time points greater than $0$.

Timed initial literals are practically very relevant: in the real
world, deterministic unconditional exogenous events are very common,
typically in the form of time windows -- within which a shop has
opened, within which humans work, within which traffic is slow, within
which there is daylight, within which a seminar room is occupied,
within which nobody answers their mail because they are all at
conferences, etc. The timed initial literals syntax is just about the
simplest way one can think of to communicate such things to a planner.

Timed initial literals can easily be compiled into artificial
constructs \cite{expressivepddl}, involving an only linear blow-up in
the instance representation and in plan length (i.e., number of
actions in it). Still it seems highly likely that handing the timed
literals over to an automated planner explicitly results in far better
performance than when one hands over the artificial (compiled)
representation. The results obtained in IPC-4 confirm this, see also
Section~\ref{results}.

\subsubsection{Syntax}
\label{pddl:timed:syntax}

The BNF notation is:

\begin{center}
  {\tt {\small $\langle$init$\rangle$ ::= (:init $\langle$init-el$\rangle$\zom)}}
\end{center}

\begin{center}
  {\tt {\small $\langle$init-el$\rangle$ ::=\req{:timed-initial-literals} (at $\langle$number$\rangle$ $\langle$literal(name)$\rangle$) }}
\end{center}

The requirement flag for timed initial literals implies the
requirement flag for durational actions, i.e., as said the language
construct is only available in PDDL2.2 level 3. The times {\tt
  $\langle$number$\rangle$} at which the timed literals occur are
restricted to be greater than $0$.  If there are also derived
predicates in the domain, then the timed literals are restricted to
not influence any of these, i.e., like action effects they are only
allowed to affect the truth values of the basic (non-derived)
predicates (IPC-4 will not use both derived predicates and timed
initial literals within the same domain).

As an illustrative example, consider a planning task where the goal is
to have completed the shopping. There is a single action {\sl
  go-shopping} that achieves the goal, and requires the (single) shop
to be open as the precondition. The shop opens at time 9 relative to
the initial state, and closes at time 20. We can express the shop
opening times by two timed initial literals:

{\small
\begin{verbatim}
(:init
  (at 9 (shop-open))
  (at 20 (not (shop-open)))
)
\end{verbatim}}

\subsubsection{Semantics}
\label{pddl:timed:semantics}

We now describe the updates that need to be made to the PDDL2.1
semantics definitions given by \citeA{fox:long:jair-03}. Adapting two
of the definitions suffices.

The first definition we need to adapt is the one that defines what a
``simple plan'', and its happening sequence, is. The original
definition by \citeA{fox:long:jair-03} is this.

\bigskip

\noindent
{\bf Definition 11 Simple Plan} \cite{fox:long:jair-03}\\
{\em A {\em simple plan}, $SP$, for a planning instance, $I$, consists
  of a finite collection of {\em timed simple actions} which are pairs
  $(t,a)$, where $t$ is a rational-valued time and $a$ is an action
  name.
  
  The {\em happening sequence}, $\{t_{i}\}_{i=0\ldots k}$ for $SP$ is
  the ordered sequence of times in the set of times appearing in the
  timed simple actions in $SP$. All $t_{i}$ must be greater than $0$.
  It is possible for the sequence to be empty (an empty plan).
  
  The {\em happening} at time $t$, $E_{t}$, where $t$ is in the
  happening sequence of $SP$, is the set of (simple) action names that
  appear in timed simple actions associated with the time $t$ in
  $SP$.}

\medskip

In the STRIPS case, the time stamps are the natural numbers $1, \dots,
n$ when there are $n$ actions/parallel action sets in the plan. The
happenings then are the actions/parallel action sets at the respective
time steps. \citeA{fox:long:jair-03} reduce the temporal planning case
to the simple plan case defined here by splitting each durational
action up into at least two simple actions -- the start action, the
end action, and possibly several actions in between that guard the
durational action's invariants at the points where other action
effects occur. So in the temporal case, the happening sequence is
comprised of all time points at which ``something happens'', i.e., at
which some action effect occurs.

To introduce our intended semantics of timed initial literals, all we
need to do to this definition is to introduce additional happenings
into the temporal plan, namely the time points at which some timed
initial literal occurs. The timed initial literals can be interpreted
as simple actions that are forced into the respective happenings
(rather than selected into them by the planner), whose precondition is
true, and whose only effect is the respective literal. The rest of
\citeA{fox:long:jair-03}'s definitions then carry over directly
(except goal achievement, which involves a little care, see below).
The PDDL2.2 definition of simple plans is this here.

\bigskip

\noindent
{\bf Definition 11 Simple Plan}\\ {\em A {\em simple plan}, $SP$, for
  a planning instance, $I$, consists of a finite collection of {\em
    timed simple actions} which are pairs $(t,a)$, where $t$ is a
  rational-valued time and $a$ is an action name. By $t_{end}$ we
  denote the largest time $t$ in $SP$, or $0$ if $SP$ is empty.
  
  Let $TL$ be the (finite) set of all timed initial literals, given as
  pairs $(t,l)$ where $t$ is the rational-valued time of occurrence of
  the literal $l$. We identify each timed initial literal $(t,l)$ in
  $TL$ with a uniquely named simple action that is associated with
  time $t$, whose precondition is TRUE, and whose only effect is $l$.

  The {\em happening sequence}, $\{t_{i}\}_{i=0\ldots k}$ for $SP$ is
  the ordered sequence of times in the set of times appearing in the
  timed simple actions in $SP$ and $TL$. All $t_{i}$ must be greater than $0$.
  It is possible for the sequence to be empty (an empty plan).
  
  The {\em happening} at time $t$, $E_{t}$, where $t$ is in the
  happening sequence of $SP$, is the set of (simple) action names that
  appear in timed simple actions associated with the time $t$ in
  $SP$ or $TL$.}

\medskip

Thus the happenings in a temporal plan are all points in time where
either an action effect, or a timed literal, occurs. The timed
literals are simple actions forced into the plan. With this
construction, \citeA{fox:long:jair-03}'s Definitions~12 (Mutex
Actions) and~13 (Happening Execution), as described (and adapted to
derived predicates) in Section~\ref{pddl:derived:semantics}, can be
kept unchanged.  They state that no action effect is allowed to
interfere with a timed initial literal, and that the timed initial
literals are true in the state that results from the execution of the
happening they are contained in. \citeA{fox:long:jair-03}'s
Definition~14 (Executability of a plan) can also be kept unchanged --
the timed initial literals change the happenings in the plan, but not
the conditions under which a happening can be executed.

The only definition we need to reformulate is that of what the {\em
  makespan} of a valid plan is. In \citeA{fox:long:jair-03}'s original
definition, this is implicit in the definition of valid plans. The
definition is this.

\bigskip

\noindent
{\bf Definition 15 Validity of a Simple Plan} \cite{fox:long:jair-03}\\
{\em A simple plan (for a planning instance, $I$) is {\em valid} if it
  is executable and produces a final state $S$, such that the goal
  specification for $I$ is satisfied in $S$.}

\medskip

The makespan of the valid plan is accessible in PDDL2.1 and PDDL2.2 by
the ``total-time'' variable that can be used in the optimization
expression. Naturally, \citeA{fox:long:jair-03} take the makespan to
be the end of the plan, the time point of the plan's final state.

In the presence of timed initial literals, the question of what the
plan's makespan is becomes a little more subtle. With
\citeA{fox:long:jair-03}'s above original definition, the makespan
would be the end of all happenings in the simple plan, which {\em
  include} all timed initial literals (see the revised Definition 11
above). So the plan would at least take as long as it takes until no
more timed literals occur. But a plan might be finished long before
that -- imagine something that needs to be done while there is
daylight; certainly the plan does not need to wait until sunset. We
therefore define the makespan to be the earliest point in time at
which the goal condition becomes (and remains) true. Formally this
reads as follows.

\bigskip

\noindent
{\bf Definition 15 Validity and Makespan of a Simple Plan}\\
{\em A simple plan (for a planning instance, $I$) is {\em valid} if it
  is executable and produces a final state $S$, such that the goal
  specification for $I$ is satisfied in $S$. The plan's {\em makespan}
  is the smallest $t \geq t_{end}$ such that, for all happenings at
  times $t' \geq t$ in the plan's happening sequence, the goal
  specification is satisfied after execution of the happening.}

\medskip

Remember that $t_{end}$ denotes the time of the last happening in the
plan that contains an effect caused by the plan's {\em actions} -- in
simpler terms, $t_{end}$ is the end point of the plan. What the
definition says is that the plan is valid if, at some time point $t$
after the plan's end, the goal condition is achieved and remains true
until after the last timed literal has occurred. The plan's makespan is
the first such time point $t$. Note that the planner can ``use'' the
events to achieve the goal, by doing nothing until a timed literal
occurs that makes the goal condition true -- but then the waiting time
until the nearest such timed literal is counted into the plan's
makespan.  (The latter is done to avoid situations where the planner
could prefer to wait millions of years rather than just applying a
single action itself.) Remember that the makespan of the plan, defined
as above, is what can be denoted by {\tt total-time} in the
optimization expression defined with the problem instance.

As with the derived predicates, the definition adaptations above
suffice to define the semantics of timed initial literals for PDDL2.2.
Some remarks on limitations of the language construct, and on future
prospects, are in Section~\ref{conclusion}.

%%% Local Variables: 
%%% mode: latex
%%% TeX-master: t
%%% TeX-master: t
%%% End: 

\section{The Competitors}
\label{systems}

We now provide an overview table listing the systems along with their
language capabilities (subset of PDDL2.2 covered), and we sketch the
competing systems.  The deterministic part of IPC-4 attracted 19
competitors (21 when counting different system versions), 12 (14) of
which were satisficing and 7 of which were optimal. Each competitor
wrote a 2-3 page extended abstract for inclusion in the IPC-4 booklet
at ICAPS 2004 \cite{Own:Comp-Proceedings}. The included sketches are
very brief outlines of these abstracts. The reader is encouraged to
read the original work by the system authors.

At this point it is appropriate to say a few words on system
development.  We allowed competitors to modify their systems during
the competition phase, i.e., while data collection was running. Our
reasons for doing so were twofold. First, some of our domains were
quite unusual -- most particularly, those containing ADL constructs
compiled to STRIPS -- so we (rightly) expected planner implementations
to have parsing etc. trouble with them. Second, from our own
experience with the competitions we knew that people will try to
enhance their planners anyway so we did not see much point in
forbidding this. We trusted the competitors to not do stupid things
like hard-coding domain names etc., and we trusted to have a diverse
enough range of domains to make tuning a ``domain-independent'' task.

An alternative would have been to collect executables before data
collection, and then have all the results collected by the organizers.
Anyone who has ever run experiments with someone else's planner knows
that such an approach is completely infeasible due to the
prototype-nature of the systems, and due to the many tiny details in
minor language changes, programming environments, etc. -- if one wants
to obtain meaningful results, at least. One could in principle apply a
strict ``any failure is counted as unsolved'' rule, but doing so would
likely lay way too much emphasis on little programming errors that
have nothing to do with the evaluated {\em algorithms}.

The competition phase was, roughly, from February 2004 until middle
May 2004. During that time, we released the domains one-by-one, and
the competitors worked on their systems, handing the results over to
us when they were ready. At the end of that phase, the competitors had
to submit an executable, and we ran some sampled tests to see that the
executable did really produce the reported results, across all the
domains. The abstracts describing the systems had to be delivered a
little earlier, by the end of April, due to timing constraints for
printing the booklet.

We start with the overview table, then sketch the satisficing and
optimal competitors.

\subsection{Planner Overview}
\label{systems:overview}

An overview of the participating satisficing planners, and their
language capabilities (defined as what language features they attacked
in IPC-4), is given in Table~\ref{tab:planners-overview}.

\begin{table}[htb]
\centering
\begin{small}
\centering
\begin{tabular}{|l||c|c|c|c|c|}
\hline    
              &  ADL     &  DP      &  Numbers    & Durations &  TL      \\
\hline
\hline
CRIKEY       &   -      &   -      &    +        &   +      &    -       \\\hline
FAP           &   -      &   -      &    -        &   -      &    -       \\\hline
FD, FDD       &   +      &   +      &    -        &   -      &    -       \\\hline
LPG-TD        &   +      &   +      &    +        &   +      &    +       \\\hline
Macro-FF      &   +      &   -      &    -        &   -      &    -       \\\hline
Marvin        &   +      &   +      &    -        &   -      &    -       \\\hline
Optop         &   +      &   +      &    +        &   +      &    +       \\\hline
P-MEP         &   +      &   -      &    +        &   +      &    +       \\\hline
Roadmapper    &   -      &   -      &    -        &   -      &    -       \\\hline
SGPlan        &   +      &   +      &    +        &   +      &    +       \\\hline
Tilsapa       &   -      &   -      &    +        &   +      &    +       \\\hline
YAHSP         &   -      &   -      &    -        &   -      &    -       \\\hline\hline
BFHSP         &   -      &   -      &    -        &   -      &    -       \\\hline
CPT           &   -      &   -      &    -        &   +      &    -       \\\hline
HSP$_a^*$     &   -      &   -      &    +        &   +      &    -       \\\hline
Optiplan      &   -      &   -      &    -        &   -      &    -       \\\hline
SATPLAN     &   -      &   -      &    -        &   -      &    -       \\\hline
SemSyn        &   +      &   -      &    +        &   -      &    -       \\\hline
TP4        &   -      &   -      &    +        &   +      &    -       \\\hline
\end{tabular}
\end{small}
\caption{\label{tab:planners-overview}An overview of the planners in IPC-4, and their language capabilities 
  (i.e., the language features they attacked in IPC-4). satisficing planners are in the top half of the table, optimal planners are in the bottom half.
  Each table entry specifies whether (``+'') or not (``-'') a planner 
  can handle a language feature. 
  ``DP'' is short for derived predicates, ``TL'' is short
  for timed initial literals. With ``Numbers'' and ``Durations'' 
  we mean numeric fluents and fixed action durations (no duration 
  inequalities), in the sense of PDDL2.1 level 2 and 3, respectively. The 
  planners 
  (and their name abbreviations) are explained below.}
\vspace{-0.0cm}
\end{table}

Observe that most of the planners treat only a small subset of
PDDL2.2: from a quick glance at the table, one sees that there are far
more (``-'') entries than (``+'') entries. Note here that, often, even
when a (``+'') sign is given, a planner may treat only a subset of the
specified language feature (as needed for the respective IPC-4
domain). For example, a planner might treat only a subset of ADL, or
only linear arithmetic for numeric variables. We leave out these
details for the sake of readability.

LPG-TD, Optop, and SGPlan are the only planners that treat the full
range of PDDL2.2, as used in IPC-4. Three satisficing planners (FAP,
Roadmapper, and YAHSP), and three optimal planners (BFHSP, Optiplan,
and SATPLAN), treat only pure STRIPS.  Derived predicates are treated
by 4 planners, timed initial literals by 5 planners, ADL by 7
planners, numeric variables by 8 planners, and action durations by 9
planners.

Table~\ref{tab:planners-overview} shows that there is a significant
amount of acceptance of the new (PDDL2.1 and PDDL2.2) language
features, in terms of implemented systems participating in the IPC.
Table~\ref{tab:planners-overview} also shows that the development on
the systems side has a tendency to be slower than the development on
the IPC language side: even the most wide-spread language feature
beyond STRIPS, action durations, is dealt with by less than half of
the 19 planners. In our opinion, this should be taken to indicate that
further language extensions should be made slowly. Some more on this
is said in Section~\ref{conclusion}.

\subsection{Satisficing Planners}
\label{systems:satisficing}

The satisficing planners -- the planners giving no guarantee on the
quality of the returned plan -- were the following. We proceed in
alphabetical order.

{\bf CRIKEY} by Keith Halsey, University of Strathclyde, Glasgow, UK.
CRIKEY is a heuristic search forward state space planner. It includes
a heuristic for relaxed plans for temporal and numeric problems and
applies a scheduler based on simple temporal networks to allow
posterior plan scheduling.

{\bf FAP} by Guy Camilleri and Joseph Zalaket, University Paul
Sabatier / IRIT CCI-CSC, France. FAP handles non-temporal non-numeric
domains. It is a heuristic planner using the relaxed planning
heuristic in an $N$-best forward search, with meta-actions (action
sequences) extracted from the relaxed planning graph to perform jumps
in the state space.

{\bf Fast Downward}, \emph{FD} for short, and {\bf Fast Diagonally
  Downward}, \emph{FDD} for short, by Malte Helmert and Silvia
Richter, University of Freiburg, Germany. FD and FDD can treat
non-temporal non-numeric domains. They apply a new heuristic estimate
based on a polynomial relaxation to the automatically inferred
multivariate or SAS$^+$ representation of the problem space.  FDD also
applies the traditional FF-style relaxed planning heuristic, i.e., it
applies both heuristics in a hybrid search algorithm.

{\bf LPG-TD} by Alfonso Gerevini, Alessandro Saetti, Ivan Serina, and
Paolo Toninelli, University of Brescia, Italy. The extensions to the
randomized local plan graph search that was already included in LPG at
IPC-3 includes functionality for PDDL2.2 derived predicates and timed
initial literals, as well as various implementation refinements. A
version tailored to computation speed, LPG-TD.speed, and a version
tailored for plan quality, LPG-TD.quality, participated.
LPG-TD.quality differs from LPG-TD.speed basically in that it does not
stop when the first plan is found but continues until a stopping
criterion is met. The LPG-TD team also ran a third version, called
``LPG-TD.bestquality'', which used, for every instance, the entire
half hour CPU time available to produce as good a plan as possible.
This third version is not included in the official data because every
team was allowed to enter at most two system versions.

{\bf Macro-FF} by Adi Botea, Markus Enzenberger, Martin M\"uller,
Jonathan Schaeffer, University of Alberta, Canada. As the name
suggests Macro-FF extends Hoffmann's FF planner with macro operators
(and other implementation refinements). Macros are learned prior to
the search and fed into the planner as new data, separated from the
operator file and the problem file.  At runtime, both regular actions
and macro-actions are used for state expansion.  Heuristic rules for
pruning instantiations of macros are added to FF's original strategy
for search control.

{\bf Marvin} by Andrew Coles and Amanda Smith, University of
Strathclyde, Glasgow, UK. Marvin extends FF by adding extra features
and preprocessing information, such as plateau-escaping macro actions,
to enhance the search algorithm.

{\bf Optop} by Drew McDermott, Yale University, USA. Optop is an
extension of the well-known UNPOP planner, with the ability to handle
a complex input language including, amongst other things, autonomous
processes running in parallel to the actions taken by the planner. The
underlying principle is a forward search with heuristic guidance
obtained from greedy-regression match graphs that are built backwards
from the goals.

{\bf P-MEP} by Javier Sanchez, Minh Tang, and Amol D. Mali, University
of Wisconsin, USA. P-MEP is short for Parallel More Expressive
Planner. Unlike most planners, P-MEP can treat numeric variables with
non-linear action effects. It employs a forward search with relaxed
plan heuristics, enhanced by relevance detection in a pre-process, and
by taking into account exclusion relations during relaxed planning.

{\bf Roadmapper} by Lin Zhu and Robert Givan, Purdue University, USA.
Roadmapper handles non-temporal non-numeric domains. It is a forward
heuristic search planner enhancing the FF heuristic with a reasoning
about landmarks. The latter are propositions that must be true at some
point in every legal plan.  Roadmapper finds landmarks in a
pre-process, arranges them in a directed road-map graph, and uses that
graph to assign weights to actions in the FF-style relaxed plans.

{\bf SGPlan} by Yixin Chen, Chih-Wei Hsu and Benjamin W. Wah,
University of Illinois, USA. The planner bootstraps heuristic search
planners by applying Lagrange optimization to combine the solution of
planning subproblems.  To split the problem, an ordering of the
planning goals is derived.  The incremental local search strategy that
is applied for Lagrange optimization on top of the individual planners
relies on the theory of extended saddle points for mixed integer
linear programming.

{\bf Tilsapa} by Bharat Ranjan Kavuluri and Senthil U., AIDB Lab, IIT
Madras, India. The planner extends the SAPA system that handles
temporal numeric domains, with the ability to handle timed initial
literals.

{\bf YAHSP} by Vincent Vidal, University of Artois, France.  YAHSP, an
acronym for \emph{yet another heuristic search planner}, searches
forward with a FF-style relaxed planning heuristic. The main
enhancement is a relaxed-plan-analysis phase that replaces actions in
the relaxed plan based on heuristic notions of better suitability in
reality, producing macro-actions in the process. The macro-actions
are, basically, as long as possible feasible sub-sequences of the
(modified) relaxed plan.

\subsection{Optimal Planners}
\label{systems:optimal}

The participating optimal planners were the following, in alphabetical
order.

{\bf BFHSP} by Rong Zhou and Eric A. Hansen, Mississippi State
University, USA. BFHSP, for breadth-first heuristic search planner,
optimizes the number of actions in the plan. The planner implements
the standard max-atom and max-pair heuristics (as well as the
max-triple heuristic, which was however not used in IPC-4). The main
difference to other systems is its search algorithm called {\em
  breadth-first heuristic search}; this improves the memory
requirements of A* search by searching the set of nodes up to a
certain threshold value in breadth-first instead of best-first manner.

{\bf CPT} by Vincent Vidal, University of Artois, France, and Hector
Geffner, University of Pompeu Fabra, Barcelona, Spain. CPT optimizes
makespan. The planner is based on constraint satisfaction, and
transforms the planning problem to a CSP.  The branching scheme that
solves the CSP makes use of several constraint propagation techniques
related to POCL, and of the temporal max-atom heuristic.

{\bf HSP$_a^*$} by Patrik Haslum, Link\"oping University, Sweden.
HSP*$_a$ is a derivate of TP4, see below, with the same expressivity,
but with a weakened version of the \emph{MaxTriple} heuristic instead
of the \emph{MaxPair} heuristic.

{\bf Optiplan} by Menkes van der Briel and Subbarao Kambhampati,
Arizona State University, USA. Optiplan is a planner based on integer
programming (IP), which itself is a consequent extension to the
encoding of the planning graph used in the planning as satisfiability
approach (see below). The compiled planning problem is solved using
the CPLEX/ILOG system. An interesting property of Optiplan is that,
due to the power of the underlying IP solver, it can optimize a great
variety of different plan metrics, in fact every plan metric that can
be expressed with a linear function.  In that respect, the planner is
unique. In IPC-4, step-optimal plans were computed because that is
typically most efficient.

{\bf SATPLAN04}, \emph{SATPLAN} for short, 
by Henry Kautz, David Roznyai, Farhad Teydaye-Saheli,
Shane Neph, and Michael Lindmark, University of Washington, USA.
SATPLAN treats non-tem\-po\-ral non-nu\-meric domains and optimizes the
number of parallel time steps. The system did not participate in IPC-3
but in IPC-1 and IPC-2, so it was a real comeback. As the planner
compiles a planning problem to a series of satisfiability tests, the
performance of the system relies on the integrated back-end SAT
solver. The graph-plan encoding scheme underwent only minor changes,
but the underlying SAT solver is much more powerful than that of 4
years ago.

{\bf SemSyn} by Eric Parker, Semsyn Software, Tempe, Arizona. The
\emph{SemSyn} system optimizes the number of actions. Semsyn is linked
to a commercial product and applies a combination of forward and
backward chaining. Both searches are variants of the original
algorithms.  Forward chaining is \emph{goal-directed}, while backward
chaining is \emph{generalized}, comprising a partition of the
top-level goal.  As the product is proprietary, detailed information
on the system is limited.

{\bf TP4-04}, \emph{TP4} for short, by Patrik Haslum, Link\"oping University, Sweden.  The
planner is an extended makespan-optimal scheduling system that does an
IDA* search routine with the max-pair heuristic.  It can deal with a
restricted form of metric planning problems with numerical
preconditions and effects.

\section{Results}
\label{results}

The CPU times for IPC-4 were measured on a machine located in
Freiburg, running 2 Pentium-4 CPUs at 3 GHz, with 6 GB main memory.
Competitors logged in remotely and ran their planners, producing one
separate ASCII result file for each solved instance, giving the
runtime taken, the quality of the found plan, as well as the plan
itself. Each planner run was allowed at most half an hour (CPU, not
real) runtime, and 1 GB memory, i.e., these were the cutoff values for
solving a single instance. As a kind of performance measure, the IPC-3
version of LPG \cite{gerevini:etal:jair-03} (called ``LPG-3''), i.e.
the most successful automatic planner from IPC-3, was also run (by its
developers, on top of the workload with their new planner version
LPG-TD). We remark that, for the sake of simplicity, to save CPU time,
and to ensure fairness, randomized planners were allowed only a single
run one ach instance, with a fixed random seed.

IPC-4 featured a lot of domain versions and instances, and a lot of
planners. The only way to fully understand the results of such a
complex event is to examine the results in detail, making sense of
them in combination with the descriptions/PDDL encodings of the
domains, and the techniques used in the respective planners. We
recommend doing so, at least to some extent, to everybody who is
interested in the results of (the deterministic part of) IPC-4. The
on-line appendix of this paper includes all the individual solution
files. The appendix also includes all \emph{gnuplot} graphics we
generated (more details below), as well as a GANNT chart for each of
the result files; the GANNT charts can be visualized using the Vega
visualization front-end~\cite{Hipke:Dissertation}.

In what follows, we provide an overview of the results, and we
highlight (what we think are) the most important points; we explain
how we decided about the IPC-4 competition awards.

More precisely, Section~\ref{results:overview} gives an overview over
the results, in terms of percentages of attacked and solved instances
per language subset and planner. Section~\ref{results:evaluation}
describes how we evaluated the results, particularly what procedure we
chose to decide about the awards. Thereafter,
Sections~\ref{results:airport} to~\ref{results:umts} in turn consider
in some more detail the results in the single domains.

\subsection{Results Overview}
\label{results:overview}

One crude way to obtain an overview over such a large data set is to
simply count the number of instances each planner {\em attacked}, i.e.
tried to solve, and those that it succeeded to solve. In our case, it
also makes sense to distinguish between the different PDDL2.2 subsets
used by the IPC-4 benchmarks. The data is displayed in
Table~\ref{tab:results-overview}.

\setlength{\tabcolsep}{3.5pt}

\begin{table}[t]
\centering
\begin{small}
\centering
\begin{tabular}{|l||rr|rr|rr|rr|rr|rr|rr||rrr|}
\hline    
              &  \multicolumn{2}{|c|}{N}       & \multicolumn{2}{|c|}{N+DP}    & \multicolumn{2}{|c|}{N+NV}    & \multicolumn{2}{|c|}{D}       & \multicolumn{2}{|c|}{D+TL}     & \multicolumn{2}{|c|}{D+NV}     & \multicolumn{2}{|c||}{D+TL+NV} & \multicolumn{3}{|c|}{All} \\\hline\hline
Number        &  \multicolumn{2}{|c|}{382}     &  \multicolumn{2}{|c|}{196}     &  \multicolumn{2}{|c|}{152}      & \multicolumn{2}{|c|}{302}     &  \multicolumn{2}{|c|}{116}      & \multicolumn{2}{|c|}{272}       &  \multicolumn{2}{|c||}{136} &  \multicolumn{3}{|c|}{1,556}        \\\hline\hline                                                            
CRIKEY       &42  &    77&\multicolumn{2}{|c|}{---}&    &       &47 &    66&\multicolumn{2}{|c|}{---}&98  &    55&\multicolumn{2}{|c||}{---}&56  &61 & 42 \\\hline
FAP           &33  &    64&\multicolumn{2}{|c|}{---}&\multicolumn{2}{|c|}{---}&\multicolumn{2}{|c|}{---}&\multicolumn{2}{|c|}{---}&\multicolumn{2}{|c|}{---}&\multicolumn{2}{|c||}{---}&33  &74  &16 \\\hline
FD            &83  &    62&84  &   100&\multicolumn{2}{|c|}{---}&\multicolumn{2}{|c|}{---}&\multicolumn{2}{|c|}{---}&\multicolumn{2}{|c|}{---}&\multicolumn{2}{|c||}{---}&83  &75  &28 \\\hline
FDD           &92  &    62&84  &   100&\multicolumn{2}{|c|}{---}&\multicolumn{2}{|c|}{---}&\multicolumn{2}{|c|}{---}&\multicolumn{2}{|c|}{---}&\multicolumn{2}{|c||}{---}&88  &75  &28 \\\hline
LPG-3      &55  &    62&\multicolumn{2}{|c|}{---}&42  &     24&45 &    62&\multicolumn{2}{|c|}{---}&56  &     50&\multicolumn{2}{|c||}{---}&51  &54  &38 \\\hline
LPG-TD        &67  &    87&75  &    74&61  &     37&76 &    62&63  &    100&96  &     50&87  &   100 &76  &71  &71 \\\hline
Macro-FF      &57  &    87&\multicolumn{2}{|c|}{---}&\multicolumn{2}{|c|}{---}&\multicolumn{2}{|c|}{---}&\multicolumn{2}{|c|}{---}&\multicolumn{2}{|c|}{---}&\multicolumn{2}{|c||}{---}&57  &100  &21 \\\hline
Marvin        &67  &    62&34  &   100&\multicolumn{2}{|c|}{---}&\multicolumn{2}{|c|}{---}&\multicolumn{2}{|c|}{---}&\multicolumn{2}{|c|}{---}&\multicolumn{2}{|c||}{---}&52  &75 &28  \\\hline
Optop         &    &      &\multicolumn{2}{|c|}{---}&\multicolumn{2}{|c|}{---}&   &      & 8  &     43&\multicolumn{2}{|c|}{---}&\multicolumn{2}{|c||}{---}& 8  & 7  & 3 \\\hline
P-MEP         &15  &    61&\multicolumn{2}{|c|}{---}& 8  &     55&24 &    45&24  &     43&13  &     32&    &       &17  &43  &38 \\\hline
Roadmapper    &28  &    49&\multicolumn{2}{|c|}{---}&\multicolumn{2}{|c|}{---}&\multicolumn{2}{|c|}{---}&\multicolumn{2}{|c|}{---}&\multicolumn{2}{|c|}{---}&\multicolumn{2}{|c||}{---}&28  &56 &12  \\\hline
SGPlan        &68  &   100&65  &   100&64  &    100&75 &    90&78  &     74&85  &    100&74  &   100 &73  &96 &96  \\\hline
Tilsapa       &    &      &\multicolumn{2}{|c|}{---}&    &       &   &      &10  &     69&    &       &62  &    63 &38  &13  &11 \\\hline
YAHSP         &77  &    87&\multicolumn{2}{|c|}{---}&\multicolumn{2}{|c|}{---}&\multicolumn{2}{|c|}{---}&\multicolumn{2}{|c|}{---}&\multicolumn{2}{|c|}{---}&\multicolumn{2}{|c||}{---}&77  &100 &21 \\\hline\hline
BFHSP         &29  &    87&\multicolumn{2}{|c|}{---}&\multicolumn{2}{|c|}{---}&\multicolumn{2}{|c|}{---}&\multicolumn{2}{|c|}{---}&\multicolumn{2}{|c|}{---}&\multicolumn{2}{|c||}{---}&29  &100 &21 \\\hline
CPT           &33  &    87&\multicolumn{2}{|c|}{---}&\multicolumn{2}{|c|}{---}&22 &   100&\multicolumn{2}{|c|}{---}&\multicolumn{2}{|c|}{---}&\multicolumn{2}{|c||}{---}&28  &100& 41 \\\hline
HSP$_a^*$     &33  &    38&\multicolumn{2}{|c|}{---}&    &       &10 &    62&\multicolumn{2}{|c|}{---}&50   &    50&\multicolumn{2}{|c||}{---}&29    &44 & 30 \\\hline
Optiplan      &34  &    87&\multicolumn{2}{|c|}{---}&\multicolumn{2}{|c|}{---}&\multicolumn{2}{|c|}{---}&\multicolumn{2}{|c|}{---}& \multicolumn{2}{|c|}{---}&\multicolumn{2}{|c||}{---}&34    &100 &21 \\\hline
SATPLAN       &46  &    87&\multicolumn{2}{|c|}{---}&\multicolumn{2}{|c|}{---}&\multicolumn{2}{|c|}{---}&\multicolumn{2}{|c|}{---}&\multicolumn{2}{|c|}{---}&\multicolumn{2}{|c||}{---}&46    &100&21  \\\hline
SemSyn        &49  &    48&\multicolumn{2}{|c|}{---}&11  &     37&\multicolumn{2}{|c|}{---}&\multicolumn{2}{|c|}{---}&\multicolumn{2}{|c|}{---}&\multicolumn{2}{|c||}{---}&40    &23  &15 \\\hline
TP4        &29  &    64&\multicolumn{2}{|c|}{---}&    &       &17 &    62&\multicolumn{2}{|c|}{---}&52  &     50&\multicolumn{2}{|c||}{---}&31    &54 & 37 \\\hline
\end{tabular}
\end{small}
\caption{\label{tab:results-overview}An overview of the IPC-4 results. Used abbreviations: ``N'' no durations, 
  ``DP'' derived predicates, ``NV'' numeric variables, ``D'' durations, ``TL'' timed initial literals. Line ``Number'' gives number of instances, all other lines give percentage values. In each entry, the right number is the percentage of instances that were attacked, and the left number is the percentage {\em of these} that were solved -- i.e. the left number is the success ratio. In the rightmost column, the middle number is the percentage of attacked instances relative to all instances that lie within the language range of the respective planner; the right number is the percentage of attacked instances over all. A dash indicates that a planner can not handle a language subset. See the detailed explanation in the text.}
  \vspace{-0.0cm}
\end{table}

Table~\ref{tab:results-overview} is complicated, and needs some
explanation. First, consider the columns in the table. The leftmost
column is, as usual, for table indexing. The rightmost column contains
data for the entire set of instances in IPC-4. The columns in between
all refer to a specific subset of the instances, in such a way that
these subsets are disjunct and exhaustive -- i.e., the instance sets
associated with columns are pairwise disjunct, and their union is the
entire set of instances. The subsets of instances are defined by the
PDDL2.2 subsets they use. The abbreviations in
Table~\ref{tab:results-overview} are explained in the caption. ``X+Y''
means that these instances use language features ``X'' and ``Y'', and
none of the other features. For example, ``N+DP'' are (only) those
instances with uniform durations (i.e., PDDL2.2 level 1) and derived
predicates.  Note that Table~\ref{tab:results-overview} does not have
a column for {\em all} possible combinations of language features --
we show only those that were used by (a non-empty subset of) the IPC-4
benchmarks.  Note also that we do not distinguish between different
domain {\em formulations}, i.e. between ADL and STRIPS. The instance
numbers in Table~\ref{tab:results-overview} are counted for domain
{\em versions}. That is, if there is an ADL and a STRIPS formulation
of a version, then each instance is counted only once.\footnote{As
  said, each domain featured several domain {\em versions}, differing
  in terms of the application constraints modelled. Within each domain
  version, there could be several different domain version {\em
    formulations}, differing in terms of the precise PDDL subset used
  to encode the {\em same} semantics. In most cases with more than one
  formulation, there were exactly two formulations, using/not using
  ADL constructs, respectively.  Competitors could choose the
  formulation they liked best and the results across different
  formulations were evaluated together.}

The first line in Table~\ref{tab:results-overview}, indexed
``Number'', simply gives the size of the instance set associated with
the column. All other lines correspond, obviously, to planning
systems. The upper half of the table contains, in alphabetical order,
the satisficing planners; the lower half contains the optimal
planners.  Note that we list only one version of LPG-TD.  This is
because the ``speed'' and ``quality'' version of this planner differ
in terms of plan quality, but not in terms of the number of
attacked/solved instances.

Let us consider the table entries in between the leftmost and
rightmost columns, i.e., the entries concerning a planner ``X'' and a
language/instance subset ``Y''. The numbers in these entries are the
{\em success ratio} and the {\em attacked ratio}. Precisely, they were
obtained as follows. We first counted the number ``y'' of instances in
``Y'' that planner ``X'' tried to solve -- our definition was to take
all domain versions (inside ``Y'') for which ``X'' delivered results,
and set ``y'' to be the total number of instances in these domain
versions. We then counted the number ``x'' of instances in ``Y'' that
planner ``X'' succeeded to solve. We obtained the first number in the
table entry -- the success ratio -- as the ratio (in percent) of ``x''
divided by ``y''. We obtained the second number in the entry -- the
attacked ratio -- as the ratio (in percent) of ``y'' divided by the
size of ``Y''. For space reasons, we rounded the percentages to the
values shown in the table. A dash in the table entry means that the
planner ``X'' can not handle the language subset ``Y''. An empty table
entry means that the planner can handle ``Y'', but did not attack any
instance in ``Y''.

In a few cases, namely in the Promela domain, see the discussion in
Section~\ref{results:promela}, we could not formulate the largest
instances in STRIPS because these (fully-grounded) representations
became too large. So, there, the numbers of test instances are
different between the ADL and STRIPS formulations {\em of the same
  domain version}.\footnote{In PSR, it also was impossible to
  formulate the largest instances in STRIPS.  But, there, we split the
  domain into different {\em versions} regarding the size and
  formulation language of the instances. We did not want to introduce
  more distinction lines in Promela because there we already had 8
  different domain versions.}  Table~\ref{tab:results-overview} uses
the number of ADL instances, not taking account of this subtlety. This
implies a slight disadvantage in terms of success ratio in columns
``N'' and ``All'', for the planners that attacked the smaller STRIPS
test suites. These planners are the following, with correct column
``N'' success ratio in parentheses: CRIKEY (51), FAP (41), LPG-TD
(74), SGPlan (80), YAHSP (92), BFHSP (35), CPT (39), HSP$_a^*$ (52),
Optiplan (41), SATPLAN (55), and TP4 (37).

The table entries in the rightmost column are obtained similarly as
above. They summarize the situation regarding the respective planners
across {\em all} used language subsets. The left and right number in
each entry give the ratio of the number of solved instances to the
number of attacked instances, and the number of attacked instances to
the number of all instances, respectively. The number in the middle of
each entry gives the ratio of the number of attacked instances to the
number of all instances {\em that lie within the language range of the
  respective planner.} We included this number in order to provide a
measure of to what extent each planner/team attacked those instances
that they could attack.

Some remarks are in order regarding the latter ratio, and the ratio of
attacked instances in general. First, there was no rule telling the
competitors to ``attack everything they can'' -- competitors were free
to choose. In retrospect, we feel that there should have been such a
rule, since it would make data interpretation easier. The way things
are, it is not known for what reason a planner did not attack an
instance subset (domain version): Bad results? Not interested?
Overlooked? Second, many planners can handle only subsets of
certain language features, like, of numeric variables (``NV''). This
can lead to too low percentages in Table~\ref{tab:results-overview},
where for simplicity we do not take account of these details. Third, 
one detail we {\em did} take account of is
that 50 of the 382 instances in column ``N'' (namely, the
``middle-compiled'' version of the PSR domain, see
Section~\ref{results:psr}) are formulated in ADL only, and are thus
not accessible to many of the planners. For planners not able to
handle ADL, when computing the middle number in the rightmost table
entry, we subtracted these 50 instances from the ``instances within
the language range'' of the respective planner. In column ``N'',
however, we took the usual ratio to the set of {\em all} instances --
in particular, this is why Macro-FF, YAHSP, BFHSP, CPT, Optiplan, and
SATPLAN all have an 87\% attacked-ratio in the ``N'' column, but a
100\% attacked-ratio as the middle number of the rightmost column.

All in all, the planners all have a pretty good coverage of the
language subset they could attack; except Optop, Tilsapa, and maybe
Semsyn. These planners/teams were probably interested only in a small
subset of the instances.\footnote{Drew McDermott, the developer of
  Optop, told us in private conversation that he tried to solve only
  the most complicated domain versions.} In the other cases where
instances were left out, planners/teams typically did not attack the
domain versions where derived predicates or timed initial literals had
been compiled away. Note that a lack of coverage may also be due to
language details not accounted for by the rather crude distinctions
made for Table~\ref{tab:results-overview}.\footnote{In private
  conversation, the LPG-TD team told us that this is the case for
  LPG-TD, and that they in fact attacked all domain versions they
  could at the time IPC-4 was running.}

% \footnote{In private conversation, the LPG-TD team
%   told us that they were actually integrating the new language
%   capabilities into their system on-line, i.e., while IPC-4 was
%   running. For some domain versions, they did not manage to implement
%   the needed capabilities in time. They did attack every domain
%   version they could. So in that sense the middle number in the
%   rightmost column should be $100$, not $71$, for LPG-TD.}

What can we conclude from Table~\ref{tab:results-overview}, in terms
of planner performance? Some spotlights are these:
\begin{itemize}
\item The success ratios of optimal planners are generally lower than
  those of satisficing planners. The largest overall success ratio of
  an optimal planner, SATPLAN, is 46 (55 when taking account of the
  above mentioned subtlety regarding ADL/STRIPS formulations in
  Promela); compared to 88 for FDD.
\item However, there are also various cases where an optimal planner
  has a higher success ratio than a satisficing one.
\item FD, FDD, and YAHSP have the best success ratios (as mentioned,
  the success ratio of YAHSP in the ``N'' column is 92 when taking
  account of ADL/STRIPS in Promela). FD and FDD also have a pretty
  good coverage in the ``N'' and ``N+DP'' columns, i.e. within PDDL2.2
  level 1 (the left out instances are the domain versions with
  compiled derived predicates).
\item SGPlan has an extremely high coverage, attacking 96\% of {\em
    all} instances. Even so, it has a very competitive success ratio.
  (Concretely, SGPlan solved 1,090 of the 1,556 IPC-4 instances; it
  attacked 1,496. The second place in terms of the %sheer
  number of instances solved is held by LPG-TD, which solved 843 out
  of the 1108 instances it could attack.)
\item LPG-3, i.e. the IPC-3 version of LPG, ranges somewhere in the
  middle-ground of the planners, in terms of success ratio and
  coverage. This indicates that significant performance improvements
  were made in difference to IPC-3. (Note that this holds for the new
  LPG version just as for the other top performance planners.)
\end{itemize}

Naturally, in our evaluation of the results, particularly in the
evaluation that formed the basis of the award decisions, we undertook
a more detailed examination of the data set. We considered, as much as
possible, {\em scaling behavior}, rather than the simple instance
counts above. In what follows, we first explain our evaluation process
in detail, then we consider each of the domains in turn.

\subsection{Results Evaluation}
\label{results:evaluation}

As said, runtime data for satisficing and optimal planners was
considered separately. For plan quality, we distinguished between
three kinds of optimization criteria: number of actions, makespan
(equalling the number of time steps in the non-temporal domains), and
metric value.  The last of these three was only defined if there was a
{\tt :metric} specification in the respective task. We compared each
planner according to only a single plan quality criterion. That is,
for every domain version the competitors told us what criterion their
planner was trying to optimize in that domain version. We evaluated
all (and only) those planners together that tried to optimize the same
criterion. Our reason for doing so is that it does not make sense to
evaluate planners based on criteria they don't actually consider.

Altogether, for every domain version we created up to 5 \emph{gnuplot}
graphics: for satisficing runtime, for optimal runtime, for number of
actions, for makespan, and for metric value. The plan quality data of
optimal planners was put into the same plots as those for satisficing
planners, to enable some comparison. In a few cases, we had to split a
runtime figure (typically, for satisficing planners) into two separate
figures because with too many runtime curves in it the graphic became
unreadable.

Based on the \emph{gnuplot} graphics, we evaluated the data in terms
of asymptotic runtime and solution quality performance. The
comparisons between the planners were made by hand, i.e. by looking at
the graphs.  While this is simplistic, we believe that it is an
adequate way of evaluating the data, given the goals of the field, and
the demands of the event. It should be agreed that what we are
interested in is the (typical) scaling behavior of planners in the
specific domains.  This excludes ``more formal'' primitive comparative
performance measures such as counts of instances solved more
efficiently -- a planner may scale worse than another planner, yet be
faster in a lot of smaller instances just due to pre-processing
implementation details.  The ideal formal measure of performance would
be to approximate the actual scaling functions underlying the
planners' data points. But it is infeasible to generate enough data,
in an event like the IPC, to do such formal approximations. So we
tried to judge the scaling behaviors by hand, laying most emphasis on
how efficiently/if at all the largest instances in a test suite were
solved by a planner.\footnote{Note that the success ratios displayed
  in Table~\ref{tab:results-overview} are also a crude way to access
  scaling behavior.}

The rest of the section contains one sub-section for each domain in
turn. Each sub-section provides the following material. First, we give
a brief description of the domain and its most important features
(such as domain versions and formulations used in IPC-4). We then
include a set of gnuplot graphics containing (what we think are) the
most important observations regarding runtime (the total set of
gnuplot graphs is too space-consuming). We discuss plan quality, with
data comparing the relative performance of pairs of planners.  We also
add some intuitions regarding structural properties of the domains,
and their possible influence on the performance of the planners.
Finally, we provide the information -- 1st and 2nd places, see below
-- underlying the decisions about the awards. This information is
contained in a text paragraph at the end of the sub-section for each
domain. %, so the un-interested reader can skip it over.

As the basis for deciding about the awards, within every domain
version, we identified a group of planners that scaled best and
roughly similar. These planners were counted as having a 1st place in
that domain version. Similarly, we also identified groups of 2nd place
planners. The awarding of prizes was then simply based on the number
of 1st and 2nd places of a planner, see
Section~\ref{awards}.\footnote{Of course, many of the decisions about
  1st and 2nd places were very close; there were numerous special
  cases due to, e.g., what planners participated in what domain
  versions; summing up the places introduces a dependency of the
  results on the number of domain versions in the individual domains.
  To hand out awards one has to make decisions, and in these decisions
  a lot of detail is bound to disappear. One cannot summarize the
  results of such a huge event with just a few prizes.}

We consider the domains in alphabetical order. Before we start, there
are a few more remarks to be made.  First, some planners are left out
of the plots in order to make the plots readable.  Specifically, we
left out LPG-TD.quality as well as LPG-3.quality (the IPC-3 version of
LPG with a preference on quality); these planners are always slower
than their counterparts with a preference on speed. We also left out
FD, since in most cases it showed very similar behavior to FDD. We
chose FDD because in a few cases its performance is superior. Note
further that we do not distinguish the runtime performance of optimal
planners optimizing different criteria.  Finally, it can make
a significant difference whether a planner is run on an ADL encoding,
or on its compilation to STRIPS; this distinction also gets lost in
our evaluation. We emphasize that we applied the simplifications only
in order to improve readability and understandability of the results.
We do not wish to imply that the planners/distinctions that are
omitted aren't important.

\subsection{Airport}
\label{results:airport}

In the Airport domain, the planner has to control the ground traffic
on an airport. The task is to find a plan that solves a specific
traffic situation, specifying inbound and outbound planes along with
their current and goal positions on the airport. The planes must not
endanger each other, i.e. they must not both occupy the same airport
``segment'' (a smallest road unit), and if plane $x$ drives behind
plane $y$ then between $x$ and $y$ there must be a safety distance
(depending on the size of $y$). These safety constraints are modelled
in terms of ``blocked'' and ``occupied'' predicates, whose value is
updated and controlled via complex ADL preconditions and conditional
effects.

There were four different versions of the domain in IPC-4: a
non-temporal version, a temporal version, a temporal version with time
windows, and a temporal version with compiled time windows. The time
windows in the latter two versions concern airplanes that are known to
land in the future, and that will thus block certain airport segments
(runways) during certain time windows. The time windows are modelled
using timed initial literals, respectively their compilation. In every
domain version, there is an ADL formulation, as well as a STRIPS
formulation obtained by compiling the ADL constructs away (resulting
in a partially grounded encoding).

Instances scale in terms of the size of the underlying airport, as
well as the number of airplanes that must be moved. One remarkable
thing about the Airport domain is that the instances were generated
based on a professional airport simulation tool. The largest instances
used in IPC-4 (number 36 to 50) correspond to a real airport, namely
Munich airport (MUC). Instance number 50 encodes a traffic situation
with 15 moving airplanes, which is typical of the situations
encountered at MUC in reality. Instances number 1 to 20 come from
smaller toy airports, instances number 21 to 35 are based on one half
of MUC airport.

\begin{figure}[t]
\begin{center}
\vspace{-0.0cm}
\begin{tabular}{cc}
\includegraphics[width=7.3cm]{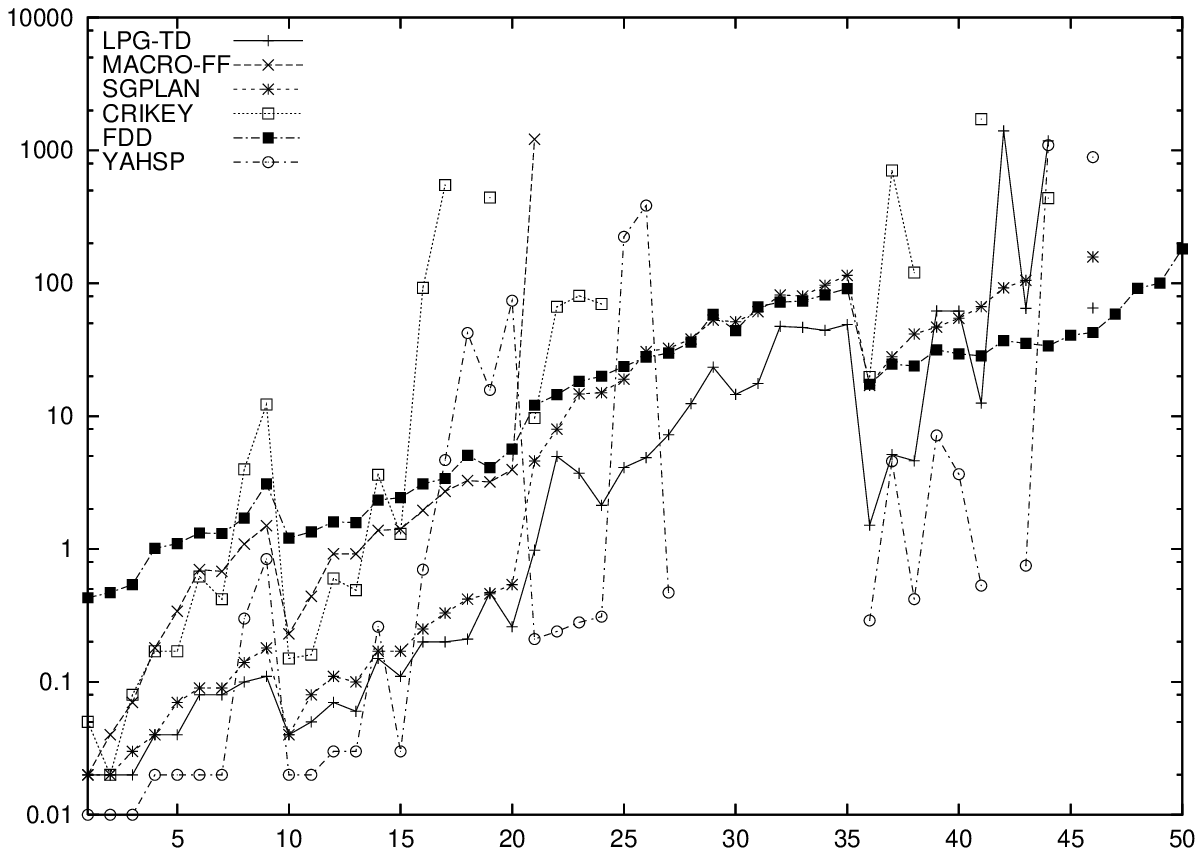} &
\includegraphics[width=7.3cm]{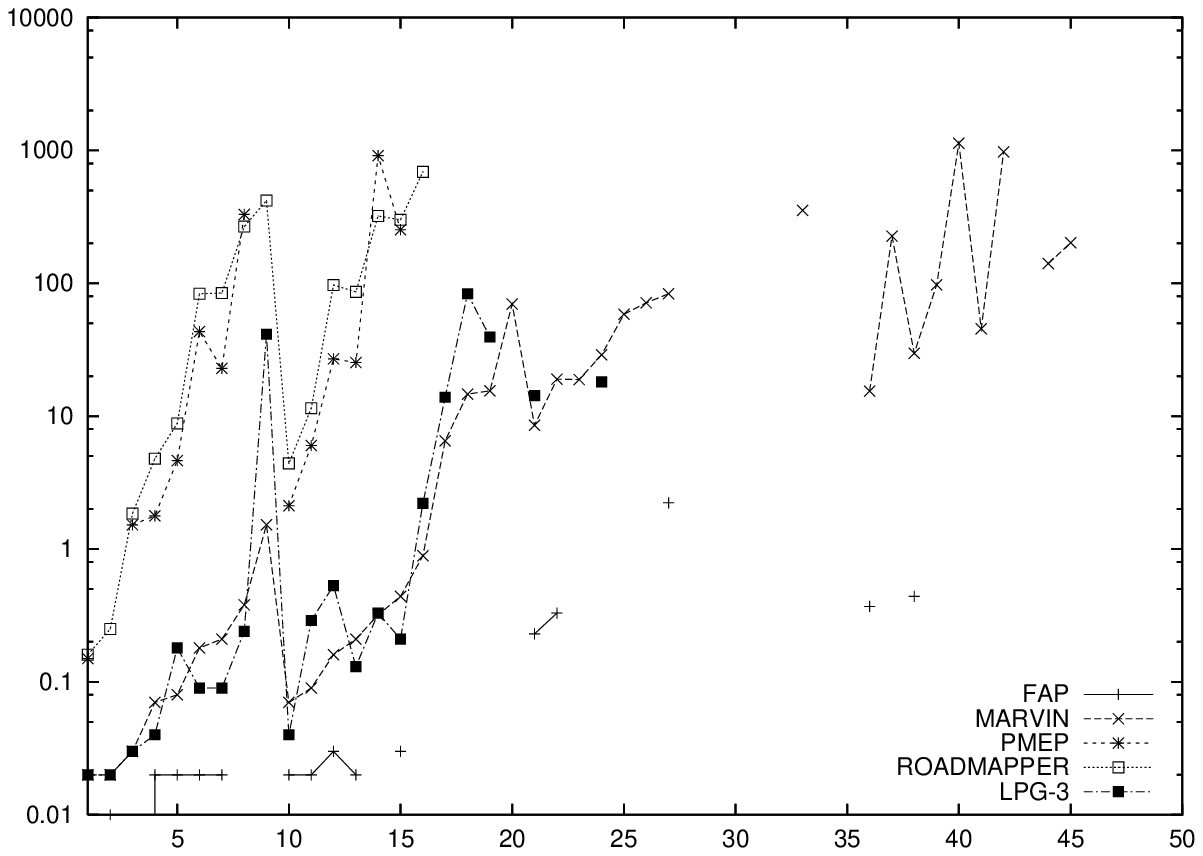}\\
(a) & (b)
\end{tabular}
\includegraphics[width=7.3cm]{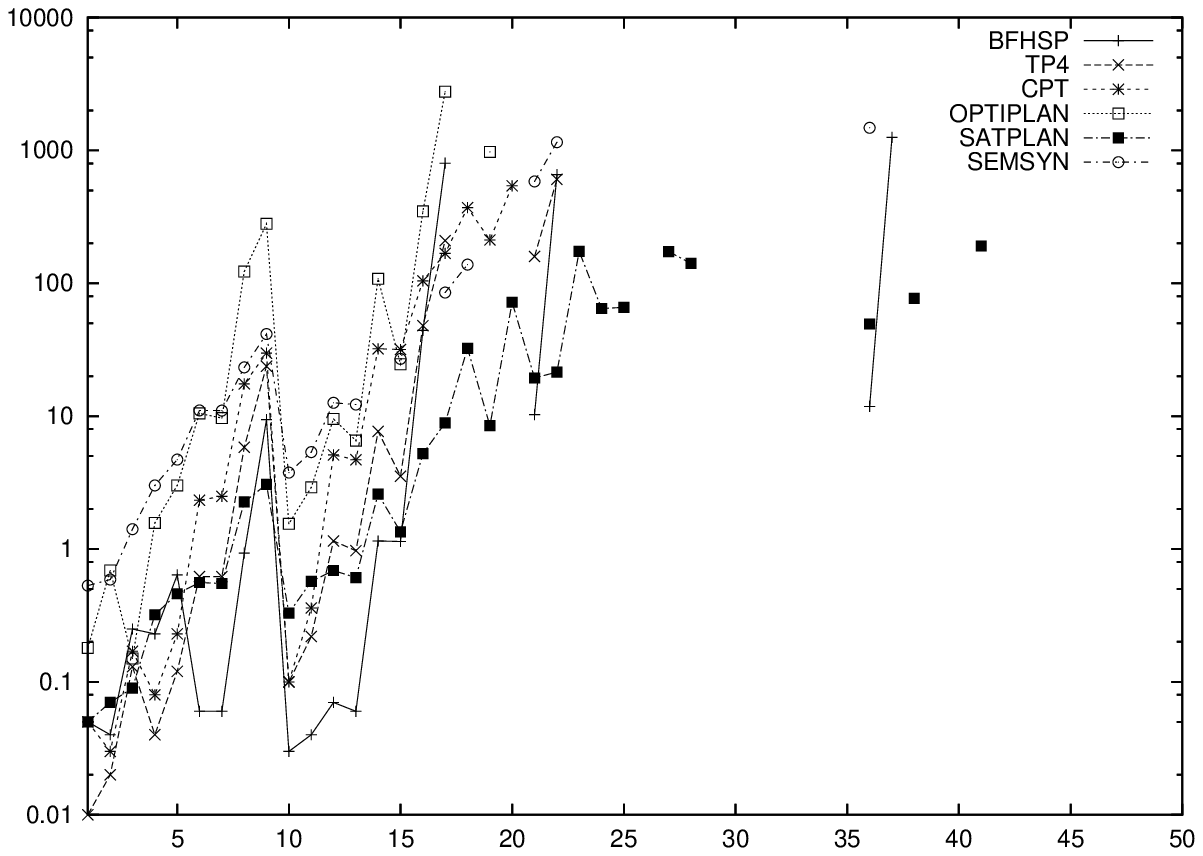}\\
(c)
\vspace{-0.0cm}
\caption{Performance in non-temporal Airport, satisficing (a) and (b), optimal (c).}
\label{airport:nontemporal}
\vspace{-0.0cm}
\end{center}
\end{figure}

Figure~\ref{airport:nontemporal} shows the runtime performance in the
non-temporal version of the domain. For readability, the set of satisficing
planners is split over two graphs. As will be the case in all graphs
displayed in the subsequent discussions, the $x$-axis denotes the
instance number (obviously, the higher the number the larger the
instance), and the $y$-axis gives CPU runtime in seconds on a
logarithmic scale. CPU time is total, including parsing and any form
of static pre-processing.

It can be observed in Figure~\ref{airport:nontemporal} that FDD is the
only planner solving all problem instances. LPG-TD and SGPlan both
scale relatively well, but fail on the largest instances. The other
planners all behave much more unreliably. Observe also that the IPC-3
version of LPG lags far behind FDD, LPG-TD, and SGPlan. For the
optimal planners, which unsurprisingly behave clearly worse than the
satisficing planners, the only clear-cut observation is a performance
advantage for SATPLAN. The other optimal planners all behave
relatively similarly; Semsyn and BFHSP are the only ones out of that
group solving a few of the larger instances.

For plan quality, there are two groups of planners, one trying to
minimize the number of actions, and one trying to minimize makespan,
i.e. the number of parallel action {\em steps}. In the former group,
the plan quality performance differences are moderate. CRIKEY, LPG-TD,
SGPlan, and YAHSP sometimes find sub-optimal plans. FDD's plans are
optimal in all cases where an optimal planner found a solution. As a
measure of comparative plan quality, from now on we provide, given
planners A and B, the min, mean, and maximum of the ratio quality(A)
divided by quality(B), for all instances that are solved by both A and
B. We call this data the ``ratio A vs B''. For CRIKEY (A) vs FDD (B),
the ratio is $0.91$ (min), $1.04$ (mean), and $1.45$ (max), $[0.91
(1.04) 1.45]$ for short. For LPG-TD.speed vs FDD the ratio is $[0.91
(1.08) 1.80]$; for LPG-TD.quality vs FDD it is $[0.91 (1.06) 1.80]$.
For SGPlan vs FDD it is $[0.91 (1.07) 2.08]$; for YAHSP vs FDD it is
$[0.91 (1.07) 1.43]$.

In the group of planners minimizing makespan, (only) Marvin has a
tendency to find very long plans. The comparison Marvin vs the optimal
SATPLAN is $[1.00 (2.46) 4.64]$. In the maximum case, SATPLAN finds a
plan with 53 steps, and Marvin's plan is 246 steps long.

An interesting observation concerns the two scaling parameters of that
domain: the number of airplanes, and the size of the airport. In all
the plots in Figure~\ref{airport:nontemporal}, and also in most plots
in Figure~\ref{airport:temporal} below, one can observe that the
instances number 26 to 35 become increasingly hard for the planners --
but in the step to instance number 36, the performance suddenly
becomes better again. As said above, the instances 21 to 35 are based
on one half of MUC airport, while the instances 36 to 50 are based on
the full MUC airport. But between instances 26 and 35, the number of
airplanes rises from 6 to 12; instance 36 contains only 2 airplanes,
instances 37 and 38 contain 3.  That is, it is easier for the planners
to address a larger airport with fewer planes. Note here the domain
combinatorics: the number of reachable states is in the order of
$S^n$, where $S$ is the number of different airport segments, and $n$
is the number of airplanes.

\begin{figure}[htb]
\begin{center}
\vspace{-0.0cm}
\begin{tabular}{cc}
\includegraphics[width=7.3cm]{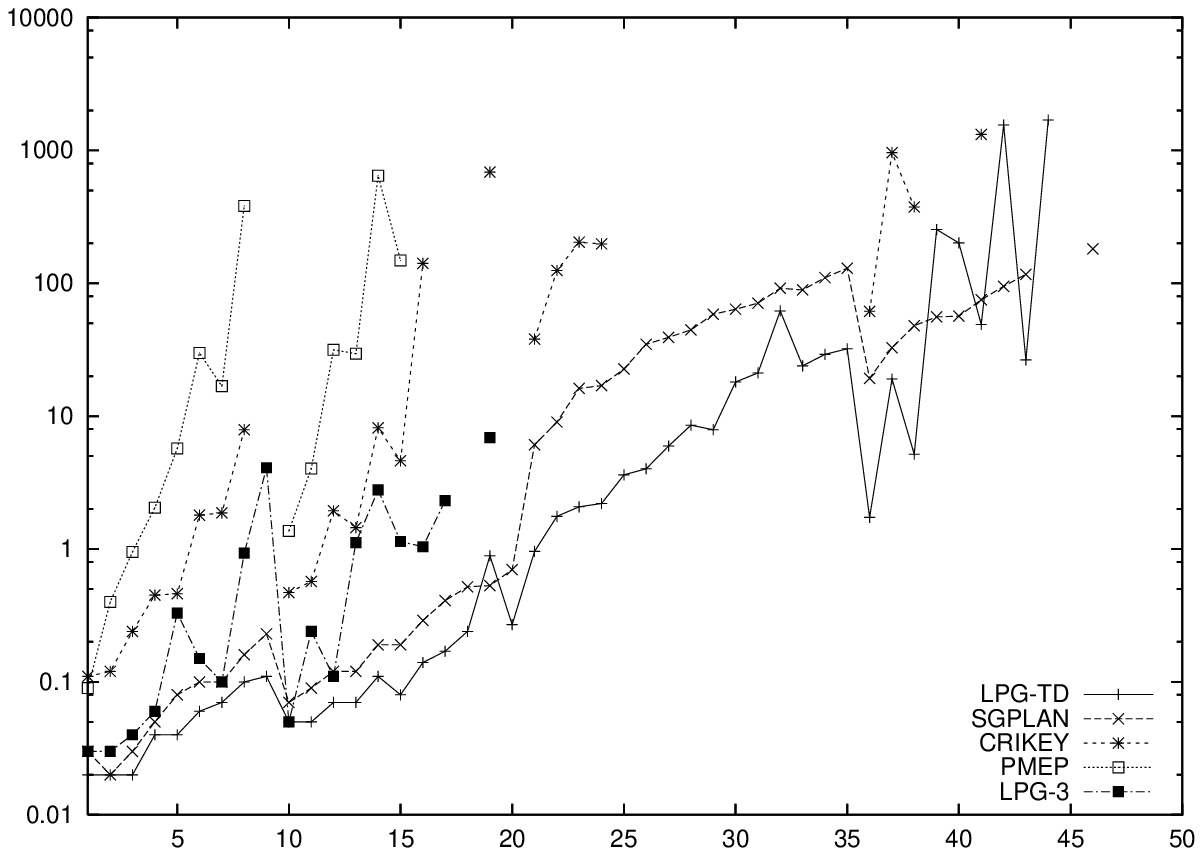} &
\includegraphics[width=7.3cm]{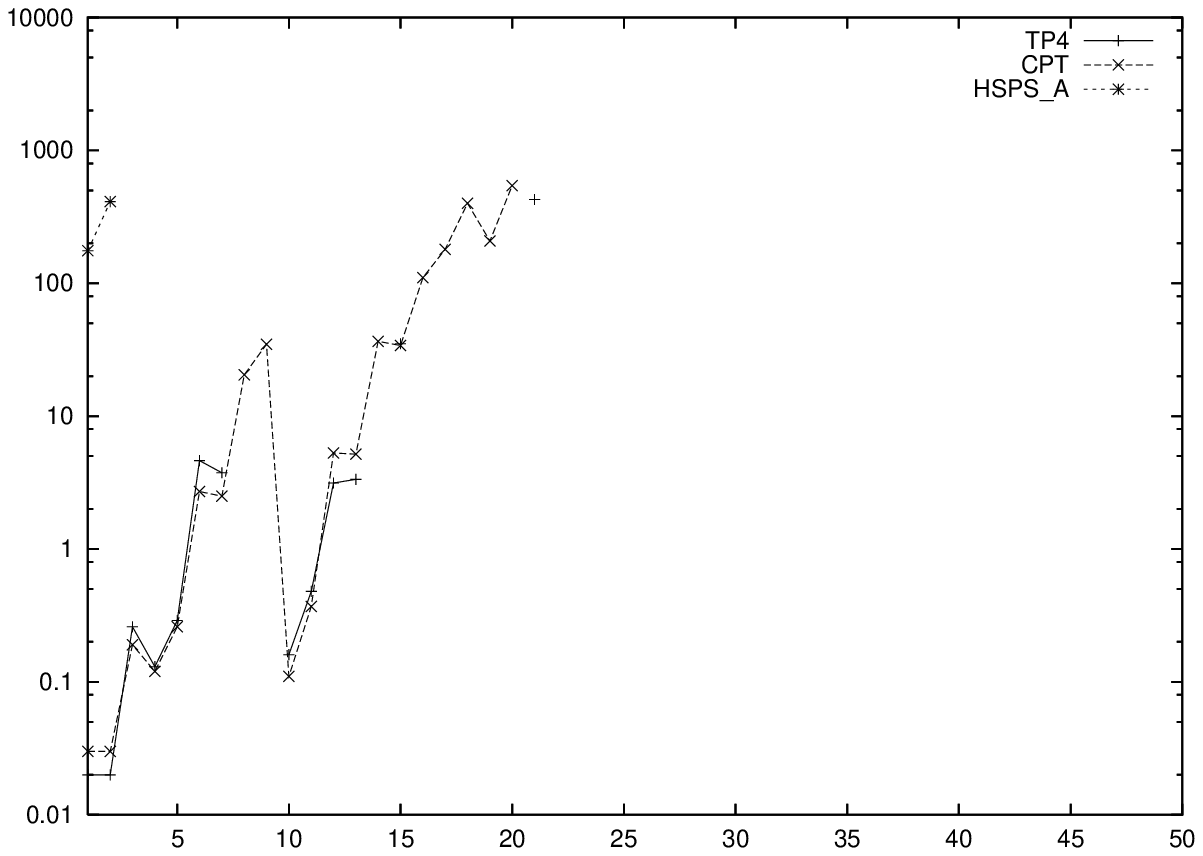}\\
(a) & (b)\\
\includegraphics[width=7.3cm]{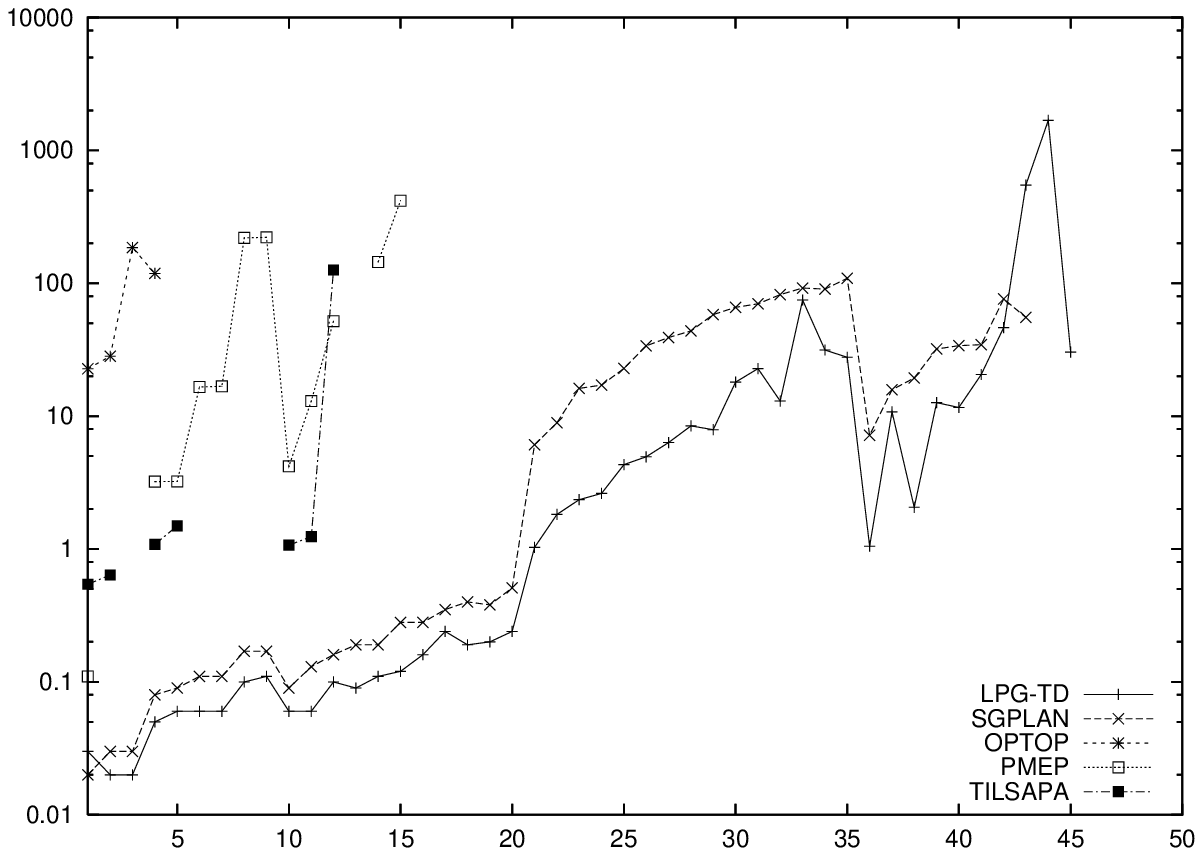} &
\includegraphics[width=7.3cm]{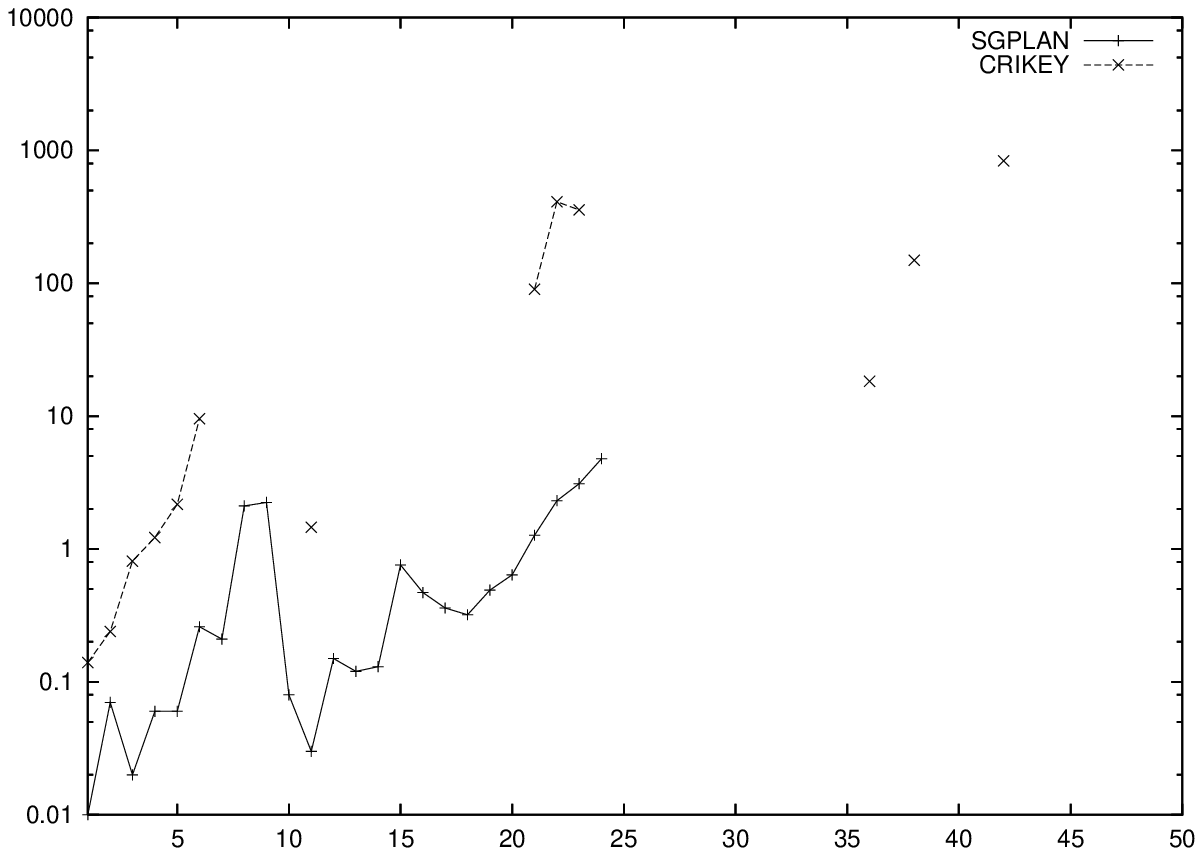}\\
(c) & (d)
\end{tabular}
\vspace{-0.0cm}
\caption{Performance in temporal Airport, satisficing (a) and optimal (b). In temporal Airport
  with time windows, satisficing, explicit encoding (c) and compiled
  encoding (d).}
\label{airport:temporal}
\vspace{0.0cm}
\end{center}
\end{figure}

Figure~\ref{airport:temporal} shows the runtime performance in the
temporal versions of the domain. We first consider the domain versions
without time windows: parts (a) and (b) of the figure. Of the
satisficing planners, LPG-TD and SGPlan both scale just as well as
they do in the non-temporal case; all other satisficing planners scale
orders of magnitude worse. Of the optimal planners, CPT scales best,
followed by TP4. The two planners minimizing the number of actions are
SGPlan and CRIKEY; SGPlan behaves somewhat worse, the ratio is $[0.95
(1.03) 1.54]$. Of the planners that minimize makespan, only LPG-TD
scales beyond the smallest instances. The ratio LPG-TD.speed vs the
optimal CPT is $[1.00 (1.15) 1.70]$, P-MEP vs CPT is $[1.00 (1.11)
1.42]$. The ratio LPG-TD.quality vs CPT is $[1.00 (1.03) 1.44]$.

No optimal planner in the competition could handle timed initial
literals, so with explicit time windows no such planner participated.
With compiled time windows, only CPT participated, scaling a little
worse than in the temporal domain version without time windows. For
the satisficing planners, the results are more interesting. With
explicit time windows, again LPG-TD and SGPlan maintain their good
scaling behavior from the simpler domain versions. To the other
planners, there is a huge runtime performance gap. SGPlan is the only
planner here that minimizes the number of actions. The makespan ratio
LPG-TD.speed vs LPG-TD.quality is $[1.00 (1.15) 2.01]$. With compiled time
windows, only SGPlan and CRIKEY participated. The former consistently
solves the smaller instances up to a certain size, the latter fails on
many of the small instances but can solve a few of the larger ones.
Neither of the planners shows reasonable runtime performance compared
to the explicit encoding of the domain. The number of actions ratio
CRIKEY vs SGPlan is $[1.00 (1.14) 1.36]$.

Our main motivation for including the Airport domain in IPC-4 was that
we were able to model it quite realistically, and to generate quite
realistic test instances -- the only thing we could not model was the
real optimization criterion (which asks to minimize the summed up
travel time of all airplanes). So the good scaling results of, at
least, the satisficing planners, are encouraging. Note, however, that
in reality the optimization criterion is of crucial importance, and
really the only reason for using a computer to control the traffic.

We remark that the domain is not overly difficult from the perspective
of relaxed-plan based heuristic planners (using the wide-spread
``ignore deletes'' relaxation). \citeA{hoffmann:jair-05} shows that
Airport instances can contain unrecognized dead ends (states from
which the goal can't be reached, but from which there is a relaxed
plan); but such states are not likely to occur very often. A dead end
in Airport can only occur when two planes block each other, trying to
get across the same segment in different directions. Due to the
topology of the airports (with one-way roads), this can happen only in
densely populated parking regions.  In case for any two airplanes the
respective paths to the goal position are disjoint, the length of a
relaxed plan provides the {\em exact} distance to the nearest goal
state. In that sense, the good runtime performance of the satisficing
planners didn't come entirely unexpected to us, though we did not
expect them to behave {\em that} well. Note that most of these
planners, particularly FDD, SGPlan, and LPG-TD, do much more than/do
things different from a standard heuristic search with goal distances
estimated by relaxed plan length.  Note also that
\citeA{hoffmann:jair-05}'s results are specific to {\em non-temporal}
domains.

In the non-temporal domain version, we awarded 1st places to FDD (and
FD) and SATPLAN; we awarded 2nd places to LPG-TD, SGPlan, Semsyn,
and BFHSP. In the temporal version, we awarded 1st places to LPG-TD,
SGPlan, and CPT; a 2nd place was awarded to TP4. In the domain version
with explicit time windows, 1st places were awarded to LPG-TD and
SGPlan.

\subsection{Pipesworld}
\label{results:pipesworld}

The Pipesworld domain is a PDDL adaptation of an application domain
dealing with complex problems that arise when transporting oil
derivative products through a pipeline system. Note that, while there
are many planning benchmarks dealing with variants of transportation
problems, transporting oil derivatives through a pipeline system has a
very different and characteristic kind of structure. The pipelines
must be filled with liquid at all times, and if you push something
into the pipe at one end, something possibly completely different
comes out of it at the other end. As a result, the domain exhibits
some interesting planning space characteristics and good plans
sometimes require very tricky maneuvers. Additional difficulties that
have to be dealt with are, for example, interface restrictions
(different types of products interfere with each other in a pipe),
tankage restrictions in areas (i.e., limited storage capacity defined
for each product in the places that the pipe segments connect), and
deadlines on the arrival time of products.

In the form of the domain used in IPC-4, the product amounts dealt
with are discrete in the sense that we assume a smallest product unit,
called ``batch''. There were six different domain versions in IPC-4:
notankage-nontemporal, notankage-temporal, tankage-nontemporal,
tankage-temporal, notankage-temporal-deadlines, and
notankage-temporal-deadlines-compiled. All versions include interface
restrictions. The versions with ``tankage'' in their name include
tankage restrictions, modelled by a number of ``tank slots'' in each
place in the network, where each slot can hold one batch (note that
the slots introduce some additional symmetry into the problem; we get
back to this below). In the versions with ``temporal'' in their name,
actions take (different amounts of) time.  The versions with
``deadlines'' in their name include deadlines on the arrival of the
goal batches, modelled by timed initial literals respectively their
compilation to standard PDDL2.1. None of the encodings uses any ADL
constructs, so of each domain version there is just one (STRIPS)
formulation.

The IPC-4 example instances were generated based on five scaling
network topologies. The smallest network topology has 3 areas
connected by 2 pipes; the largest network topology has 5 places
connected by 5 pipes. For each network we generated 10 instances, with
growing numbers of batches, and of batches with a goal location. So
altogether we had 50 instances per domain version, numbered
consecutively. Between domain versions, the instances were, as much as
possible, transferred by adding/removing constructs. E.g., the
instances in tankage-nontemporal (tankage-temporal) are exactly the
same as those in notankage-nontemporal (notankage-temporal) except
that tankage restrictions were added.

We do not include graphs for the domain versions featuring deadlines.
In the version with explicit deadlines (modelled by timed initial
literals), only LPG-TD and Tilsapa participated, of which LPG-TD
scaled up to middle-size instances; Tilsapa solved only a few of the
smallest instances. In the domain version with compiled deadlines,
only CPT participated, solving a few small instances.

\begin{figure}[t]
\begin{center}
\vspace{-0.0cm}
\begin{tabular}{cc}
\includegraphics[width=7.3cm]{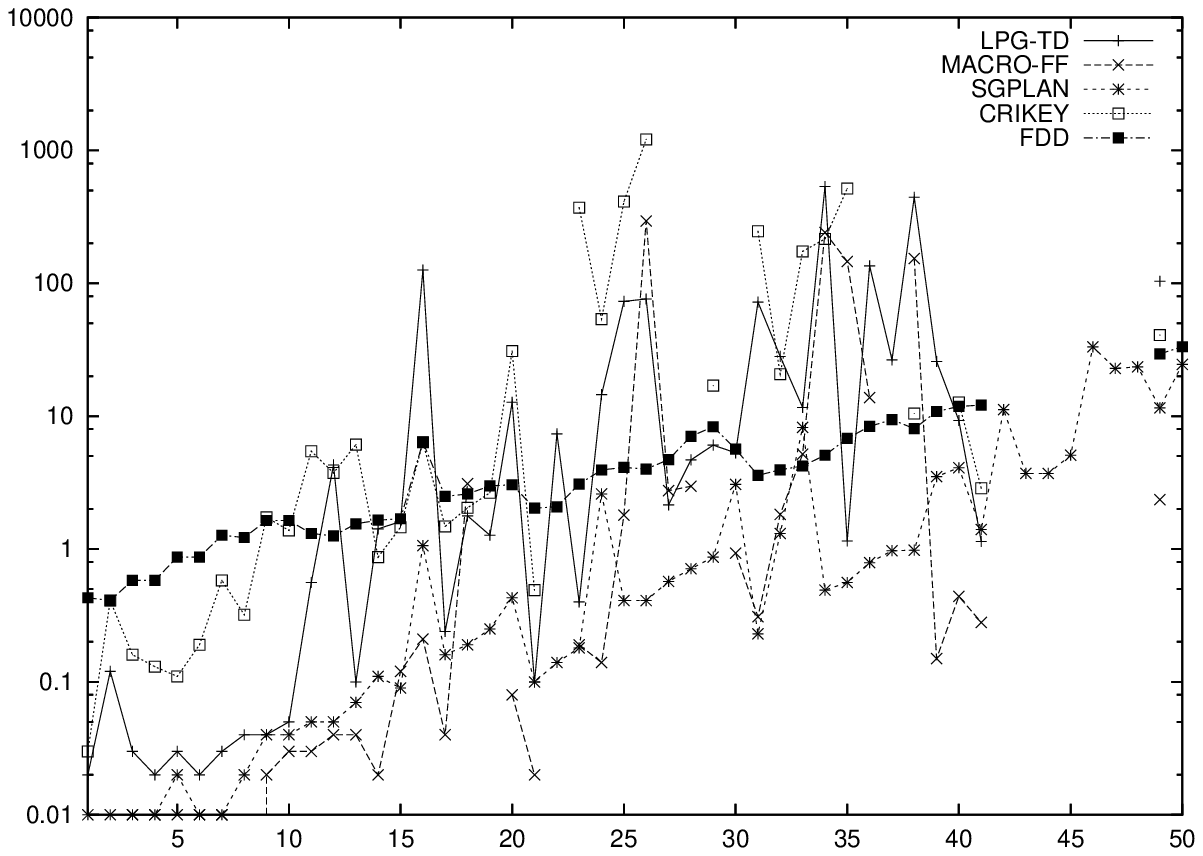} &
\includegraphics[width=7.3cm]{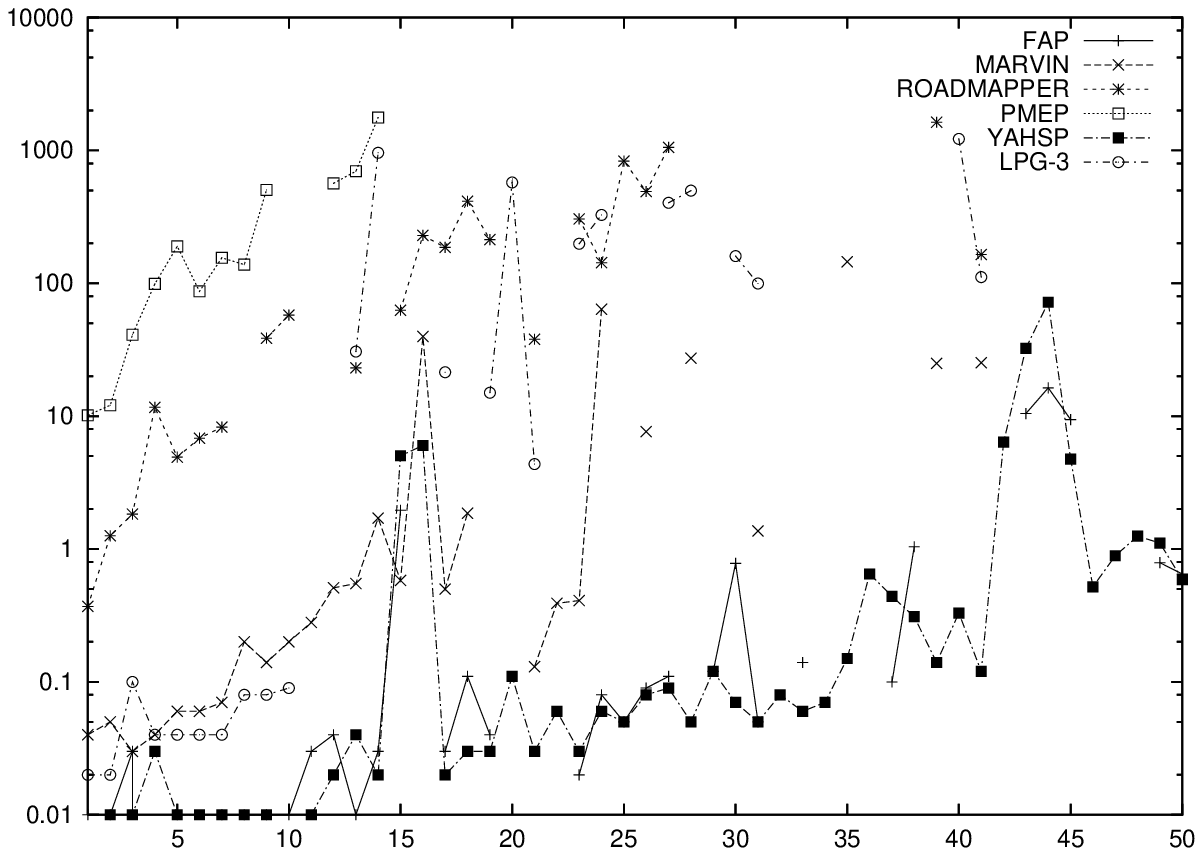}\\
(a) & (b)
\end{tabular}
\includegraphics[width=7.3cm]{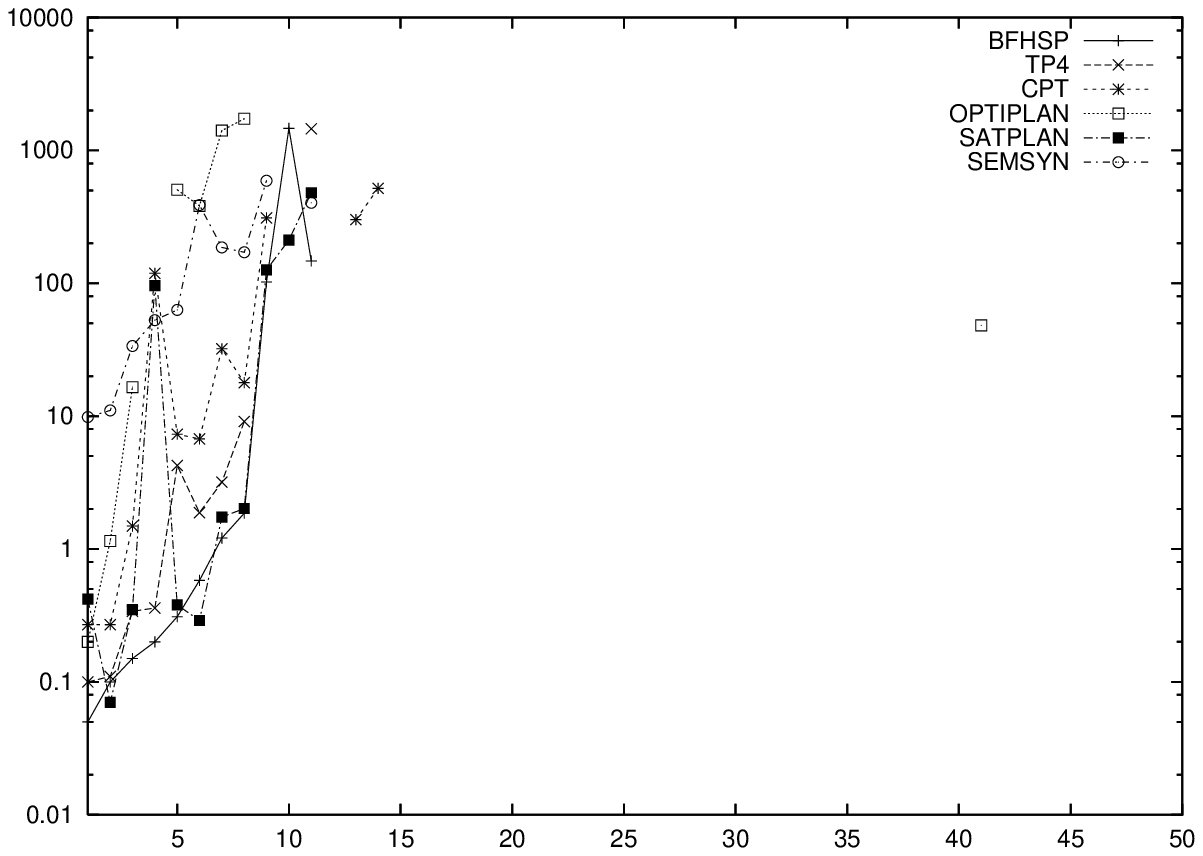}\\
(c)
\vspace{0.0cm}
\caption{Non-temporal Pipesworld, no tankage constraints, satisficing (a) and (b), 
  optimal (c).}
\label{pipesworld:nontankagenontemp}
\vspace{0.0cm}
\end{center}
\end{figure}

Figure~\ref{pipesworld:nontankagenontemp} shows the results for the
non-temporal domain version without tankage restrictions. Parts (a)
and (b) contain the results for satisficing planners in the respective
non-temporal domain version. We observe that YAHSP and SGPlan are the
only planners that can solve all instances. YAHSP is a lot faster in
most instances; but it finds excessively long plans: the ratio YAHSP
vs SGPlan is $[0.38 (1.77) 14.04]$, where in the maximum case SGPlan needs
72 actions, YAHSP 1,011. For the other planners, plan quality does not
vary much, with the exception of Marvin, whose makespan ratio vs the
optimal CPT is $[0.88 (1.60) 2.19]$.  LPG-TD and FDD both solve all but a
few of the largest instances. Note that, as in Airport, the IPC-3
version of LPG is outperformed dramatically by the best IPC-4
planners.

Part (c) of Figure~\ref{pipesworld:nontankagenontemp} shows the
results for the optimal planners.  The runtime curves are extremely
similar, nearing indistinguishable.  Note that Optiplan is the only
optimal planner that solves an instance with a very large size
parameter. However, in Pipesworld, like in Airport, due to the domain
combinatorics, planners are likely to find it easier to solve a large
network with little traffic, than a small network with a lot of
traffic.\footnote{On the other hand, note at this point that the
  networks in Pipesworld do by far not grow as much as in Airport,
  where they grow from microscopic toy airports for the smallest
  instances to a real-world airport for the largest instances. As
  said, the smallest network in Pipesworld has 3 areas connected by 2
  pipes, and the largest network has 5 places connected by 5 pipes.}
The other optimal planners here may just not have tried to run their
planner on larger instances when it already failed on the smaller ones
-- we explicitly advised people to save machine workload by not
insisting on spending half an hour runtime on instances that were
probably infeasible for their planners anyway. In cases like this one
here, this admittedly is a potentially misleading guideline.

\begin{figure}[t]
\begin{center}
\vspace{-0.0cm}
\begin{tabular}{cc}
\includegraphics[width=7.3cm]{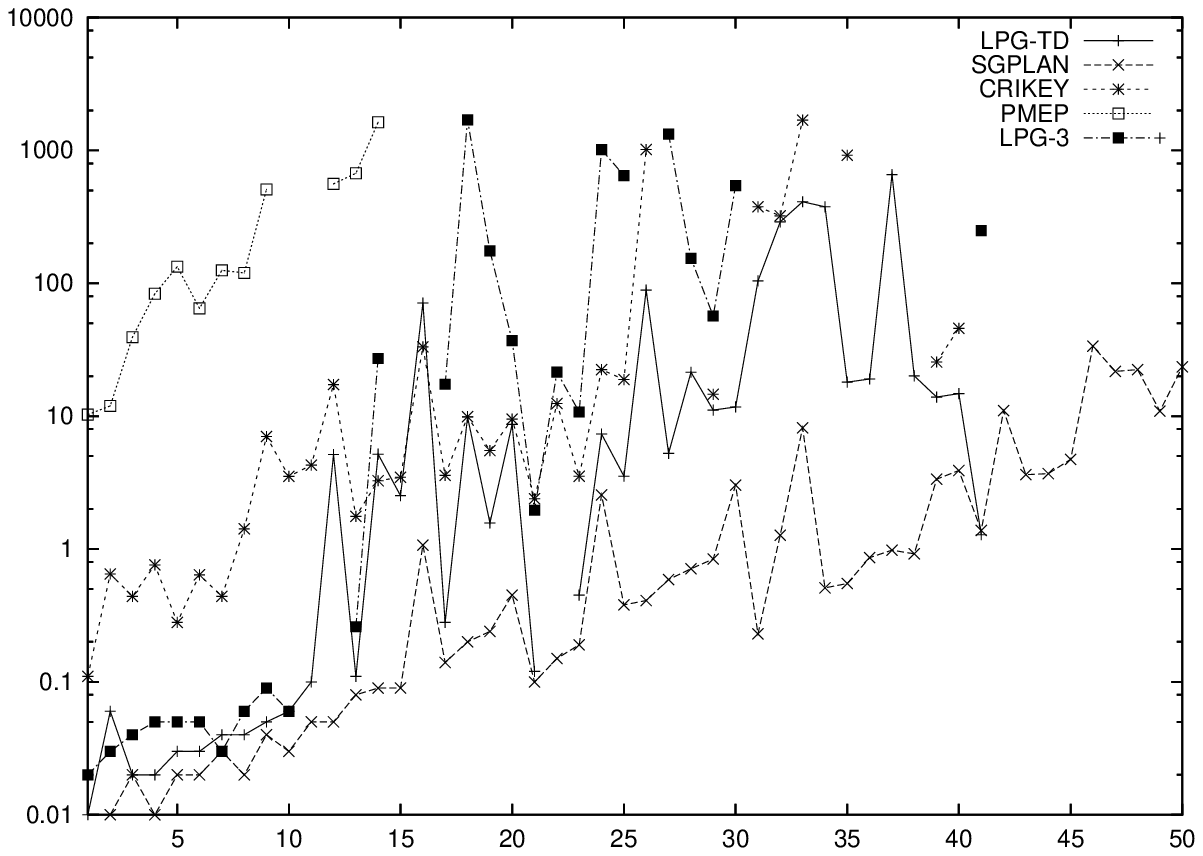} &
\includegraphics[width=7.3cm]{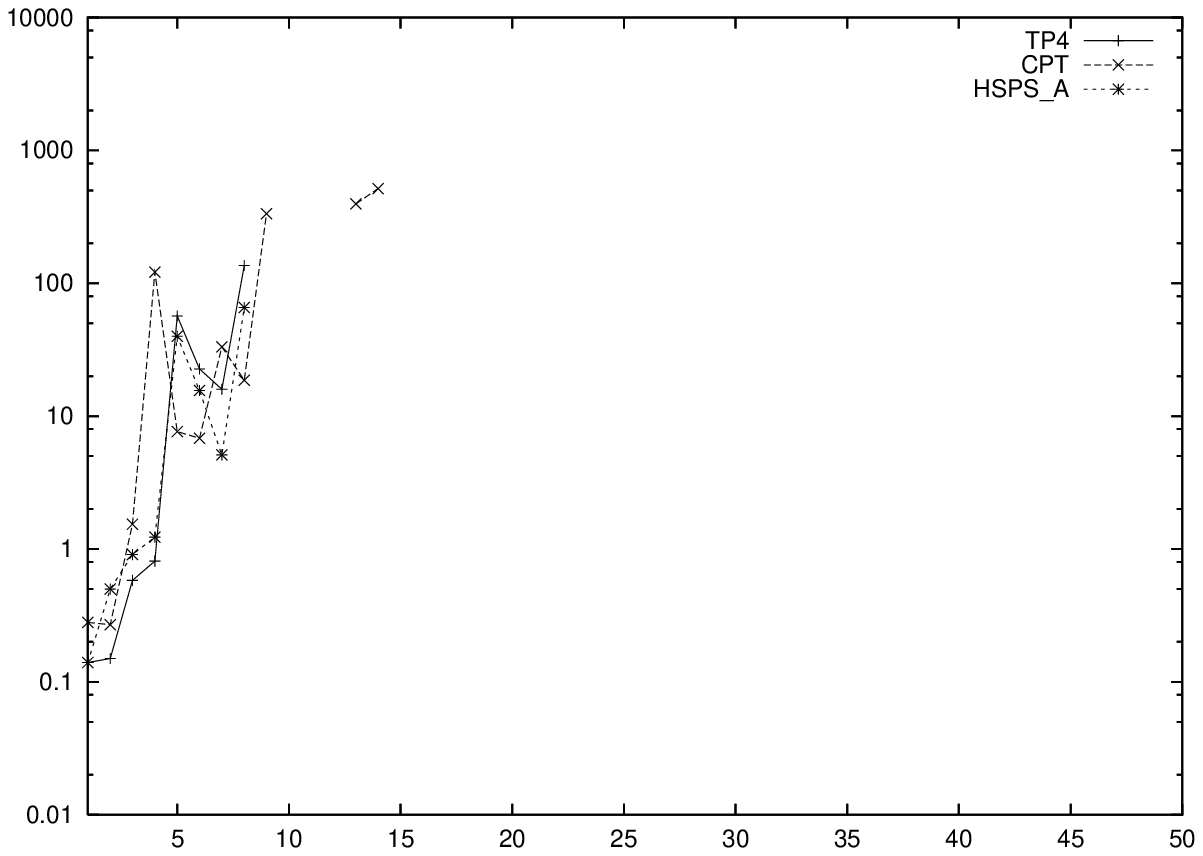}\\
(a) & (b)
\end{tabular}
\vspace{0.0cm}
\caption{Temporal Pipesworld without tankage constraints, satisficing (a), optimal (b).}
\label{pipesworld:nontankagetemp}
\vspace{0.0cm}
\end{center}
\end{figure}

In parts (a) and (b) of Figure~\ref{pipesworld:nontankagetemp}, we
display the results in the temporal domain version without tankage
restrictions. Part (a) shows a clear win of SGPlan over the other
planners. We find it particularly remarkable that, as before in
Airport, SGPlan is just as good in the temporal domain version as in
the nontemporal one.  Again, LPG-3 is outperformed by far. As for the
optimal planners in part (b) of the figure, CPT scales best; TP4 and
HSP$_a^*$ scale similarly. The only planners minimizing the number of
actions here are CRIKEY and SGPlan; the ratio is $[0.35 (1.24) 5.67]$,
showing quite some variance with a slight mean advantage for SGPlan.
Of the planners minimizing makespan, LPG-TD and P-MEP sometimes return
long plans; in one instance (number 2), LPG-TD's plan is extremely
long (makespan 432). More precisely, the ratio LPG-TD.speed vs the
optimal CPT is $[1.00 (4.55) 36.00]$; the ratio P-MEP vs CPT is $[1.19
(2.25) 4.59]$; the ratio LPG-TD.quality vs CPT is $[0.73 (1.05) 1.62]$
(i.e., in particular the peak at instance 2 disappears for
LPG-TD.quality). Note that, strangely at first sight, sometimes LPG-TD
finds better plans here than the optimal CPT. This is due to the
somewhat simpler model of durative actions that CPT uses
\cite{vidal:geffner:aaai-04}, making no distinction between the start
and end time points of actions.

\begin{figure}[t]
\begin{center}
\vspace{-0.0cm}
\begin{tabular}{cc}
\includegraphics[width=7.3cm]{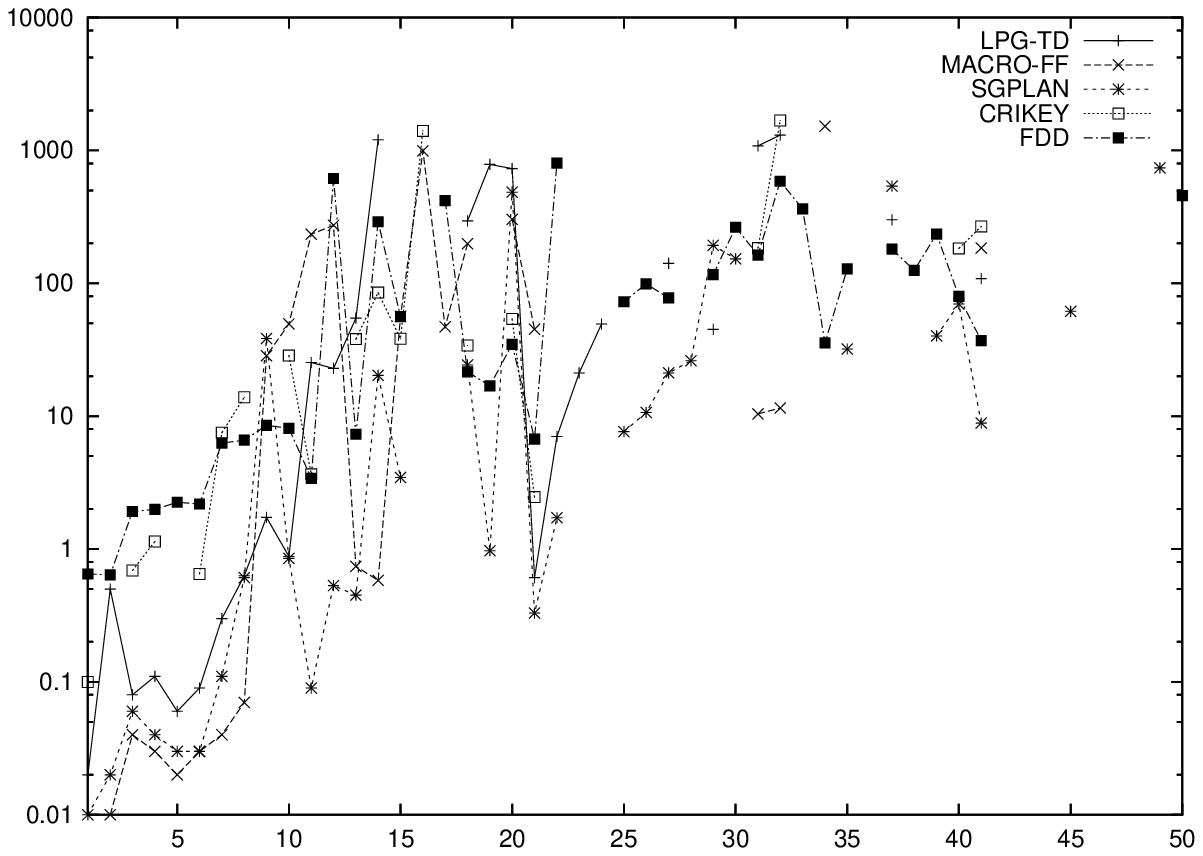} &
\includegraphics[width=7.3cm]{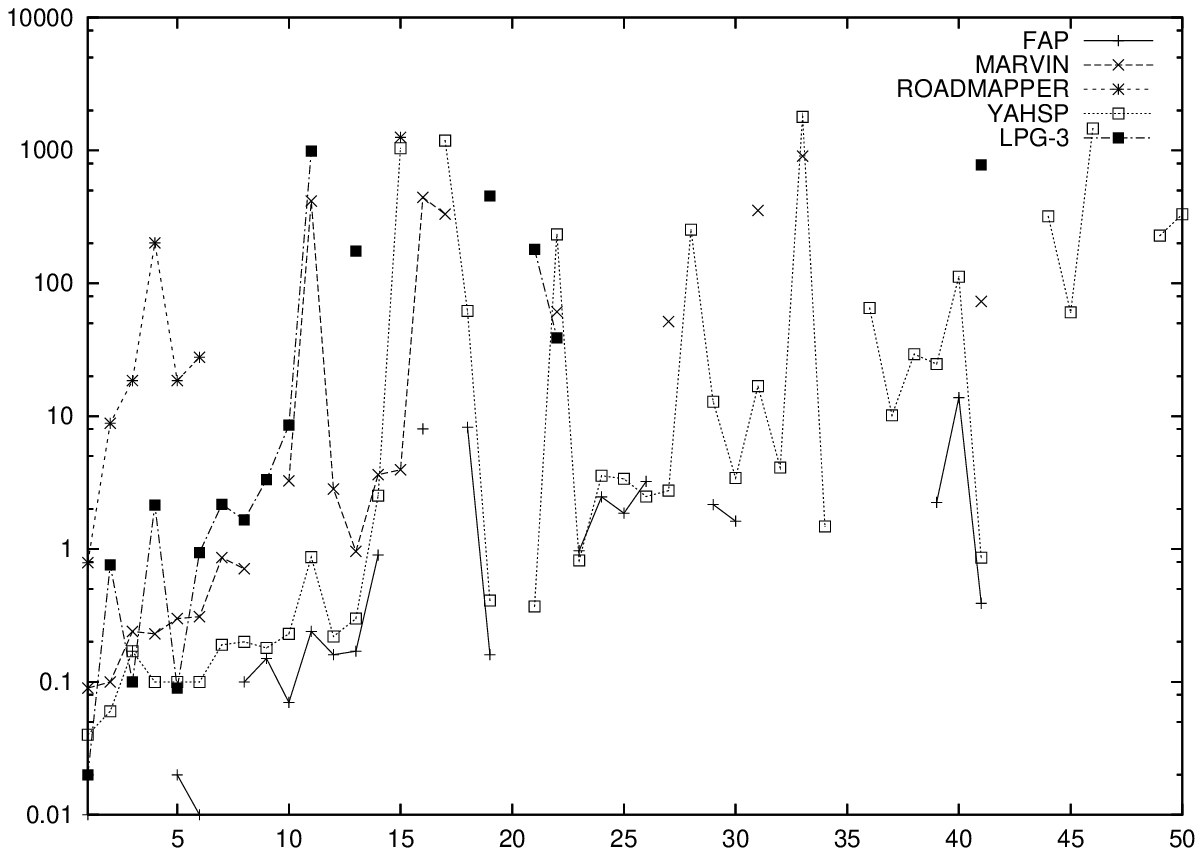}\\
(a) & (b)
\end{tabular}
\includegraphics[width=7.3cm]{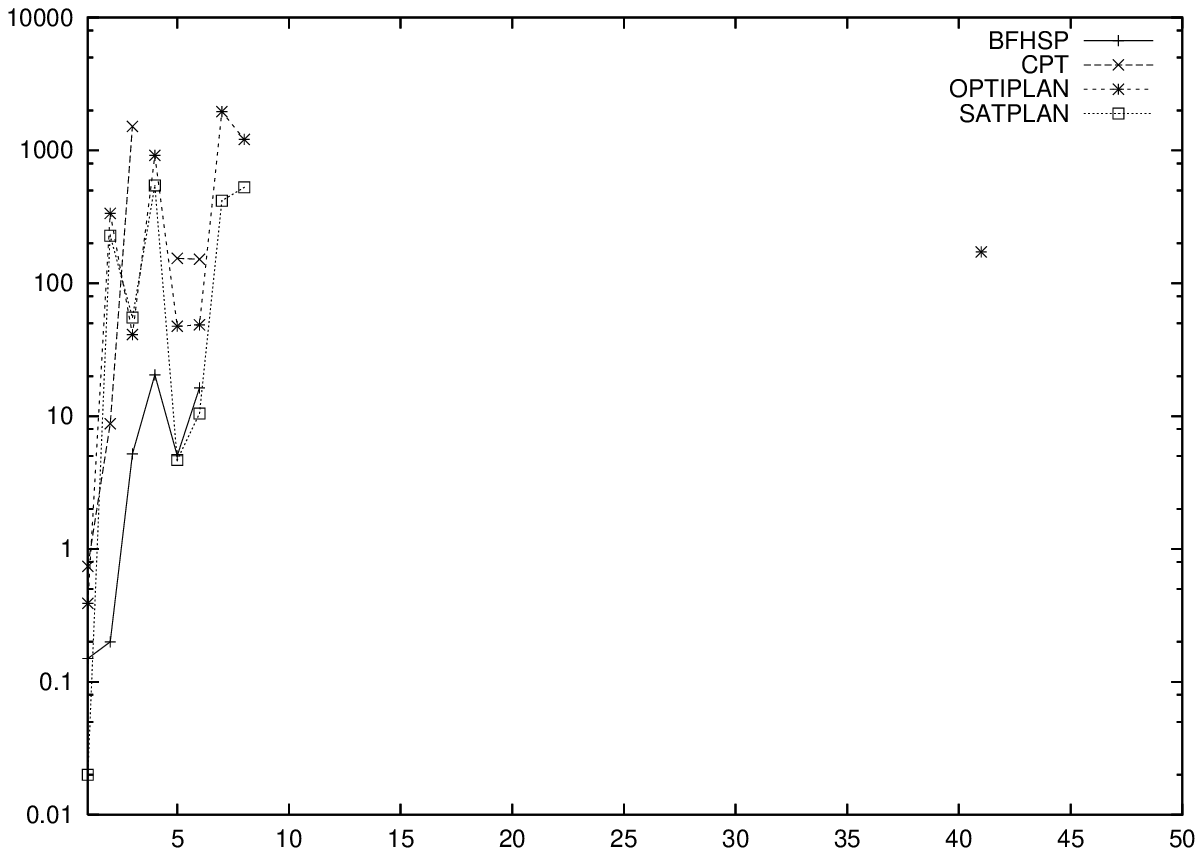}\\
(c)\\
\begin{tabular}{cc}
\includegraphics[width=7.3cm]{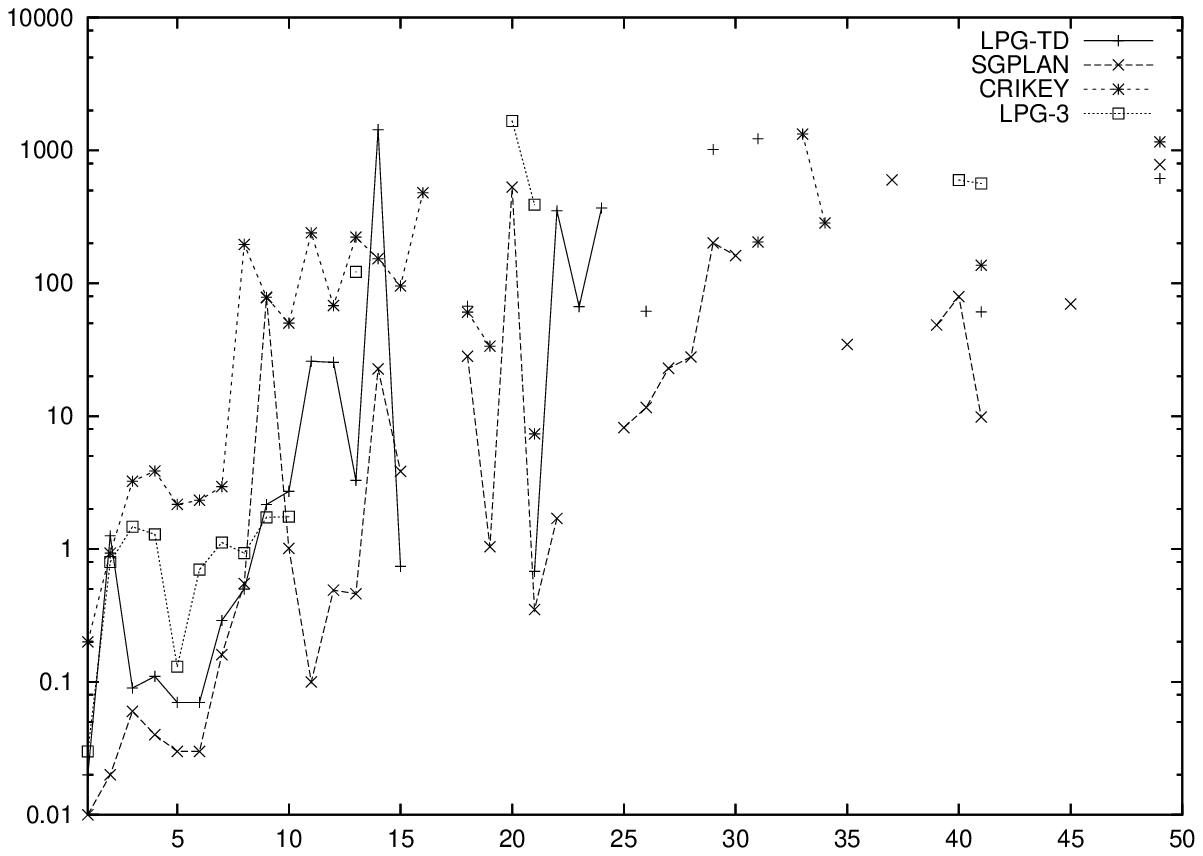} &
\includegraphics[width=7.3cm]{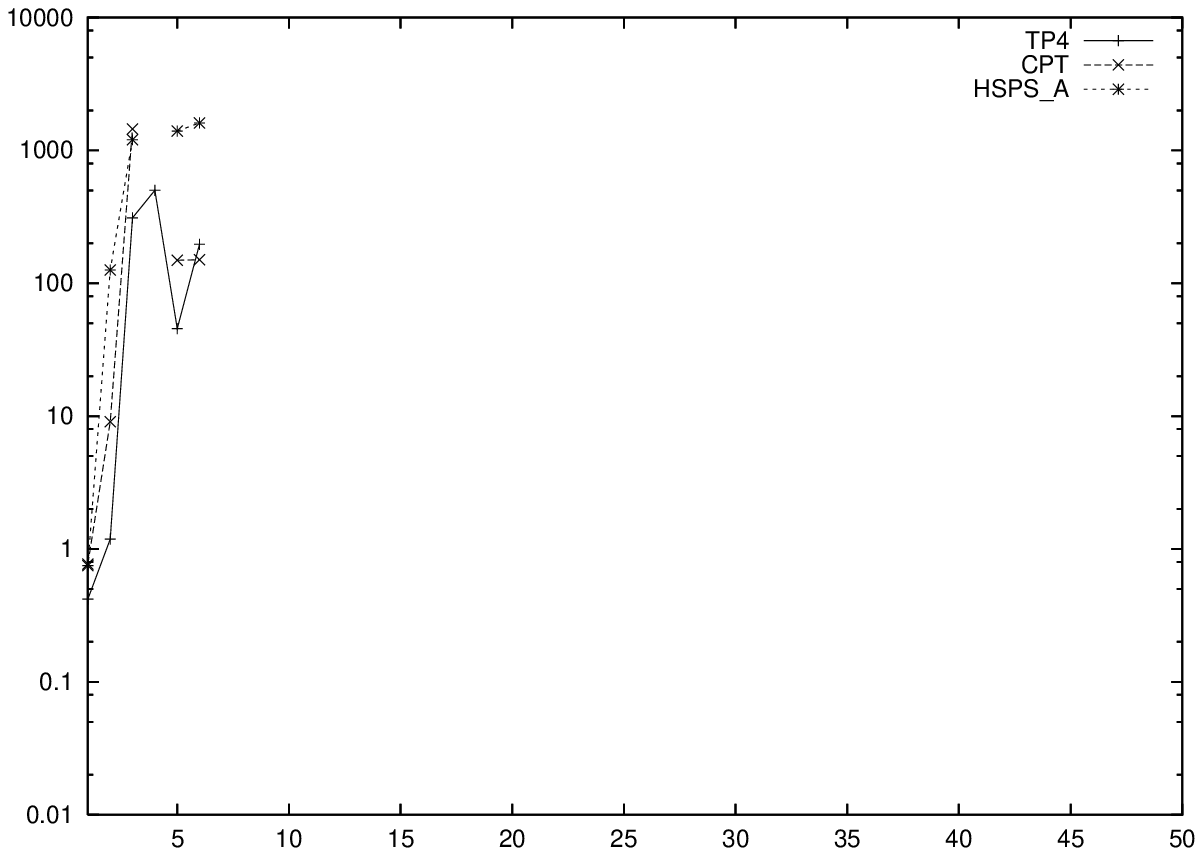}\\
(d) & (e)
\end{tabular}
\vspace{0.0cm}
\caption{Pipesworld with tankage constraints: non-temporal satisficing (a) and (b), 
  optimal (c), temporal satisficing (d), optimal (e).}
\label{pipesworld:tankage}
\vspace{0.0cm}
\end{center}
\end{figure}

In Figure~\ref{pipesworld:tankage}, results are displayed for the two
domain versions that do feature tankage restrictions. Very generally,
we see that the planners have much more trouble with these domain
versions than with their less constrained counterparts.

In the non-temporal domain version with tankage constraints, for the
satisficing planners in parts (a) and (b) we observe that, as before,
YAHSP is most efficient, solving more instances than any of the other
planners.  Again, this efficiency is bought at the cost of overlong
plans: the ratio YAHSP vs. SGPlan is $[0.66 (1.89) 6.14]$, where in
the maximum case SGPlan needs 64 actions, YAHSP 393. For the other
planners, plan quality does not vary as much (e.g. the ratio FDD vs
SGPlan is $[0.56 (0.96) 1.38]$. Of the optimal planners displayed in
part (c), same non-temporal domain version, SATPLAN and Optiplan scale
a little better than the other planners, solving two and three more
instances, respectively.  SATPLAN tends to be faster than Optiplan in
the solved cases. Part (d) displays the results for the satisficing
planners in the respective temporal domain version. SGPlan clearly
scales best, once again keeping its performance from the corresponding
non-temporal domain version. LPG-TD is followed relatively closely by
CRIKEY but can solve some more instances. Regarding plan quality,
CRIKEY and SGPlan minimize plan length, and the ratio CRIKEY vs SGPlan
is $[0.62 (1.37) 2.54]$. For makespan, LPG-TD again has a peak in
instance 2; the ratio LPG-TD.speed vs CPT is $[1.00 (2.55) 7.27]$,
maximum 160 vs 22 units; the ratio LPG-TD.quality vs CPT is $[0.86
(0.95) 1.18]$ (LPG-TD.quality's makespan in instance 2 is $26$). Note
that, as above, LPG-TD.quality can find better plans than CPT due to
the somewhat simpler model of durative actions used in CPT. In the
optimal track, TP4 performs a little better, in terms of runtime, than
the similar-performing CPT and HSP$_a^*$; all the planners solve only
a few of the smallest instances.

We remark that we were rather surprised by the relatively good scaling
behavior of some planners, particularly of the fastest satisficing
planners in the domain versions without tankage restrictions.
\citeA{hoffmann:jair-05} shows that, in Pipesworld, there can be
arbitrarily deep local minima under relaxed plan distances (i.e., the
distance from a local minimum to a better region of the state space
may be arbitrarily large). No particularly unusual constructs are
needed to provoke local minima -- for example, a cycle of pipe
segments, as also occurs in the network topologies underlying the
IPC-4 instances, suffices. Not less importantly, in the limited
experiments we did during testing, FF \cite{hoffmann:nebel:jair-01}
and Mips \cite{edelkamp:jair-03} generally scaled much worse than,
e.g., YAHSP and SGPlan do as shown above.
% So the lookahead and
% problem decomposition techniques employed by these planners,
% respectively, seem to be quite effective in this domain.

As for the domain versions that do feature tankage restrictions, these
are harder than their counterparts. There are two important
differences between the domain versions with/without tankage
restrictions. First, the problem is more constrained (remember that,
modulo the tankage constraints, the IPC-4 instances are identical in
both respective versions). Second, the tank slots, that model the
restrictions, introduce additional symmetry into the problem: a
planner can now choose into which (free) slot to insert a batch. This
choice does not make a difference, but enlarges the state space
exponentially in the number of such choices (to the basis of the
number of tank slots). We considered the option to use a ``counter''
encoding instead, to avoid the symmetry: one could count the product
amounts in the areas, and impose the tankage restrictions based on the
counter values. We decided against this because symmetry is a
challenging aspect of a benchmark.

In domain version notankage-nontemporal, we awarded 1st places to
YAHSP and SGPlan; we awarded 2nd places to LPG-TD and FDD. In version
notankage-temporal, we awarded 1st places to SGPlan and CPT; we
awarded 2nd places to LPG-TD, TP4, and HSP$_a^*$. In version
tankage-nontemporal, we awarded 1st places to YAHSP and FDD; we
awarded 2nd places to SGPlan, SATPLAN, and Optiplan. In version
tankage-temporal, we awarded a 1st place to SGPlan; we awarded 2nd
places to LPG-TD and TP4. In version notankage-temporal-deadlines, we
awarded a 1st place to LPG-TD.

\subsection{Promela}
\label{results:promela}

Promela is the input language of the model checker
SPIN~\cite{Holzmann04}, used for specifying communication protocols.
Communication protocols are distributed software systems, and many
implementation bugs can arise, like deadlocks, failed assertions, and
global invariance violations. The model checking
problem~\cite{ModelChecking} is to find those errors by returning a
counter-example, or to verify correctness by a complete exploration of
the underlying state-space. \citeA{edelkamp:spin-03} developed an
automatic translation of this problem, i.e., of the Promela language,
into (non-temporal) PDDL, making it possible to apply state-of-the-art
planners without any modification.

For IPC-4, two relatively simple communication protocols were selected
as benchmarks -- toy problems from the Model-Checking area. One is the
well-known Dining-Philosophers protocol, the other is a larger
protocol called Optical-Telegraph. The main point of using the domain,
and these protocols, in IPC-4 was to promote the connection between
Planning and Model-Checking, and to test how state-of-the-art planners
scale to the basic problems in the other area.

The IPC-4 instances exclusively require to find deadlocks in the
specified protocols -- states in which no more transitions are
possible. The rules that detect whether or not a given state is a
deadlock are most naturally modelled by derived predicates, with
complex ADL formulas in the rule bodies. To enable broad
participation, we provided compilations of derived predicates into
additional actions, and of ADL into STRIPS. Edelkamp's original
translation of Promela into PDDL also makes use of numeric variables.
For the non-numeric planners, the translation was adapted to use
propositional variables only -- for the relatively simple
Dining-Philosophers and Optical-Telegraph protocols, this was
possible.

Precisely, the domain versions and formulations were the following.
First, the versions were split over the modelled protocol,
Dining-Philosophers or Optical-Telegraph. Second, the versions were
split over the used language: with/without numeric variables, and
with/without derived predicates.\footnote{We split with/without
  numeric variables into different domain versions, rather than
  formulations, to encourage the use of numeric planning techniques.
  We split with/without derived predicates into different versions
  since, through the modelling of derived predicates as new actions,
  the plans become longer.} So, all in all, we had 8 domain versions.
Each of them used ADL constructs. In those 4 versions without numeric
variables, we also provided STRIPS formulations, obtained from the ADL
formulations with automatic compilation software based on FF's
pre-processor \cite{hoffmann:nebel:jair-01}, producing fully grounded
encodings. All Promela domain versions are non-temporal.

Here, we show plots only for the results obtained in the non-numeric
domain versions. In the domain versions using both numeric variables
and derived predicates, {\em no} planner participated. In the numeric
version of Dining-Philosophers (without derived predicates), only
SGPlan and P-MEP participated. SGPlan solved all instances while P-MEP
solved only a few of the smallest ones. In the numeric version of
Optical-Telegraph (without derived predicates), only SGPlan
participated, solving some relatively small instances.

\begin{figure}[t]
\begin{center}
\vspace{-0.0cm}
\begin{tabular}{cc}
\includegraphics[width=7.3cm]{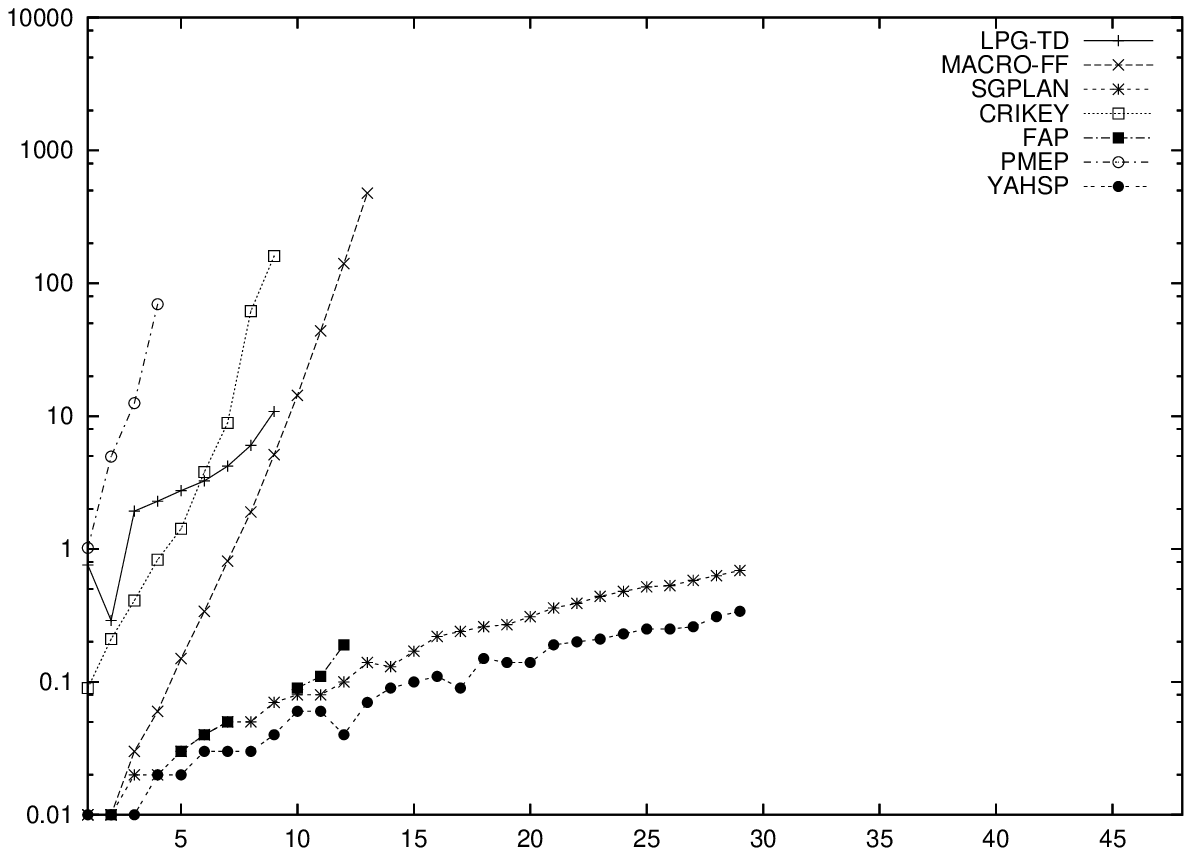} & 
\includegraphics[width=7.3cm]{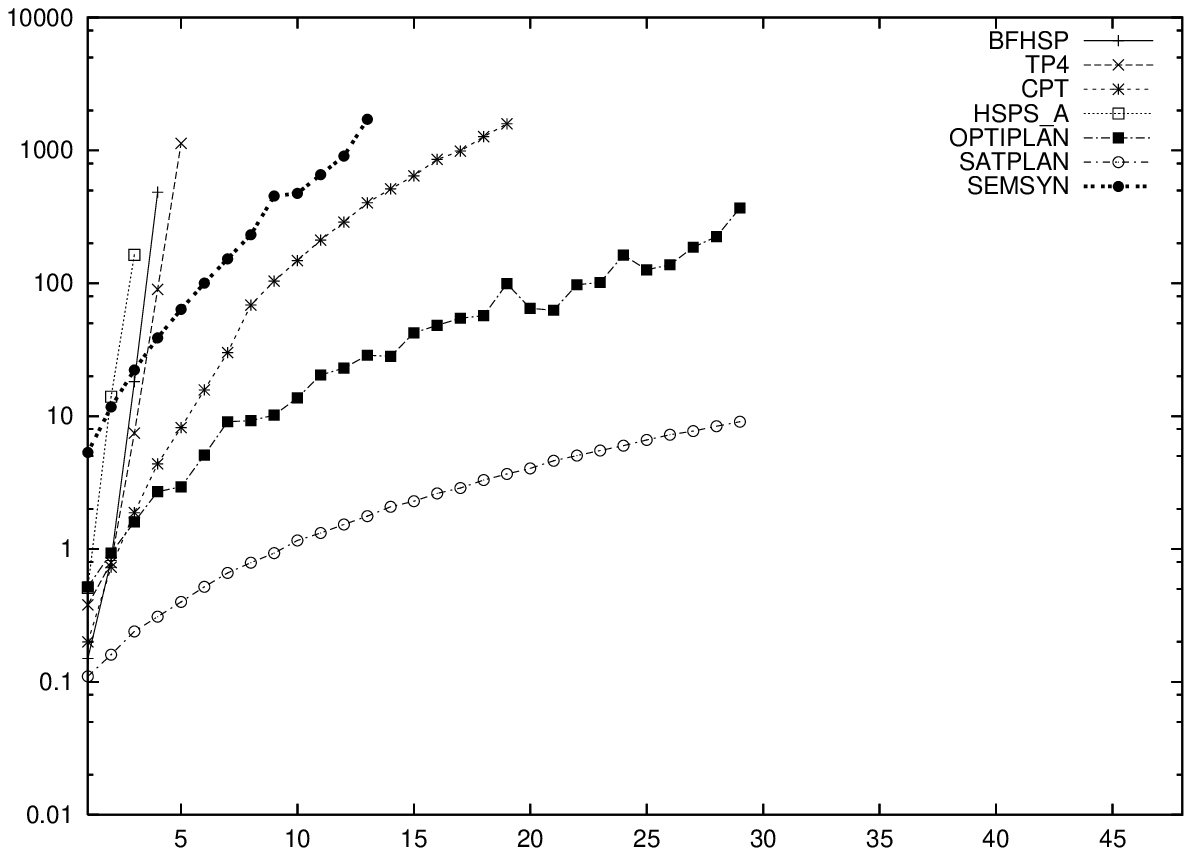}\\
(a) & (b)
\end{tabular}
\includegraphics[width=7.3cm]{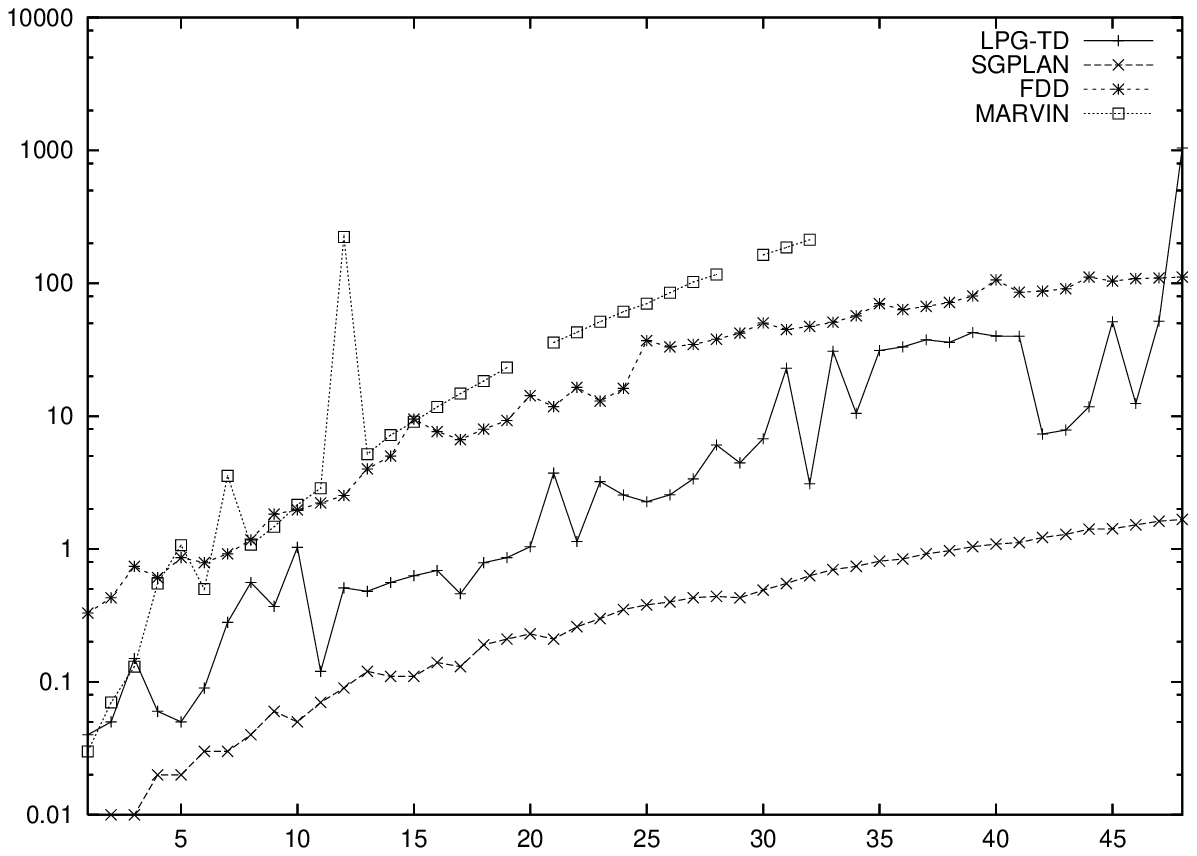}\\
(c)
\caption{Performance in Promela/Dining Philosophers. Encoding without derived predicates, satisficing (a), 
  optimal (b). Encoding with derived predicates, satisficing (c).}
\label{fig:phil}
\vspace{0.0cm}
\end{center}
\end{figure}

Parts (a) and (b) of Figure~\ref{fig:phil} show the results for
Dining-Philosophers, without derived predicates, i.e. with additional
actions to derive the deadlocks. In the the satisficing track, YAHSP
and SGPlan clearly show the best performance. The other planners all
lag several orders of magnitude behind. In the optimal track, SATPLAN
clearly scales best, followed by Optiplan. Observe that, in difference
to what we have seen in Airport and Pipesworld, {\em the optimal
  planners are just as efficient as the satisficing ones.} This is
evident from the plots. Most particularly, SATPLAN and Optiplan solve
just as many instances, up to $30$ philosophers (instance number $x$
features $x+1$ philosophers), like YAHSP and
SGPlan.\footnote{Importantly, these planners handle only STRIPS, and
  for compilation reasons there were STRIPS instances up to $n=30$
  only. So these planners solve {\em all} their respective test
  instances. We discuss this in a little more detail further below.}
SATPLAN even does so in comparable time. Such a competitivity of
optimal planners with satisficing ones has not been seen in any test
suite used in any of the last three competitions.\footnote{In the first
  IPC, in 1998, apart from the fact that the ``heuristic search''
  satisficing planners were still in their infancy, Mystery and Mprime
  were used as benchmarks. These can't be solved particularly
  efficiently by any planner up to today, except, maybe, FD and FDD
  \cite{helmert:icaps-04}.} The efficiency of SATPLAN and Optiplan in
Dining-Philosophers is probably due the fact that the needed number of
parallel time steps is {\em constantly} $11$ across all instances (see
below).  In a (standard) planning graph
\cite{blum:furst:ijcai-95,blum:furst:ai-97}, the goals are first
reached after $7$ steps; so only 4 unsuccessful iterations are made
before a plan is found.

Regarding plan quality, there is one group of planners trying to
minimize the number of actions, and one group trying to minimize the
makespan (the number of parallel action steps). The results clearly
point to an important difference between this domain and the other
IPC-4 domains. Namely, there is only a single scaling parameter $n$ --
the number of philosophers -- and only a single instance per value of
$n$.  The optimal number of actions is a linear function of $n$,
precisely $11n$: basically, one blocks all philosophers in sequence,
taking $11$ steps for each. The optimal makespan is, as mentioned
above, constantly $11$: one can block all philosophers in parallel.
The only sub-optimal plan found by any planner minimizing the number
of actions is that of CRIKEY for $n=8$, containing 104 instead of 88
actions. Of the planners minimizing makespan, only P-MEP finds
sub-optimal plans: it solves only the smallest four instances,
$n=2,3,4,5$, with a makespan of 19, 25, 30, and 36, respectively.

Figure~\ref{fig:phil} (c) shows the results in the domain version that
uses derived predicates to detect the deadlock situations. No optimal
planner could handle derived predicates which is why there are no
results for this. Of the satisficing planners, SGPlan clearly shows
the best performance. FDD and LPG-TD are roughly similar -- note that,
while LPG-TD is faster than FDD in most examples, the behavior of FDD
in the largest instances indicates an advantage in scaling. There is a
sudden large increase in LPG-TD's runtime, from the second largest
instance (that is solved in 52 seconds) to the largest instance (that
is solved in 1045 seconds). In difference to that, FDD solves the
largest instance in almost the same time as the second largest one,
taking 111 seconds instead of 110 seconds (in fact, FDD's runtime
performance shows little variance and is pretty much a linear function
in the instance size). We remark that the plans for the largest of
these instances are huge: more than 400 steps long, see below. SGPlan
generates these plans in little more than a single second CPU time.

Regarding plan quality, in this domain version the optimal number of
actions is $9n$, and the optimal makespan is constantly $9$. All
planners except Marvin tried to minimize the number of actions. FD and
SGPlan always find optimal plans. FDD, LPG-TD.speed, and
LPG-TD.quality sometimes find slightly sub-optimal plans. Precisely,
the ratio FDD vs FD is $[1.00 (1.08) 1.44]$; LPG-TD.speed vs FD is $[1.00
(1.10) 2.19]$; LPG-TD.quality vs FD is $[1.00 (1.03) 1.24]$. As for Marvin,
the makespan of its plans is, roughly, linear in $n$; it always lies
between $7n$ and $9n$.

At this point, there is an important remark to be made about a detail
regarding our compilation techniques in these domain versions. Observe
the huge gap in planner performance between Figure~\ref{fig:phil} (a)
and (b), and Figure~\ref{fig:phil} (c). As we have seen, the
difference in plan length (makespan) is not very large -- $11n$ ($11$)
compared to $9n$ ($9$). Indeed, the apparent performance difference is
due to a {\em compilation} detail, inherent in the IPC-4 instances,
rather than to planner performance. In the version without derived
predicates, we were able to compile only the instances with up to
$n=30$ philosophers from ADL into STRIPS. For larger values of $n$,
the (fully-grounded) STRIPS representations became prohibitively
large. In the version {\em with} derived predicates, the blow-up was a
lot smaller, and we could compile {\em all} instances, with up to
$n=49$ philosophers. SGPlan, YAHSP, SATPLAN, and Optiplan can all
handle only STRIPS. So, as mentioned above, in Figure~\ref{fig:phil}
(a) and (b) these planners actually solve all instances in their
respective test suite; they would probably scale up further when
provided with STRIPS representations of instances with larger $n$
values.

\begin{figure}[htb]
\begin{center}
\vspace{-0.0cm}
\begin{tabular}{cc}
\includegraphics[width=7.3cm]{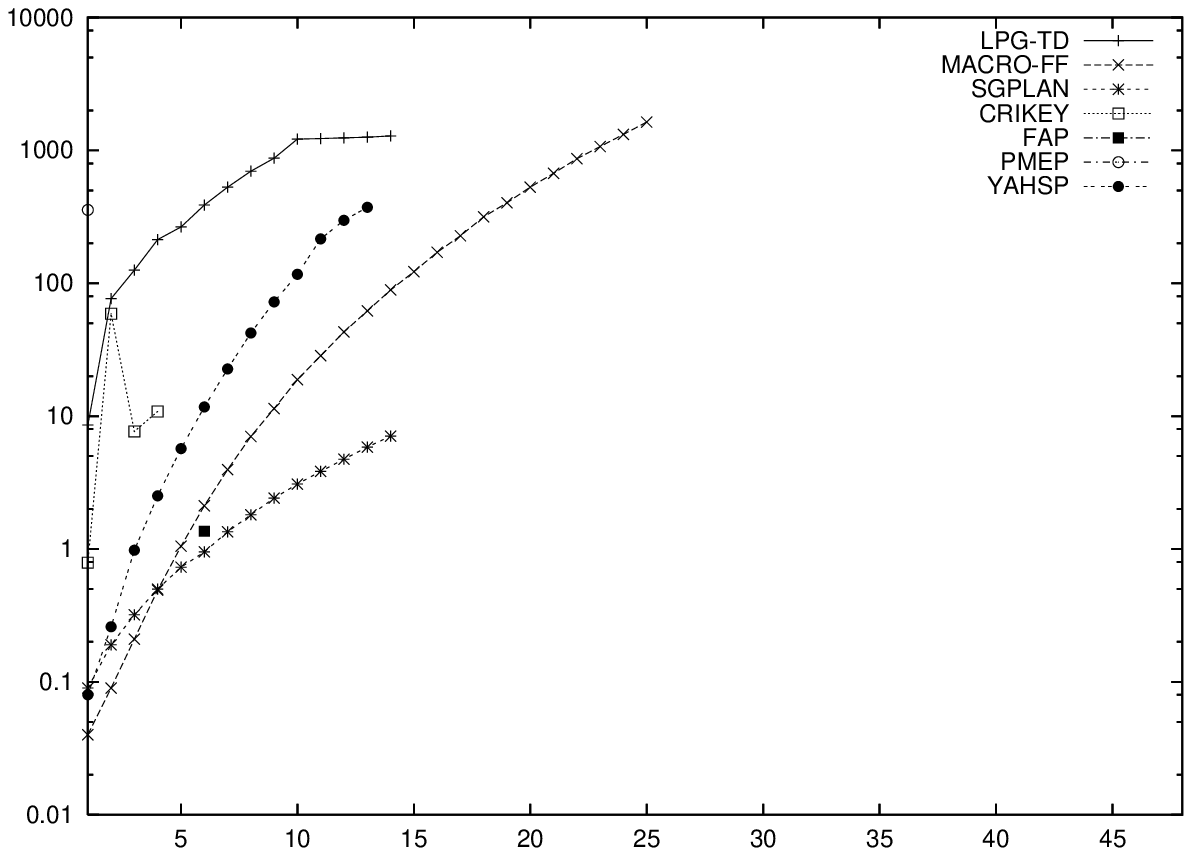} & 
\includegraphics[width=7.3cm]{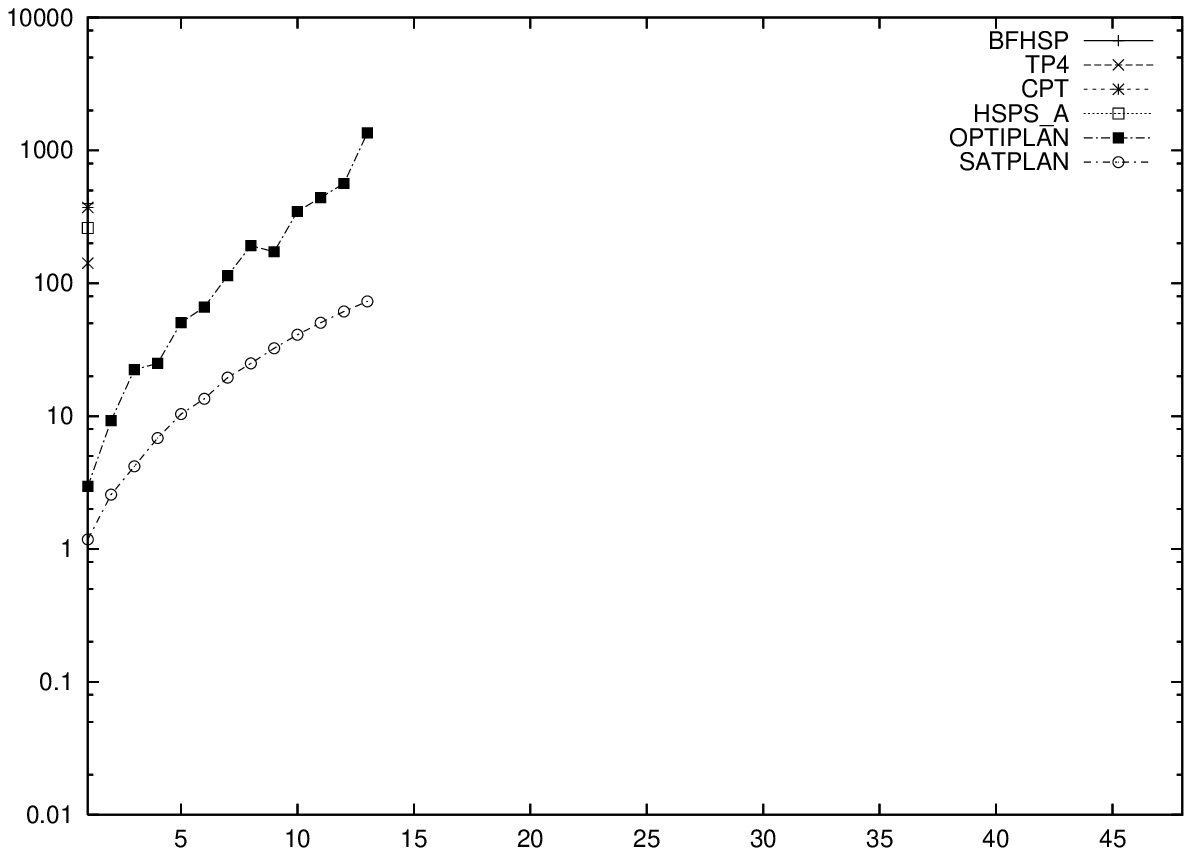}\\
(a) & (b)
\end{tabular}
\includegraphics[width=7.3cm]{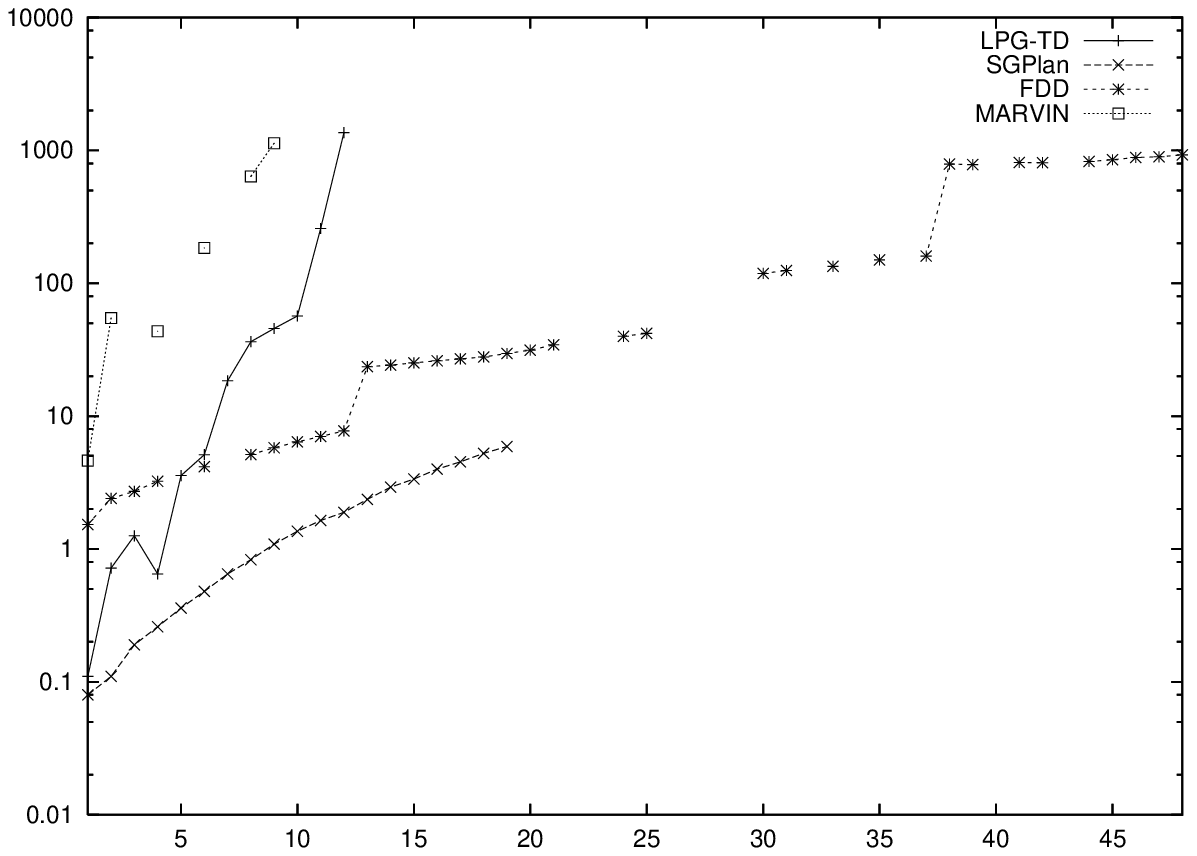}\\
(c)
\caption{Performance in Promela/Optical Telegraph. Encoding without derived predicates, satisficing (a), 
  optimal (b). Encoding with derived predicates, satisficing (c).}
\label{fig:opt}
\vspace{0.0cm}
\end{center}
\end{figure}

The Optical-Telegraph protocol is slightly more complex than the
Dining-Philosopher example, i.e. it involves more complicated and
indirect interactions between the communicating processes, leading to
longer solutions. There is still just one scaling parameter, the
number $n$ of ``telegraph station pairs'', and a single instance per
value of $n$. Each telegraph station pair is a pair of processes that
goes through a rather complicated internal communication structure,
enabling the exchange of data. The telegraph station pair shares with
the outside world -- i.e., with the other telegraph station pairs --
two ``control channels'' that must be occupied as a prerequisite to
the internal exchange of data. Thus, the role of the control channels
is pretty similar to the role of the (shared) forks in the
Dining-Philosopher example. Instance number $x$ in IPC-4 features
$x+1$ telegraph station pairs.

Figure~\ref{fig:opt} shows in parts (a) and (b) the performance of the
satisficing and optimal planners, respectively, in the domain version
using additional actions to derive deadlocks. In the group of
satisficing planners, Macro-FF clearly performs best. Of the other
satisficing planners, SGPlan is the most efficient. It is important to
note that, here, we were able to compile only the instances up to
$n=15$ from ADL into STRIPS. So SGPlan and LPG-TD solve all their
respective test instances. Of the optimal planners, SATPLAN and
Optiplan solve the instances up to $n=14$, i.e., they failed to solve
the largest instance in the STRIPS test suite they attacked. SATPLAN
is much faster than Optiplan. Each of the other optimal planners could
solve only the smallest instance, $n=2$. Figure~\ref{fig:opt} (c)
shows that FDD, which handles the ADL formulation, is by far the most
successful satisficing planner in the domain version encoding
Optical-Telegraph with derived predicates for deadlock detection; of
the other planners, SGPlan scales best, solving all instances in the
STRIPS set, up to $n=20$.

Regarding plan quality, in Optical-Telegraph without derived
predicates the optimal number of actions is $18n$, and the optimal
makespan is constantly $13$: optimal sequential plans block all
telegraph station pairs in sequence, taking $18$ steps for each; in
parallel plans, some simultaneous actions are possible {\em within}
each telegraph station pair. In Optical-Telegraph {\em with} derived
predicates, the optimal number of actions is $14n$, and the optimal
makespan is constantly $11$. In the competition results, all planners
returned the optimal plans in these test suites, in all cases -- with
a single exception. The plans found by Marvin in the version with
derived predicates have makespan $14n$ in all solved cases.

As we have seen, the results in Promela are, over all, quite different
from, e.g., those we have seen in Airport and Pipesworld. There is
only a single scaling parameter and a single instance per size, and
optimal makespan is constant. This leads to rather smooth runtime and
plan quality curves, as well as to an unusual competitivity of optimal
planners with satisficing planners. The scalability of the planners
both shows that current planners {\em are} able to efficiently solve
the most basic Model-Checking benchmarks (Dining-Philosophers), and
that they are not very efficient in solving more complex
Model-Checking benchmarks (Optical-Telegraph). We remark that
\citeA{hoffmann:jair-05} shows that there exist arbitrarily deep local
minima under relaxed plan distances in Optical-Telegraph, but not in
Dining-Philosophers, where there exists a large upper bound (31) on
the number of actions needed to escape a local minimum. It is not
clear, however, how much these theoretical results have to do with the
planner performance observed above. The worst-cases observed by
Hoffmann all occur in regions of the state space {\em not} entered by
plans with optimal number of actions -- as found by the IPC-4
participants, in most cases.

In Dining-Philosophers without derived predicates, we awarded 1st
places to YAHSP, SGPlan, and SATPLAN; we awarded a 2nd place to
Optiplan. In Dining-Philosophers with derived predicates, we awarded a
1st place to SGPlan, and 2nd places to FDD and LPG-TD. In
Optical-Telegraph without derived predicates, we awarded 1st places to
Macro-FF and SATPLAN; we awarded 2nd places to SGPlan and Optiplan.
In Optical-Telegraph with derived predicates, we awarded a 1st place
to FDD, and a 2nd place to SGPlan. In the numeric version of
Dining-Philosophers, we awarded a 1st place to SGPlan.

\subsection{PSR}
\label{results:psr}

PSR is short for {\em Power Supply Restoration}. The domain is a PDDL
adaptation of an application domain investigated by Sylvie Thi\'ebaux
and other researchers
\cite{thiebaux:etal:uai-96,thiebaux:cordier:ecp-01}, which deals with
reconfiguring a faulty power distribution system to resupply customers
affected by the faults. In the original PSR problem, various numerical
parameters such as breakdown costs and power margins need to be
optimized, subject to power capacity constraints.  Furthermore, the
location of the faults and the current network configuration are only
partially observable, which leads to a tradeoff between acting to
resupply lines and acting to reduce uncertainty.  In contrast, the
version used for IPC-4 is set up as a pure goal-achievement problem,
numerical aspects are ignored, and total observability is assumed.
Temporality is not a significant aspect even in the original
application, and the IPC-4 domain is non-temporal.

We used four domain versions of PSR in IPC-4. Primarily, these
versions differ by the size of the problem instances encoded. The
instance size determined in what languages we were able to formulate
the domain version. The domain versions are named 1. {\em large}, 2.
{\em middle}, 3. {\em middle-compiled}, and 4. {\em small}. Version 1
has the single formulation {\em adl-derivedpredicates}: in the most
natural formulation, the domain comes with derived predicates to model
the flow of electricity through the network, and with ADL formulas to
express the necessary conditions on the status of the network
connectors. Version 2 has the formulations {\em
  adl-derivedpredicates}, {\em simpleadl-derivedpredicates}, and {\em
  strips-derivedpredicates}.  Version 3 has the single formulation
{\em adl}, and version 4 has the single formulation {\em strips}. As
indicated, the formulation names simply give the language used. In
version 2, ADL constructs were compiled away to obtain the simpler
formulations. In version 3, derived predicates were compiled away by
introducing additional artificial actions; due to the increase in plan
length (which we discuss in some detail further below), we turned this
into a separate domain version, rather than a formulation. As for the
{\em strips}\ domain version, to enable encoding of reasonably-sized
instances for this we adopted a different fully-grounded encoding
inspired by the work of \citeA{bertoli:etal:ecai-02}. The encoding is
generated from a description of the problem instance by a tool
performing some of the reasoning devoted to the planner under the
other domain versions.  Still we were only able to formulate
comparatively small instances in pure STRIPS.

\begin{figure}[t]
\begin{center}
\vspace{-0.0cm}
\begin{tabular}{cc}
\includegraphics[width=7.3cm]{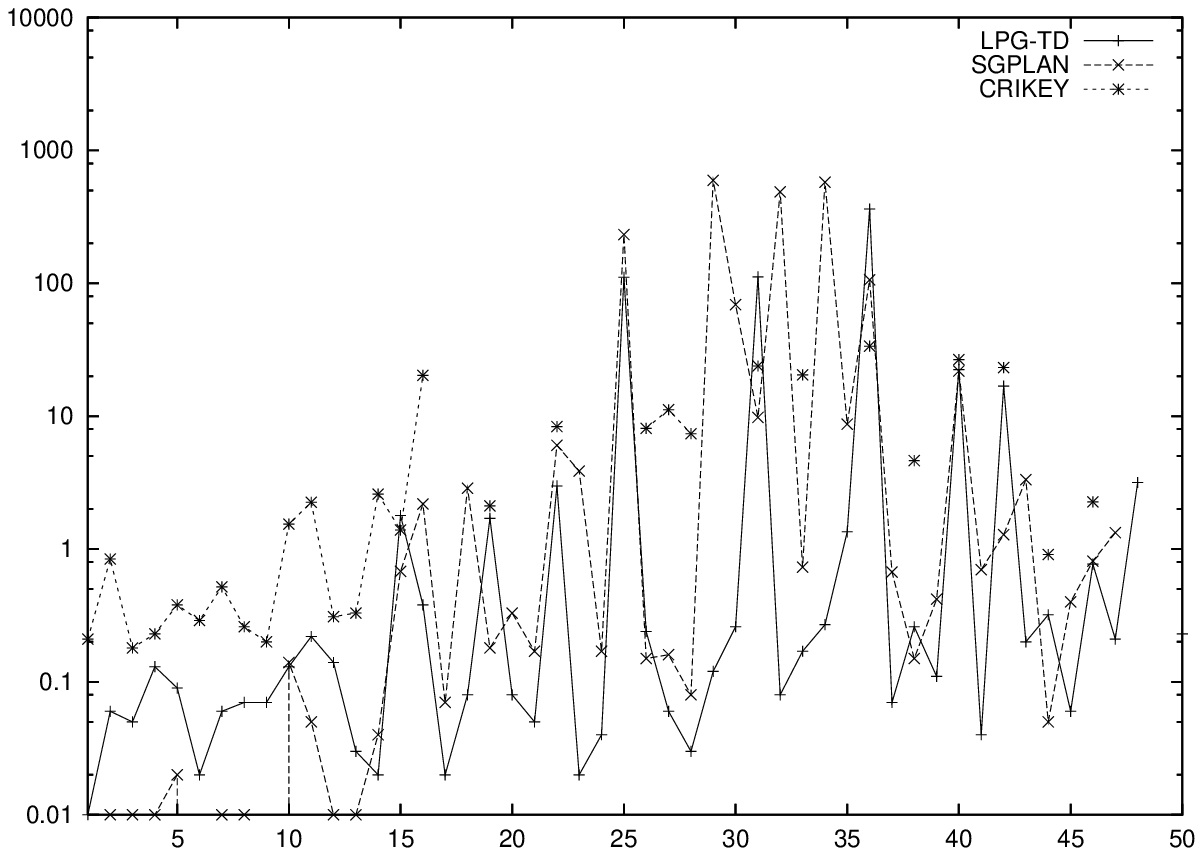} &
\includegraphics[width=7.3cm]{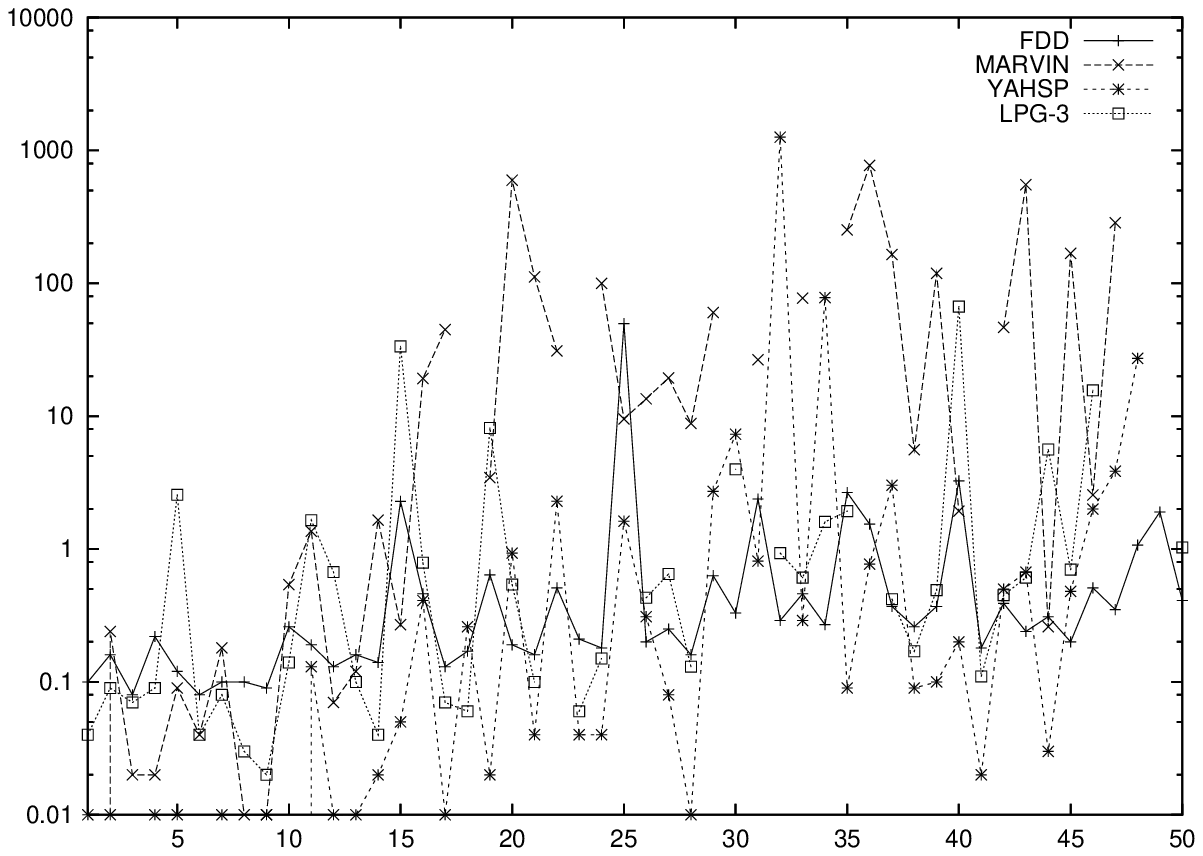} \\
(a) & (b) \\
\includegraphics[width=7.3cm]{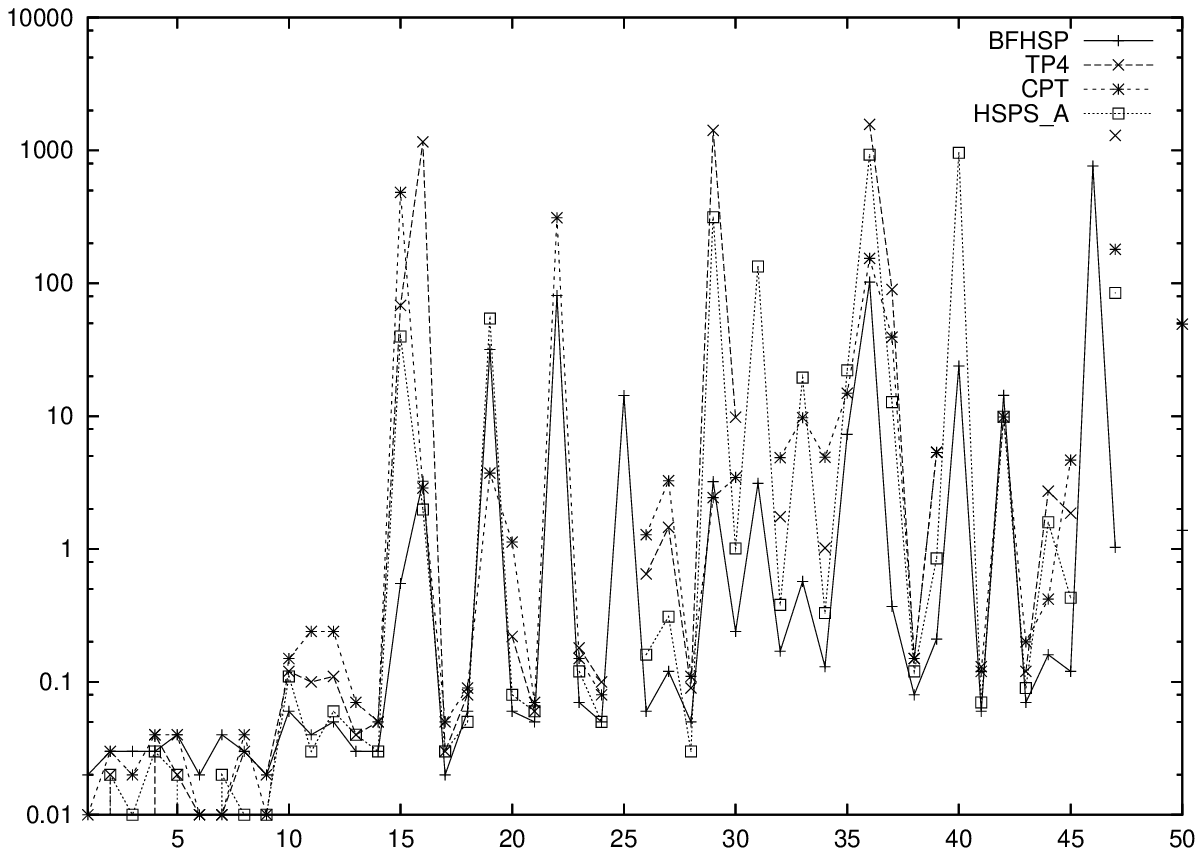} &
\includegraphics[width=7.3cm]{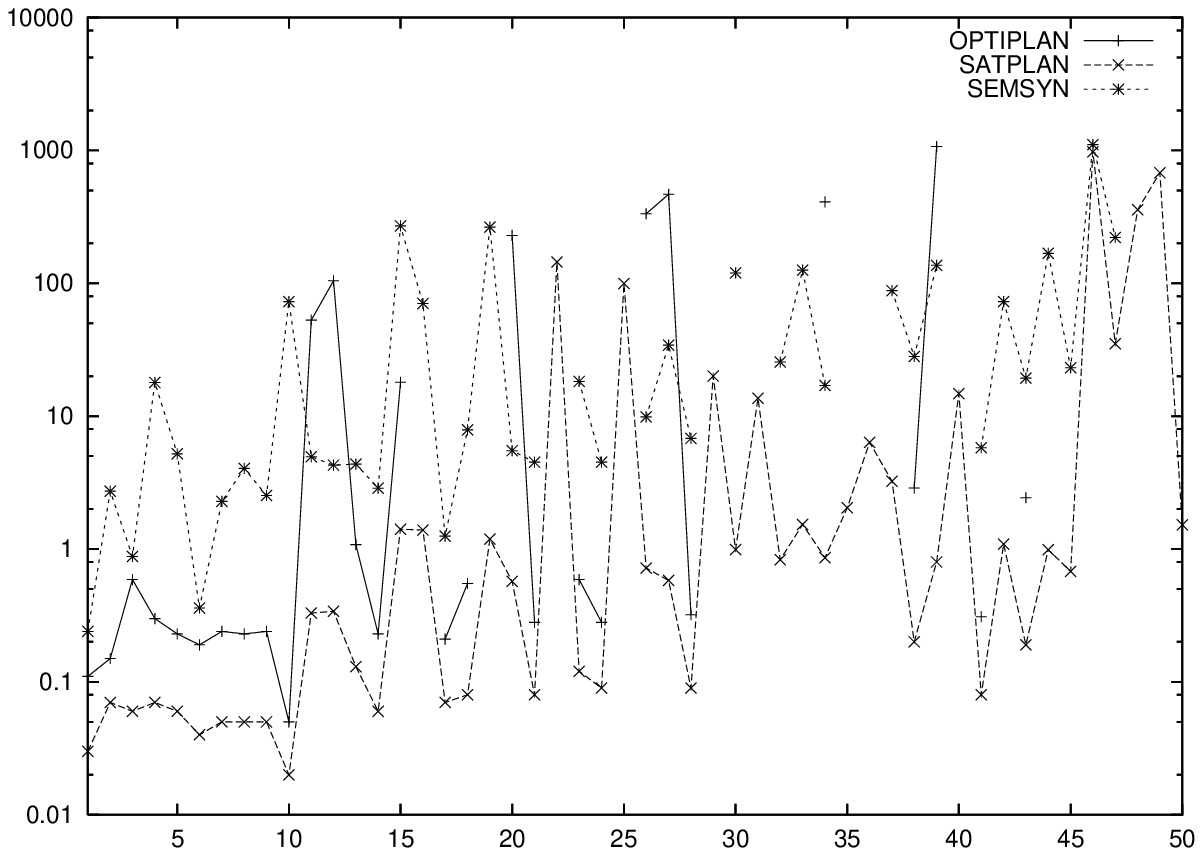} \\
(c) & (d)
\end{tabular}
\caption{PSR \emph{small}, satisficing (a) and (b), optimal (c) and (d).}
\label{psr:small}
\end{center}
\vspace{-0.0cm}
\end{figure}

Starting with the performance in the smallest instances, we depict the
performance of satisficing and optimal planners in the PSR STRIPS
domain in Figure~\ref{psr:small}. Both result graphs are divided into
two because they are completely unreadable otherwise. Most planners
show a lot of variance in this domain, blurring the performance
differences (observable) between the individual systems. Still it is
possible to identify systems that behave better than others. Of the
satisficing systems, only FDD can solve all the 50 instances; it does
so consistently fast (the results for FD are almost identical). Of the
other planners, LPG-TD, SGPlan, and YAHSP have the best success ratio:
they solve 49, 47, and 48 instances, respectively; CRIKEY solves 29
instances, Marvin 41. As for the optimal systems, here the only system
solving the entire test suite is SATPLAN. BFHSP solved 48 instances,
CPT 44, HSP$_a^*$ 44, Optiplan 29, Semsyn 40, TP4 38.

For plan quality, once again there are groups of planners trying to
minimize plan length (number of actions), and makespan. In the former
group, we take as a performance measure the plans found by BFHSP,
which are optimal in that sense, and which, as said above, we have for
all but 2 of the 50 instances. The ratio CRIKEY vs BFHSP is $[1.00
(1.72) 3.44]$; the ratio FDD vs BFHSP is $[1.00 (1.02) 1.52]$; the
ratio LPG-TD.speed vs BFHSP is $[1.00 (5.52) 12.70]$; the ratio
LPG-TD.quality vs BFHSP is $[1.00 (1.82) 8.32]$; the ratio SGPlan vs
BFHSP is $[1.00 (1.01) 1.24]$; the ratio YAHSP vs BFHSP is $[1.00
(1.00) 1.05]$. In the planner group minimizing makespan, all planners
except Marvin are optimal. The ratio Marvin vs SATPLAN is $[1.00
(1.28) 2.07]$.

Apart from the observations to be made within the groups of
satisficing respectively optimal planners, there is something to be
said here on the relationship between these two groups. Like in
Dining-Philosophers, we have the rather unusual situation that the
optimal planners are just as efficient as the satisficing ones.
Indeed, by solving all instances, SATPLAN is superior to most of the
satisficing planners. We remark at this point that
\citeA{hoffmann:jair-05} shows the existence of arbitrarily deep local
minima in PSR, i.e., regions where it takes arbitrarily many step to
escape a local minimum under relaxed plan distances. For example,
local minima arise naturally because a relaxed plan is able to supply
{\em and} not supply a line at the same time, and thereby does not
have to make the crucial distinction between faulty and non-faulty
lines.  Now, these observations hold for the {\em original} domain
formulation, with derived predicates and ADL constructs. As said, the
IPC-4 STRIPS formulation of PSR is obtained from that by a combination
of complicated pre-processing machines. These are not likely to make
the real structure of the domain more amenable to relaxed plan
distances. While the satisficing planners participating in PSR {\em
  small} are by no means exclusively dependent on relaxed plan
distances, for most of them these distances do form an important part
of the search heuristics.

\begin{figure}[t]
\begin{center}
\vspace{-0.0cm}
\begin{tabular}{cc}
\includegraphics[width=7.3cm]{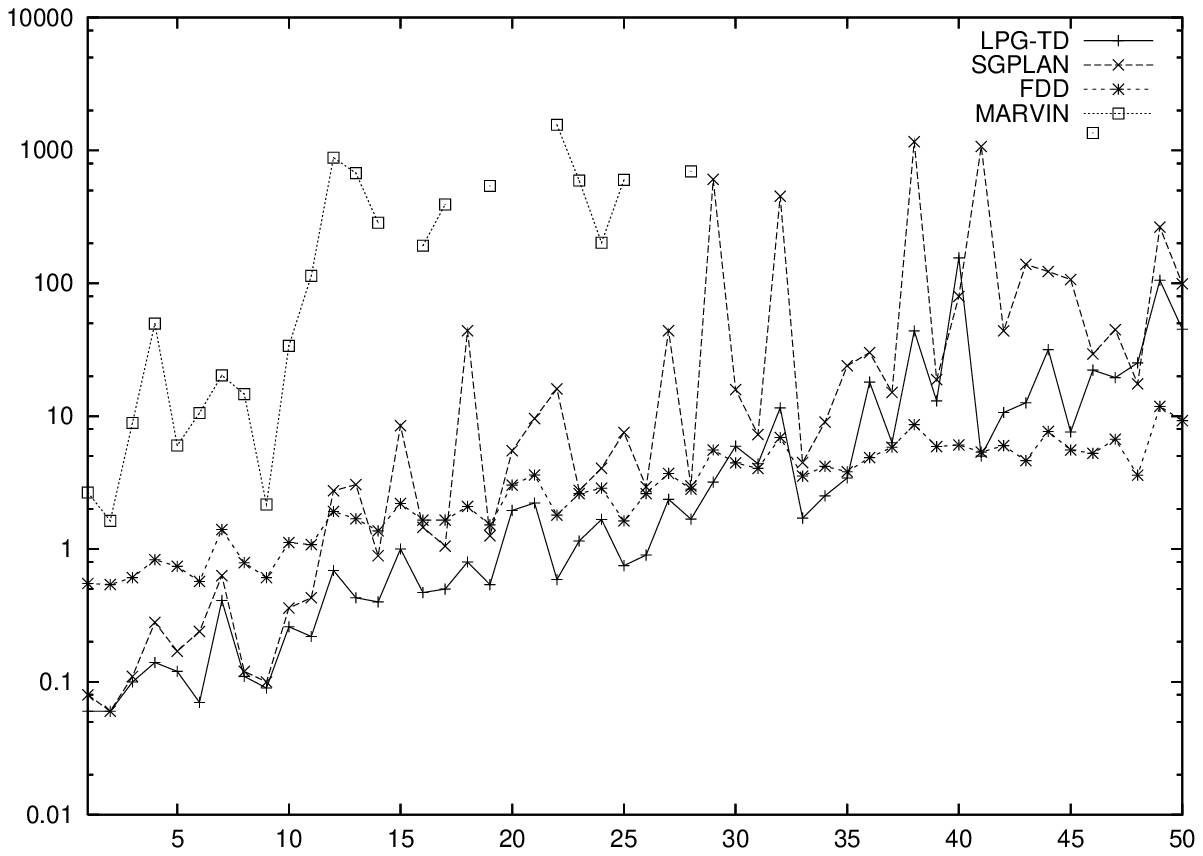} & 
\includegraphics[width=7.3cm]{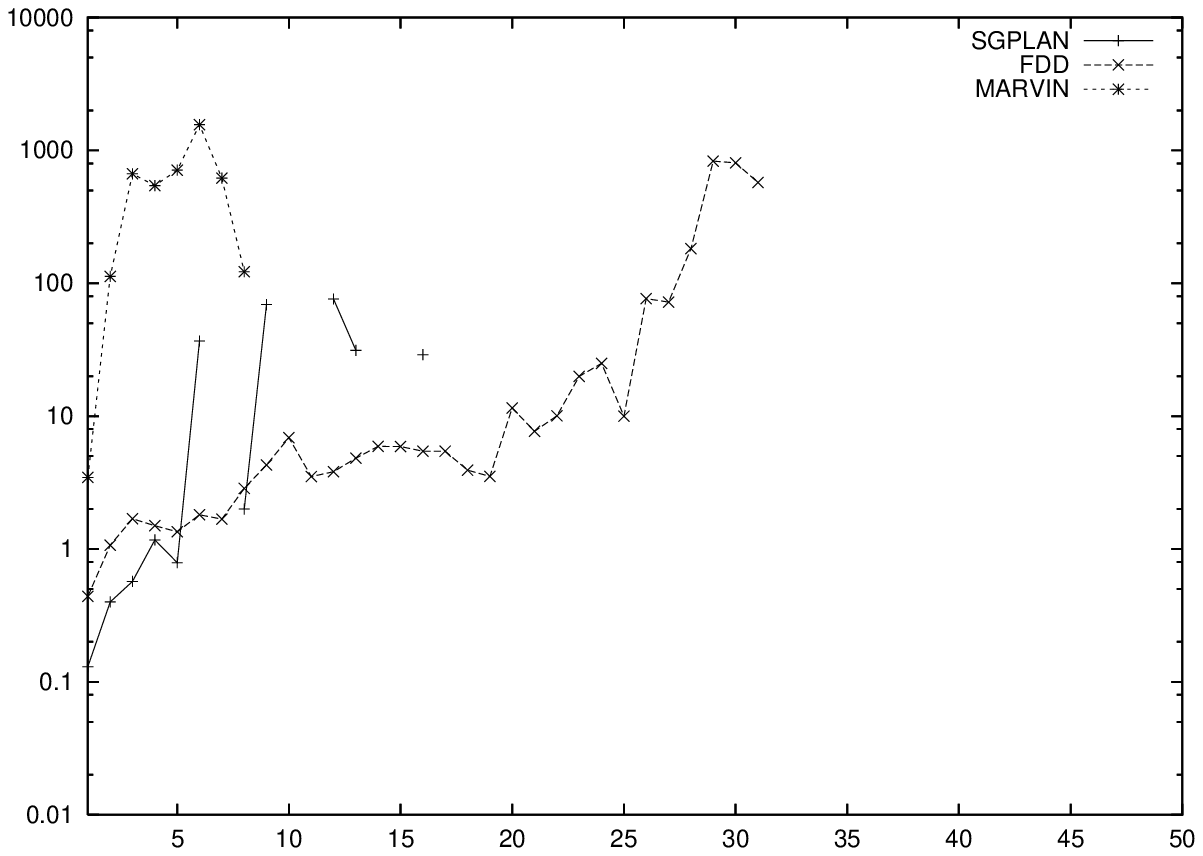} \\
(a) & (b)
\end{tabular}
\caption{PSR, \emph{middle} (a) and \emph{large} (b), satisficing planners.}
\label{psr:middle-large}
\end{center}
\vspace{-0.0cm}
\end{figure}

Figure~\ref{psr:middle-large} gives the results for the PSR domain
versions, {\em middle} and {\em large}, with larger instances, encoded
with ADL constructs and/or derived predicates. No optimal planner
could handle these language constructs, so unfortunately the above
comparison between the two groups can not be continued. In the middle
sized instances, FDD, SGPlan, and LPG-TD all scale through the entire
test suite. FDD indicates better scaling behavior in the largest
instances. In the domain version with large instances (available only
in a formulation using ADL), FDD indeed shows that it outperforms the
other planners (at least, SGPlan, that also participates here) by far.
The data for FD is almost identical to that for FDD.

Regarding plan quality, the only participating planner in each of {\em
  middle} and {\em large} that tries to minimize makespan is Marvin,
so we have no basis for a comparison.  Of the other planners, SGPlan
generally finds the plans with the smallest number of actions.
Precisely, in the {\em middle} version the plan quality ratios are as
follows. The ratio FDD vs SGPlan is $[0.67 (1.23) 2.33]$ (FD: $[0.67 (1.25)
2.85]$). The ratio LPG-TD.speed vs SGPlan is $[0.67 (1.82) 7.27]$; the
maximum case is instance number 49, where LPG-TD.speed takes 80 steps
and SGPlan 11. The ratio LPG-TD.quality vs SGPlan is $[0.60 (1.08) 7.27]$,
with the same maximum case. In the {\em large} version, only FD and
FDD solve a considerable number of instances; the ratio FD vs FDD is
$[0.60 (1.01) 1.30]$.

In the PSR domain version with middle-size instances and derived
predicates compiled into ADL, i.e., in domain version {\em
  middle-compiled}, only Macro-FF and SGPlan participated.  Macro-FF
scaled relatively well, solving 32 instances up to instance number 48.
SGPlan solved only 14 instances up to instance number 19. Both
planners try to minimize the number of actions, and the ratio SGPlan
vs Macro-FF is $[0.51 (1.28) 1.91]$.

It is interesting to observe that, in PSR, when compiling derived
predicates away, the plan length increases much more than what we have
seen in Section~\ref{results:promela} for Dining-Philosophers and
Optical-Telegraph. In the latter, the derived predicates just replace
two action applications per process, detecting that the respective
process is blocked. In PSR, however, as said the derived predicates
model the flow of electricity through the network, i.e., they encode
the transitive closure of the underlying graph. The number of
additional actions needed to simulate the latter grows, of course,
with the size of the graph. This phenomenon can be observed in the
IPC-4 plan length data. The PSR {\em middle-compiled} instances are
identical to the {\em middle} instances, except for the compilation of
derived predicates. Comparing the plan quality of Macro-FF in {\em
  middle-compiled} with that of SGPlan in {\em middle}, we obtain the
remarkably high ratio values $[7.17 (12.53) 25.71]$. The maximum case
is in instance number 48 -- the largest instance solved by Macro-FF --
where Macro-FF takes 180 steps to solve the compiled instance, while
SGPlan takes only 7 steps to solve the original
instance.\footnote{While Macro-FF is not an optimal planner, and the
  optimal plan lengths for the {\em middle-compiled} instances are not
  known, it seems highly unlikely that these observations are only due
  to overlong plans found by Macro-FF.}

In domain version {\em small}, we awarded 1st places to FDD (and FD),
and SATPLAN; we awarded 2nd places to LPG-TD, SGPlan, YAHSP, and
BFHSP. In domain version {\em middle}, we awarded a 1st place to FDD
(and FD); we awarded 2nd places to LPG-TD and SGPlan. In domain
version {\em middle-compiled}, we awarded a 1st place to Macro-FF. In
domain version {\em large}, we awarded a 1st place to FDD (and FD).

\subsection{Satellite}
\label{results:satellite}

The {\em Satellite} domain was introduced in IPC-3 by
\citeA{long:fox:jair-03}. It is motivated by a NASA space application:
a number of satellites has to take images of a number of spatial
phenomena, obeying constraints such as data storage space and fuel
usage. In IPC-3, there were the domain versions {\em Strips}, {\em
  Numeric}, {\em Time} (action durations are expressions in static
variables), and {\em Complex} (durations {\em and} numerics, i.e. the
``union'' of Numeric and Time). The numeric variables transport the
more complex problem constraints, regarding data capacity and fuel
usage.

For IPC-4, the domain was made a little more realistic by additionally
introducing time windows for the sending of the image data to earth,
i.e. to antennas that are visible for satellites only during certain
periods of time. We added the new domain versions {\em
  Time-timewindows}, {\em Time-timewindows-compiled}, {\em
  Complex-timewindows}, and {\em Complex-timewindows-compiled}, by
introducing time windows, explicit and compiled, into the IPC-3 {\em
  Time} and {\em Complex} versions, respectively.\footnote{In the new
  domain versions derived from Complex, we also introduced utilities
  for the time window inside which an image is sent to earth. For each
  image, the utility is either the same for all windows, or it
  decreases monotonically with the start time of the window, or it is
  random within a certain interval. Each image was put randomly into
  one of these classes, and the optimization requirement is to
  minimize a linear combination of makespan, fuel usage, and summed up
  negated image utility.} None of the domain versions uses ADL
constructs, so of all versions there was only a single (STRIPS)
formulation. The instances were, as much as possible, taken from the
original IPC-3 instance suites. Precisely, the Strips, Numeric, Time,
and Complex versions each contained the 20 instances posed in IPC-3 to
the fully-automatic planners, plus the 16 instances posed in IPC-3 to
the hand-tailored planners. For Time-timewindows and
Time-timewindows-compiled, we extended the 36 instances from Time with
the time windows. Similarly, for Complex-timewindows and
Complex-timewindows-compiled, we extended the 36 instances from
Complex with the time windows.  That is, in the newly added domain
versions the sole difference to the previous instances lies in the
time windows.

\begin{figure}[t]
\begin{center}
\vspace{-0.0cm}
\begin{tabular}{cc}
\includegraphics[width=7.3cm]
{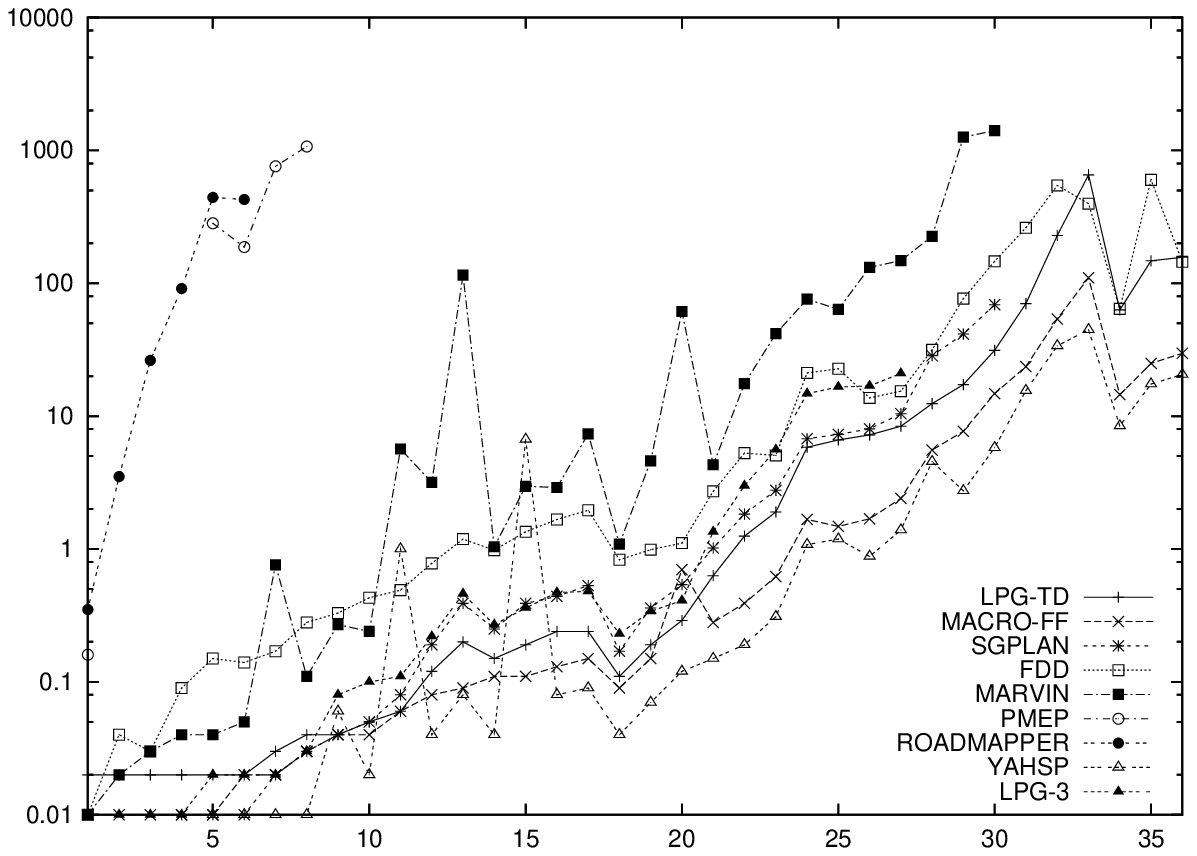} & 
\includegraphics[width=7.3cm]
{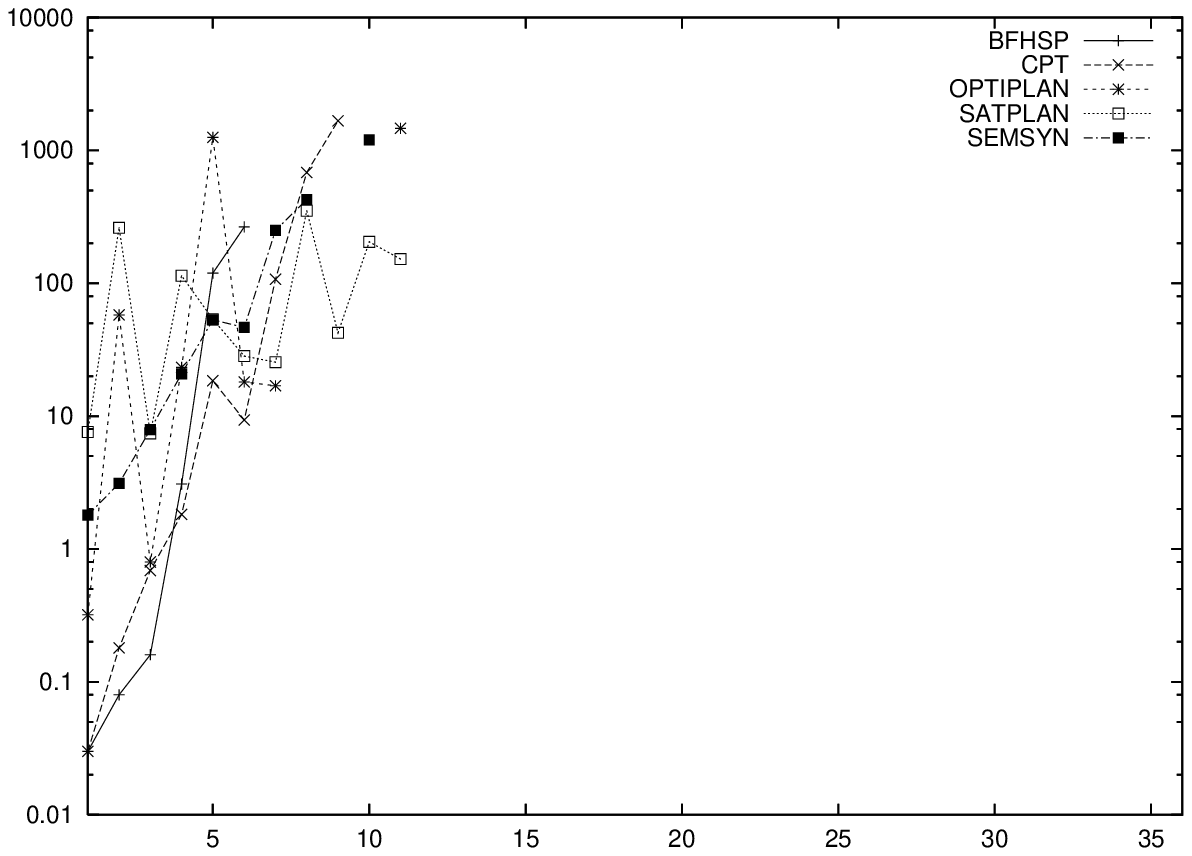} \\
(a) & (b) \\
\includegraphics[width=7.3cm]
{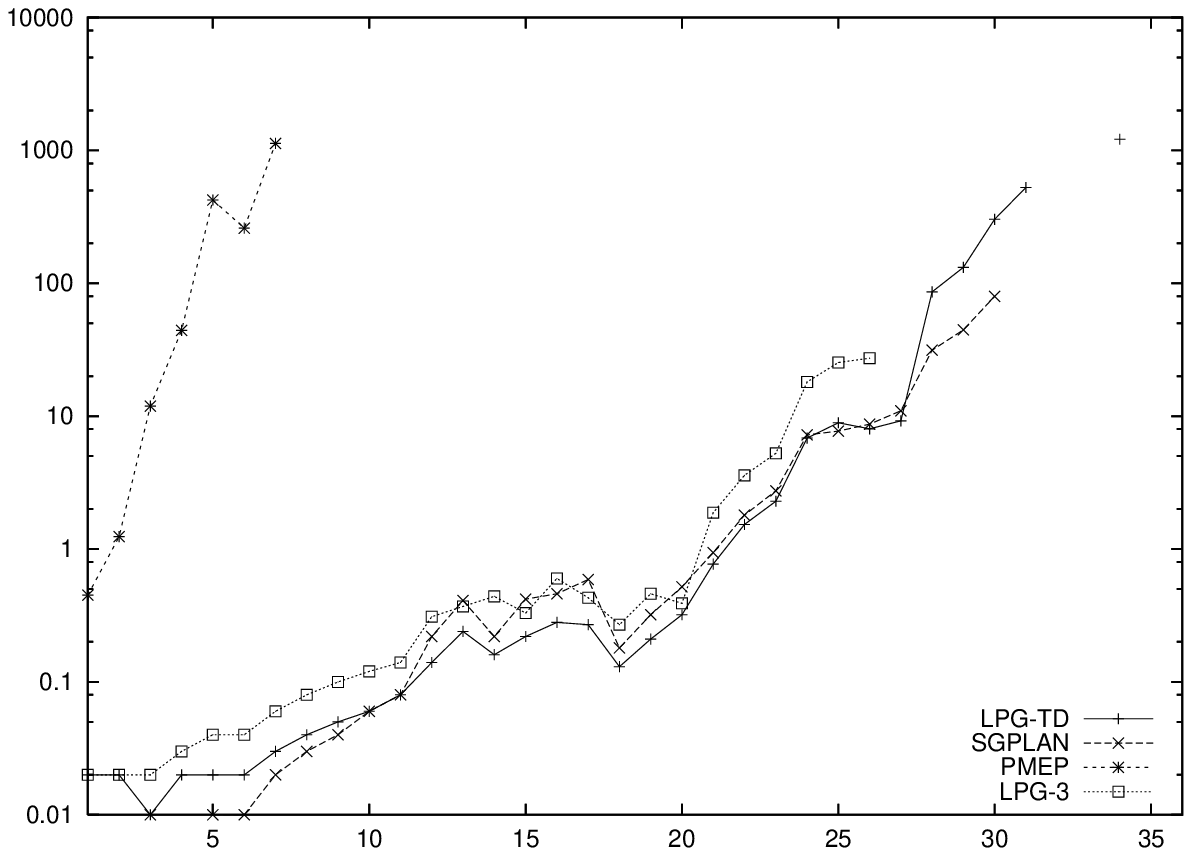} &
\includegraphics[width=7.3cm]
{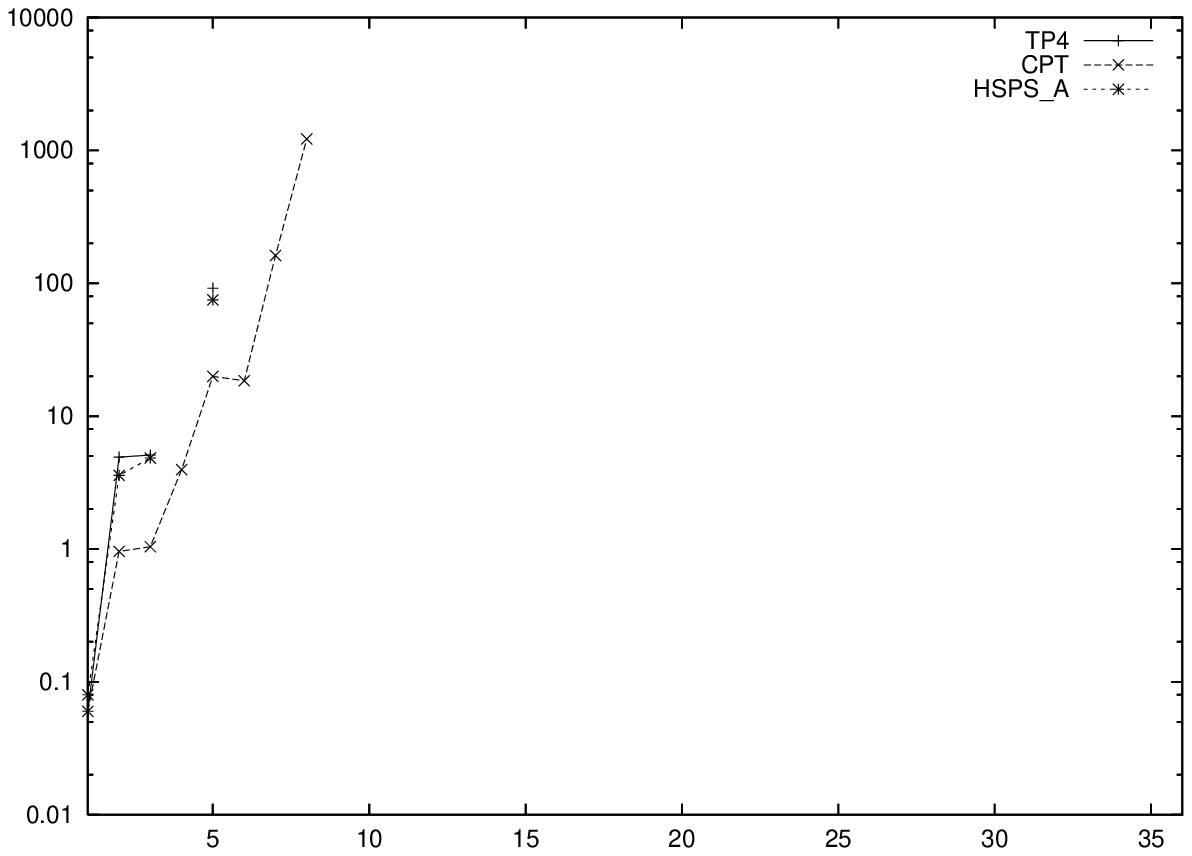} \\
(c) & (d) 
\end{tabular}
\caption{Satellite; Strips satisficing (a), optimal (b), and Time satisficing (c), optimal (d).}
\label{sat:nontandt}
\end{center}
\vspace{-0.0cm}
\end{figure}

Let us first consider the results in the simpler versions of the
Satellite domain. Figure~\ref{sat:nontandt} shows the results for the
Strips and Time versions. To many of the satisficing planners, the
Strips version does not pose a serious problem, see
Figure~\ref{sat:nontandt} (a). Macro-FF and YAHSP show the best
runtime behavior, followed closely by LPG-TD and FDD (as well as FD).
As for the optimal planners in Figure~\ref{sat:nontandt} (b), none of
them scales up very far. SATPLAN is most efficient, solving 11
instances; CPT and Semsyn each solve 9 instances, Optiplan solves 8,
BFHSP solves only 6. In the Time version, LPG-TD and SGPlan behave
similarly up to instance number 30, but LPG-TD solves two more,
larger, instances. The Time optimal planners, see
Figure~\ref{sat:nontandt} (d), are clearly headed by CPT.

Regarding the efficiency of the satisficing planners in the Strips
test suite, we remark that these instances were solved quite
efficiently at IPC-3 already, e.g. by FF and LPG-3. Further,
\citeA{hoffmann:jair-05} shows that, in this domain, under relaxed
plan distance, from any non-goal state, one can reach a state with
strictly smaller heuristic value within at most 5 steps. In that
sense, the results depicted in Figure~\ref{sat:nontandt} (a) didn't
come as a surprise to us.

\begin{sloppypar}
Let us consider plan quality in Strips and Time. In both domain
versions, as before the addressed optimization criteria are plan
length, and makespan. In Strips, the plan quality behavior of all the
planners trying to minimize plan length is rather similar. The overall
shortest plans are found by LPG-TD.quality. Precisely, the ratio FDD
vs LPG-TD.quality is $[0.96 (1.19) 1.48]$; the ratio FD vs
LPG-TD.quality is $[0.96 (1.20) 1.53]$; the ratio LPG-TD.speed vs
LPG-TD.quality is $[1.00 (1.12) 1.31]$; the ratio Macro-FF vs
LPG-TD.quality is $[0.95 (1.03) 1.17]$; the ratio Roadmapper vs
LPG-TD.quality is $[1.00 (1.35) 1.71]$; the ratio SGPlan vs
LPG-TD.quality is $[0.99 (1.07) 1.70]$; the ratio YAHSP vs
LPG-TD.quality is $[0.97 (1.22) 1.93]$. Of the planners trying to
optimize makespan in the Strips version, only Marvin and P-MEP are
satisficing; Marvin is the only planner that scales to the larger
instances. The ratio P-MEP vs SATPLAN (which solves a superset of the
instances solved by P-MEP) is $[1.01 (2.22) 3.03]$. The ratio Marvin
vs SATPLAN is $[1.08 (2.38) 4.00]$.
\end{sloppypar}

In the Time version, the only planner minimizing plan length (i.e.,
the number of actions) is SGPlan, so there is no basis for comparison.
The only satisficing planners trying to minimize makespan are P-MEP
and LPG-TD. In the small instances solved by CPT (actually, a superset
of those solved by P-MEP), the ratio P-MEP vs CPT is $[1.10 (3.71)
6.49]$.  The ratio LPG-TD.speed vs CPT is $[1.33 (3.37) 5.90]$; the
ratio LPG-TD.quality vs CPT is $[0.85 (1.24) 1.86]$. Note that, like
we have seen in Pipesworld before, sometimes LPG-TD finds better plans
here than the optimal CPT. As said, this is due to the somewhat
simpler model of durative actions that CPT uses
\cite{vidal:geffner:aaai-04}, making no distinction between the start
and end time points of actions.

\begin{figure}[t]
\vspace{-0.0cm}
\begin{center}
\begin{tabular}{cc}
\includegraphics[width=7.3cm] {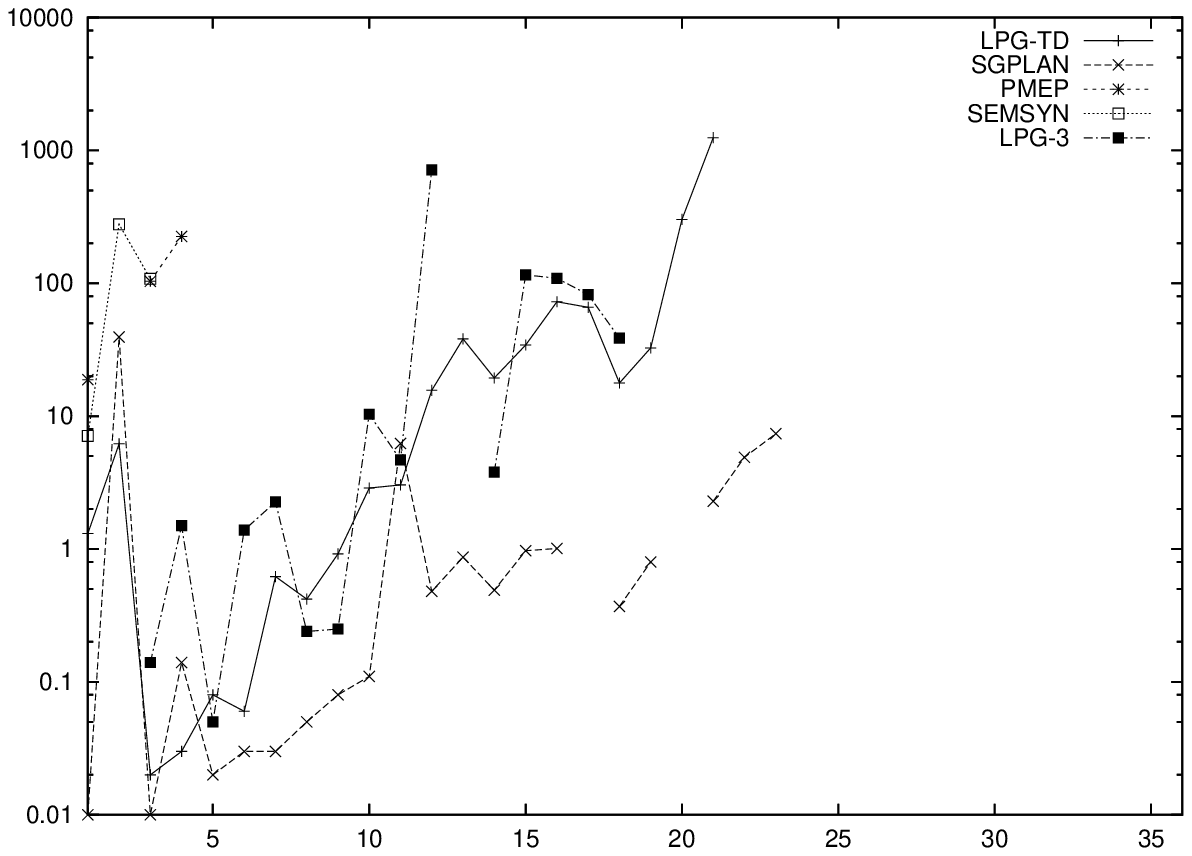} &
\includegraphics[width=7.3cm]{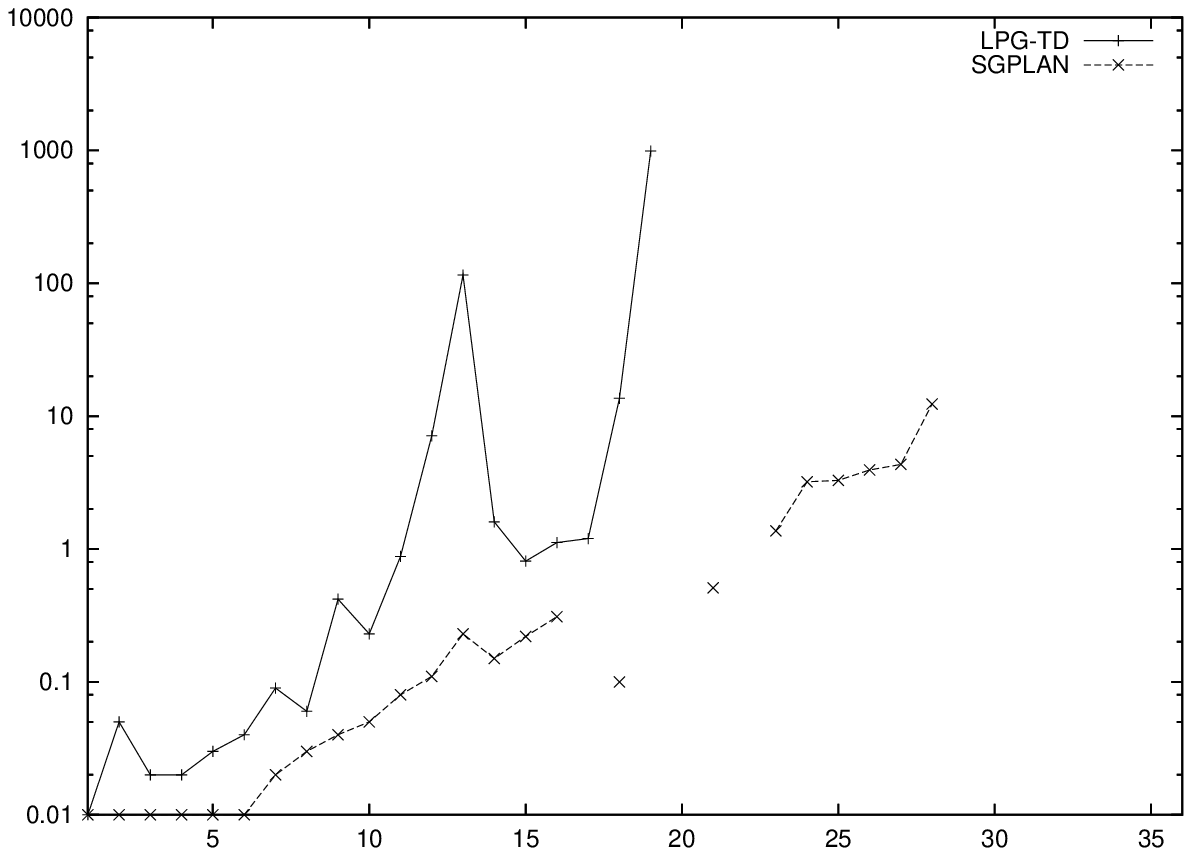} \\
(a) & (b) \\
\includegraphics[width=7.3cm]{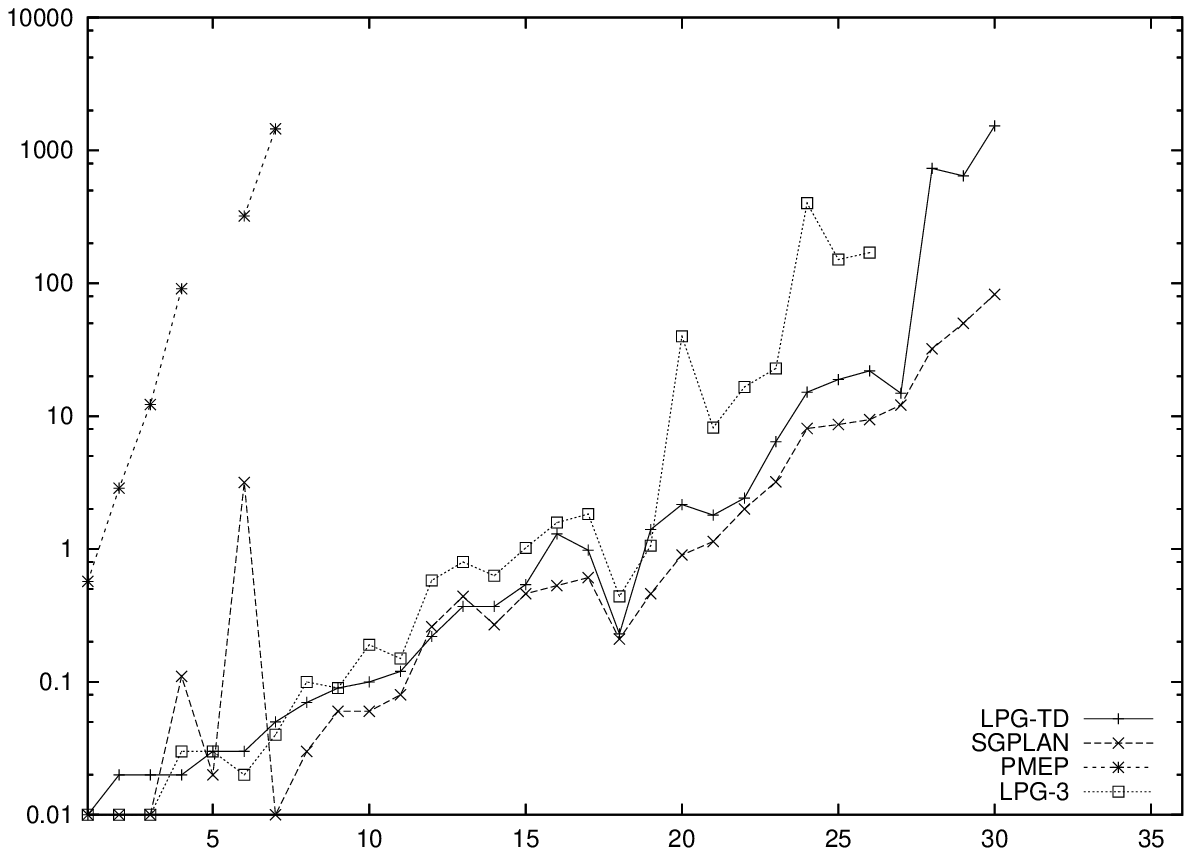} &
\includegraphics[width=7.3cm]{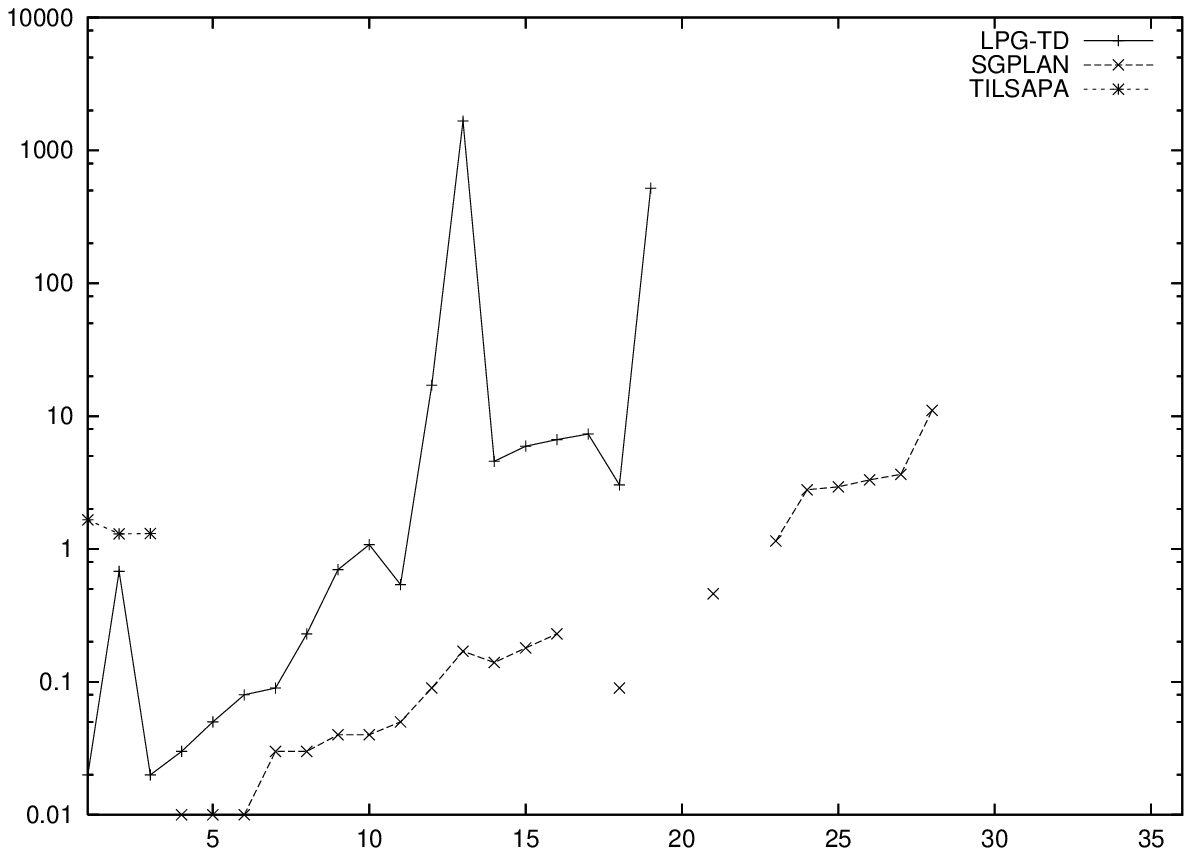} \\
(c) & (d)
\end{tabular}
\caption{Satellite; Numeric (a), Time-timewindows (b), satisficing Complex (c), Complex-timewindows (d).}
\label{sat:ntwc}
\end{center}
\vspace{-0.0cm}
\end{figure}

Figure~\ref{sat:ntwc} (a) shows the results in Satellite Numeric,
together for all participating planners -- the only optimal planner
that participated here was Semsyn. SGPlan and LPG-TD scale best; they
solve the same number of instances, but SGPlan solves some larger ones
and is at least an order of magnitude faster in those instances solved
by both. Regarding plan quality, SGPlan was the only planner here
trying to minimize plan length. The other planners tried to minimize
the metric value of the plan, i.e., the quality metric specified in
the instance files, which is fuel usage. The best plans in this
respect are found by LPG-TD.quality; it is unclear to us why Semsyn
was marked to optimize the metric value here, and why it does not find
the optimal plans. The ratio LPG-TD.speed vs LPG-TD.quality is $[1.00
(2.26) 4.69]$; the ratio P-MEP vs LPG-TD.quality is $[1.01 (2.16)
3.03]$; the ratio Semsyn vs LPG-TD.quality is $[1.00 (1.28) 1.58]$.

In Time-timewindows, no optimal planner could participate due to the
timed initial literals. Of the satisficing planners, see
Figure~\ref{sat:ntwc} (b), only SGPlan and LPG-TD participated, which
both scaled relatively well, with a clear advantage for SGPlan.
Regarding plan length, SGPlan minimizes the number of actions, and
LPG-TD the makespan. The ratio LPG-TD.speed vs LPG-TD.quality is
$[1.00 (1.22) 1.51]$. In Time-timewindows-compiled, only SGPlan and
CPT participated. SGPlan largely maintained its performance from the
Time-timewindows version, CPT could solve only 3 of the smallest
instances. Plan quality can't be compared due to the different
criteria (plan length and makespan) that are minimized.

Figure~\ref{sat:ntwc} (c) shows the performance of the satisficing
planners in Satellite Complex. SGPlan and LPG-TD scale well, in
particular much better than the only other competitor, P-MEP. In the
largest three instances, SGPlan shows a clear runtime advantage over
LPG-TD. In the optimal track, only TP4 and HSP$_a^*$ competed here.
Both solved the same four very small instances, numbers 1, 2, 3, and
5, in almost the same runtime. SGPlan is the only planner minimizing
the number of actions. Of the other planners, which all try to
minimize makespan, only LPG-TD scales up to large instances. The ratio
LPG-TD.speed vs LPG-TD.quality is $[1.01 (2.76) 4.71]$. The ratio
LPG-TD.speed vs TP4 is $[1.81 (2.09) 2.49]$; the ratio LPG-TD.quality
vs TP4 is $[0.93 (1.07) 1.19]$; the ratio P-MEP vs TP4 is $[1.27
(2.25) 3.32]$.  As above for CPT, the better plan (in one case) found
by LPG-TD.quality is due to the somewhat simpler action model used in
TP4 \cite{haslum:geffner:ecp-01}.

Figure~\ref{sat:ntwc} (d) shows the performance of all planners in
Complex-timewindows -- as above, due to the timed initial literals no
optimal planner could compete here. SGPlan scales clearly best,
followed by LPG-TD; Tilsapa solves only the 3 smallest instances. Of
these 3 participating planners, each one minimizes a different quality
criterion: number of actions for SGPlan, makespan for Tilsapa, and
metric value -- the aforementioned linear combination of makespan,
fuel usage, and summed up negated image utility -- for LPG-TD. So the
only useful comparison is that between LPG-TD.speed and
LPG-TD.quality, where the ratio is $[1.00 (2.33) 7.05]$.\footnote{Due
  to the negated image utility, the quality values here can be
  negative.  For LPG-TD.speed and LPG-TD.quality here, this happened
  in 5 instances. We skipped these when computing the given ratio
  values, since putting positive and negative values together doesn't
  make much sense -- with negative quality, the number with larger
  absolute value represents the better plan. It would probably have
  been better to define, in this domain version, an image penalty
  instead of an image utility, and thus obtain strictly positive
  action ``costs''.}  In Complex-timewindows-compiled, the only
participating planner was SGPlan. It maintained its good scalability
from the Complex domain version with explicit time windows.

In domain version Strips, we awarded 1st places to Macro-FF, YAHSP,
and SATPLAN; we awarded 2nd places to FDD (and FD), LPG-TD, CPT,
Optiplan, and Semsyn. In domain version Time, we awarded 1st places to
LPG-TD and CPT; we awarded a 2nd place to SGPlan. In each of the
domain versions Numeric, Time-timewindows, Complex, and
Complex-timewindows, we awarded a 1st place to SGPlan and a 2nd place
to LPG-TD.

\subsection{Settlers}
\label{results:settlers}

The {\em Settlers} domain was also introduced in IPC-3
\cite{long:fox:jair-03}. It features just a single domain version,
which makes extensive use of numeric variables. These variables carry
most of the domain semantics, which is about building up an
infrastructure in an unsettled area, involving the building of houses,
railway tracks, sawmills, etc.  In IPC-3, no planner was able to deal
with the domain in an efficient way -- the best IPC-3 planner in
Settlers, Metric-FF \cite{hoffmann:jair-03}, solved only the smallest
six instances of the test suite. For these reasons, we included the
domain into IPC-4 as a challenge for the numeric planners. We used the
exact same domain file and example instances as in IPC-3, except that
we compiled away some universally quantified preconditions to improve
accessibility for planners. The quantifiers were not nested, and
ranged over a fixed set of domain constants, so they could easily be
replaced by conjunctions of atoms.

\begin{figure}[t]
\begin{center}
\includegraphics[width=7.3cm]{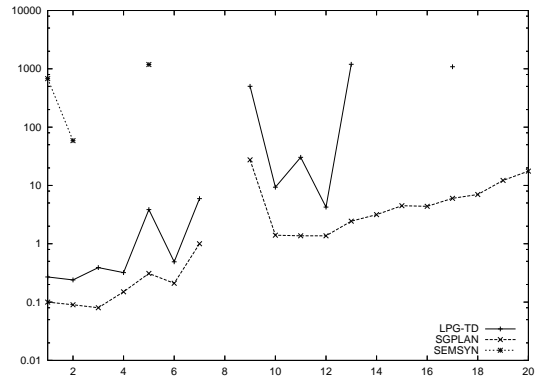}
\end{center}
\caption{Performance of all planners in Settlers.}
\label{settlers:all}
\end{figure}

Figure~\ref{settlers:all} shows the runtime results obtained in IPC-4.
The efficiency increase compared to IPC-3 is, obviously, dramatic:
SGPlan solves every instance within less than 15 seconds (instance
number 8 is unsolvable).\footnote{We remark that instance $8$ is
  trivially unsolvable. One of the goals can only be achieved by
  actions having a static precondition that is false in the initial
  state.  This can be detected by simple reachability analyses like,
  e.g., planning graphs, even without taking account of delete lists.}
LPG-TD solves 13 instances out of the set, Semsyn solves 3 (numbers
1,2, and 5). We remark that the solutions for these tasks, as returned
by SGPlan, are huge, up to more than 800 actions in the largest tasks.
On the one hand, this once again demonstrates SGPlan's capability to
find extremely long plans extremely quickly. On the other hand,
SGPlan's plans in Settlers might be unnecessarily long, to some
extent. In the largest instance solved by Semsyn, number 5, Semsyn
finds a plan with 94 actions, while the plan found by SGPlan has 264
actions. In the largest instance solved by LPG-TD, number 17,
LPG-TD.quality takes 473 actions while SGPlan takes 552. LPG-TD
minimizes the metric value of the plans, a linear combination of the
invested labor, the resource use, and the caused pollution. The ratio
LPG-TD.speed vs LPG-TD.quality is $[1.00 (1.21) 3.50]$.

We awarded a 1st place to SGPlan and a 2nd place to LPG-TD.

% Figure~\ref{settlers:all} shows the runtime results obtained in IPC-4.
% The efficiency increase compared to IPC-3 is, obviously, dramatic:
% SGPlan solves every instance within less than 15 seconds (instance
% number 8 is unsolvable). We awarded a 1st place to SGPlan and a 2nd
% place to LPG-TD. We remark that the solutions for these tasks, as
% returned by SGPlan, are huge, up to more than 800 actions in the
% largest tasks. On the one hand, this once again demonstrates SGPlan's
% capability to find extremely long plans extremely quickly. On the
% other hand, SGPlan's plans in Settlers are probably overly long. In
% those instances solved by both LPG-TD and SGPlan, SGPlan's plans are
% around 20\% longer.

\subsection{UMTS}
\label{results:umts}

UMTS applications require comparatively much time to be started on a
hand-held device, since they have to communicate with the network
several times. If there is more than one application that has been
called, this yields a true bottleneck for the user.  Therefore, the
applications set-up is divided into several parts to allow different
set-up modules to work concurrently.  The task in the domain is to
provide a good schedule -- minimizing the used time -- for setting up
timed applications, respecting the dependencies among them.

In IPC-4, there were the six domain versions {\em UMTS}, {\em
  UMTS-timewindows}, {\em UMTS-timewindows-compiled}, {\em UMTS-flaw},
{\em UMTS-flaw-timewindows}, and {\em UMTS-flaw-timewindows-compiled}.
All of these domain versions are temporal, and make use of numeric
variables to model the properties of the applications to be set up.
ADL constructs are not used. UMTS is our standard model of the domain.
In UMTS-timewindows there are additional time windows regarding the
executability of set up actions, encoded with timed initial literals.
In UMTS-timewindows-compiled the same time windows are compiled into
artificial constructs. The remaining three domain versions result, as
their names suggest, from adding a {\em flaw} construction to their
respective counterparts. The flaw construction is practically
motivated. It consists of an extra action that has an important
sub-goal as its add effect, but that deletes another fact that can't
be re-achieved. So the flaw action can't be used in a plan, but it can
be used in relaxed plans, i.e. when ignoring the delete effects.  In
particular, adding an important sub-goal, the flawed action provides a
kind of ``short-cut'' for relaxed plans, where the short-cut does not
work in reality. This can lead to overly optimistic heuristic values.
Thus the flaw may confuse the heuristic functions of relaxed-plan
based heuristic planners. We used the flawed domain versions in IPC-4
to see whether the latter would be the case.

In the IPC-4 test suites, all instances, irrespective of their
number/size, contain 10 applications. The main scaling parameter is
the number of applications that must actually be set up. For IPC-4
instances number 1 \dots 5, a single application must be set up; for
instances number 6 \dots 10, two applications must be set up, and so
on, i.e., the number of needed applications is $\lfloor x/5 \rfloor$
where $x$ is the index of the instance.

\begin{figure}[t]
\begin{center}
\vspace{-0.0cm}
\begin{tabular}{cc}
\includegraphics[width=7.3cm]{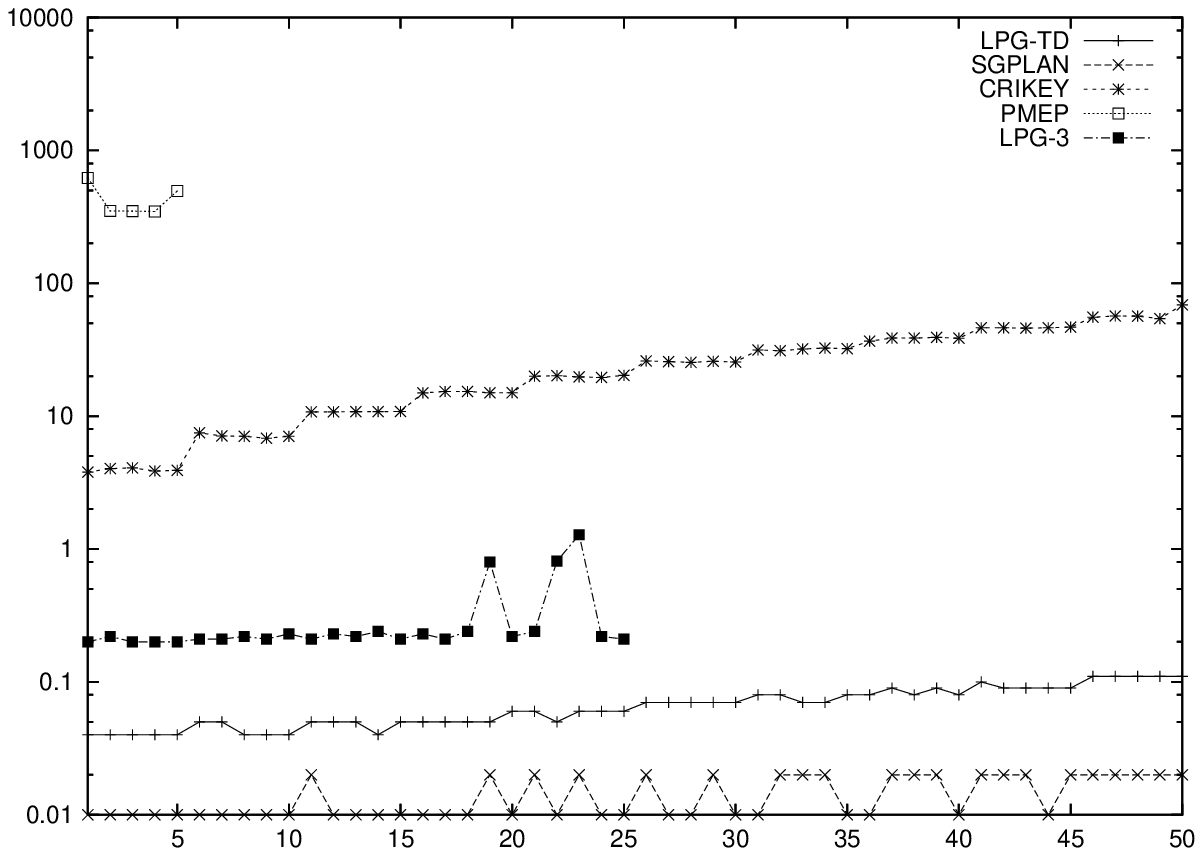} &
\includegraphics[width=7.3cm]{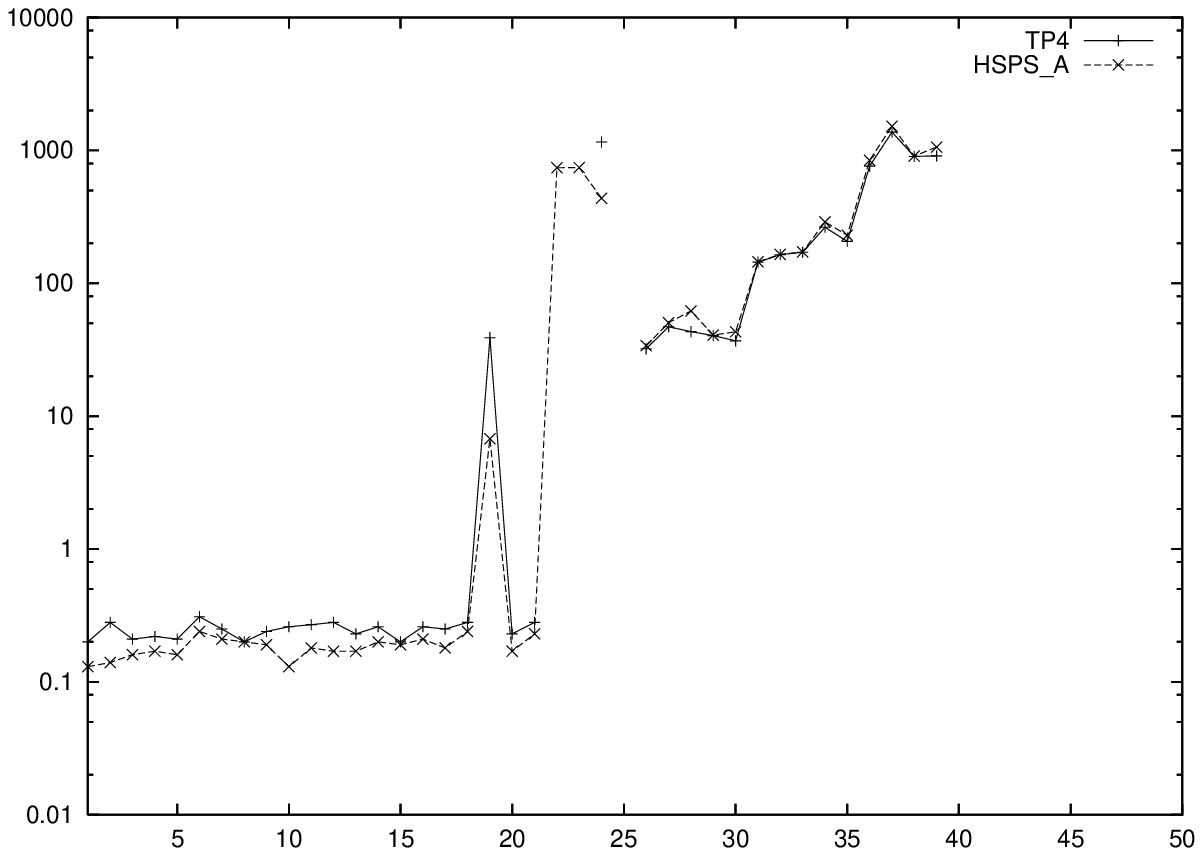} \\
(a) & (b) \\
\includegraphics[width=7.3cm]{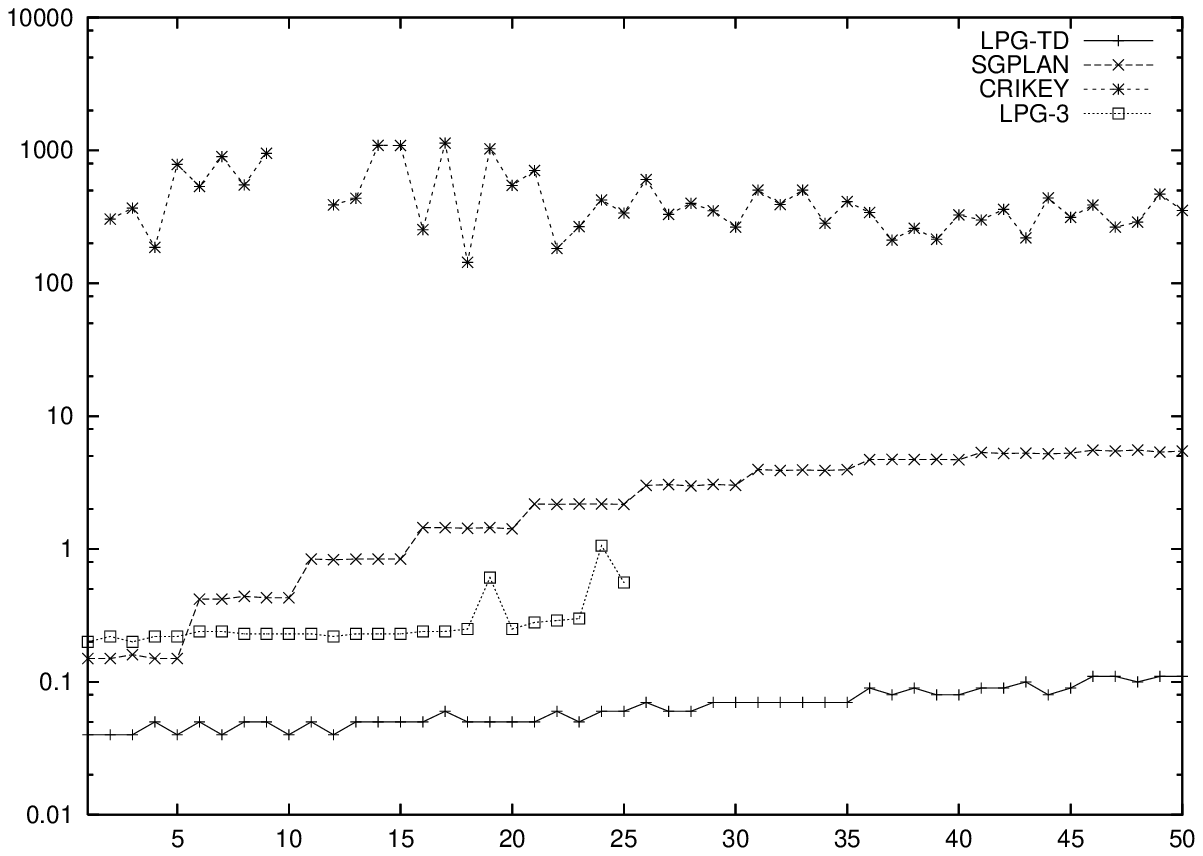} &
\includegraphics[width=7.3cm]{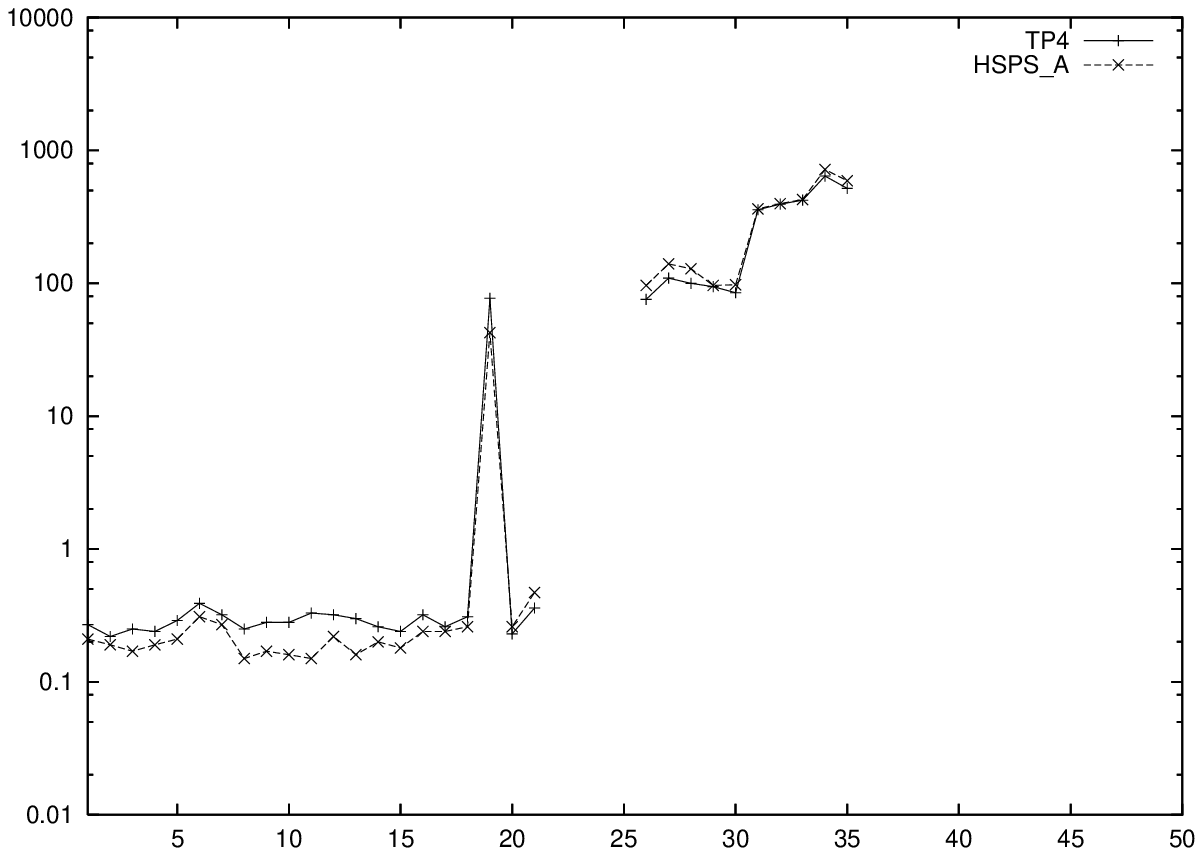}\\
(c) & (d)
\end{tabular}
\vspace{-0.0cm}
\end{center}
\caption{UMTS, satisficing (a) optimal (b); UMTS-flaw, satisficing (c) optimal (d).}
\label{umts:notw}
\end{figure}

Figure~\ref{umts:notw} (a) and (b) shows the IPC-4 performance in the
basic domain version. Obviously, SGPlan and LPG-TD have no difficulty
at all with the domain -- they solve every instance within split
seconds. CRIKEY takes more time, but also scales up nicely. As for the
optimal planners, only TP4 and HSP$_a^*$ were able to handle the
domain syntax, due to the combination of numeric variables and action
durations. As Figure~\ref{umts:notw} (b) shows they scaled relatively
similarly, with a slight advantage for HSP$_a^*$. Note that the two
planners {\em scaled better than P-MEP and LPG-3 in the satisficing
  track.}

We remark at this point that UMTS is mainly intended as a benchmark
for {\em optimal} planners minimizing makespan. The domain is a pure
scheduling problem by nature. As in most scheduling problems, it is
trivial to find {\em some} plan -- for example, one can simply
schedule all applications in sequence.\footnote{Note that this is not
  possible in the Airport domain -- that can also be viewed as a type
  of scheduling problem \cite{hatzack:nebel:ecp-01} -- due to the
  restricted space on the airport. In that sense, the Airport domain
  incorporates more planning aspects than UMTS.} The only point of the
domain, and, indeed, of using a computer to solve it, is to provide
{\em good} schedules, that is, schedules with the smallest possible
execution time, corresponding to the makespan of the plan.

That said, we observe that the satisficing planners in IPC-4 were
quite good at finding near-optimal plans in UMTS. In fact, as we will
see in detail in the following, the only planner finding highly
non-optimal plans was LPG-3.  Let's consider the basic domain version
treated in Figure~\ref{umts:notw} (a) and (b). Two of the
participating planners, CRIKEY and SGPlan, try to minimize the number
of actions (i.e., the wrong optimization criterion). Their data is
identical, i.e. the plan ``lengths'' are the same for all instances.
Precisely, both planners find, for each instance number $x$, a plan
with $\lfloor x/5 \rfloor*8$ actions in it. This is, in fact, the
optimal (smallest possible) number of actions -- remember what we said
above about the scaling in the IPC-4 test suites. The participating
planners trying to minimize makespan are LPG-3, LPG-TD, P-MEP,
HSP$_a^*$, and TP4. P-MEP solves only the smallest 5 instances,
finding the optimal plans. The ratio LPG-TD.speed vs HSP$_a^*$ is
$[1.00 (1.02) 1.11]$. The ratio LPG-TD.quality vs HSP$_a^*$ is $[1.00 (1.00)
1.03]$. The ratio LPG-3 vs HSP$_a^*$ is $[1.00 (1.53) 2.27]$; we remark that
the plan quality data used here is for ``LPG-3.bestquality'', which
takes the entire available half hour time trying to optimize the found
plan. Looking at the plots, one sees that the LPG-3 makespan curve is
much steeper than those of the other planners.

In Figure~\ref{umts:notw} (c) and (d), we see the performance of the
IPC-4 planners in the same basic domain version, but with the flaw
construct. LPG-TD remains unaffected, but SGPlan and CRIKEY become a
lot worse. Particularly, CRIKEY now takes several minutes to solve
even the smallest instances. We take this to confirm our intuition
that the flaw can (but does not necessarily) confuse the heuristic
functions of relaxed-plan based heuristic planners. Whether or not the
heuristic function becomes confused probably depends on details in the
particular way the relaxed plans are constructed (such a construction
may, e.g., choose between different actions based on an estimate of
how harmful they are to already selected actions). In the optimal
track, when introducing the flaw construct into the domain, the
performance of TP4 and HSP$_a^*$ becomes slightly worse, and nearly
indistinguishable.  Regarding plan quality, there now is a quality
difference in terms of the number of actions needed by CRIKEY and
SGPlan. SGPlan's plans still all have the smallest possible
``length'', $\lfloor x/5 \rfloor*8$. Those of CRIKEY have a lot of
variance, and are longer; the ratio CRIKEY vs SGPlan is $[1.07 (1.72)
3.88]$.  In the group minimizing makespan, the observations are very
similar to the unflawed domain version above.  The ratio LPG-TD.speed
vs HSP$_a^*$ is $[1.00 (1.01) 1.04]$, and the ratio LPG-TD.quality vs
HSP$_a^*$ is $[1.00 (1.00) 1.01]$. Although it solves only the smaller
half of the instances, the ratio LPG-3.bestquality vs HSP$_a^*$ is
$[1.00 (1.48) 2.27]$.

\begin{figure}[t]
\begin{center}
\vspace{-0.0cm}
\begin{tabular}{cc}
\includegraphics[width=7.3cm]{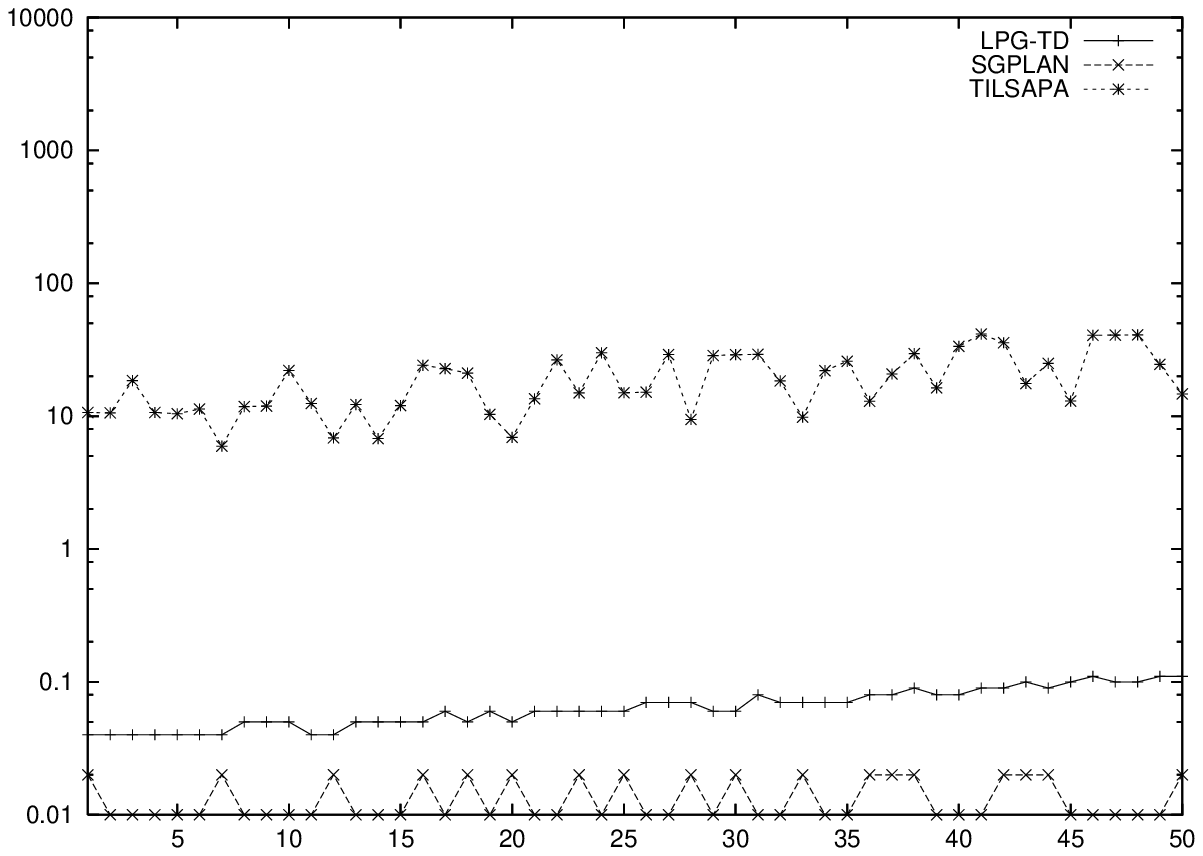} &
\includegraphics[width=7.3cm]{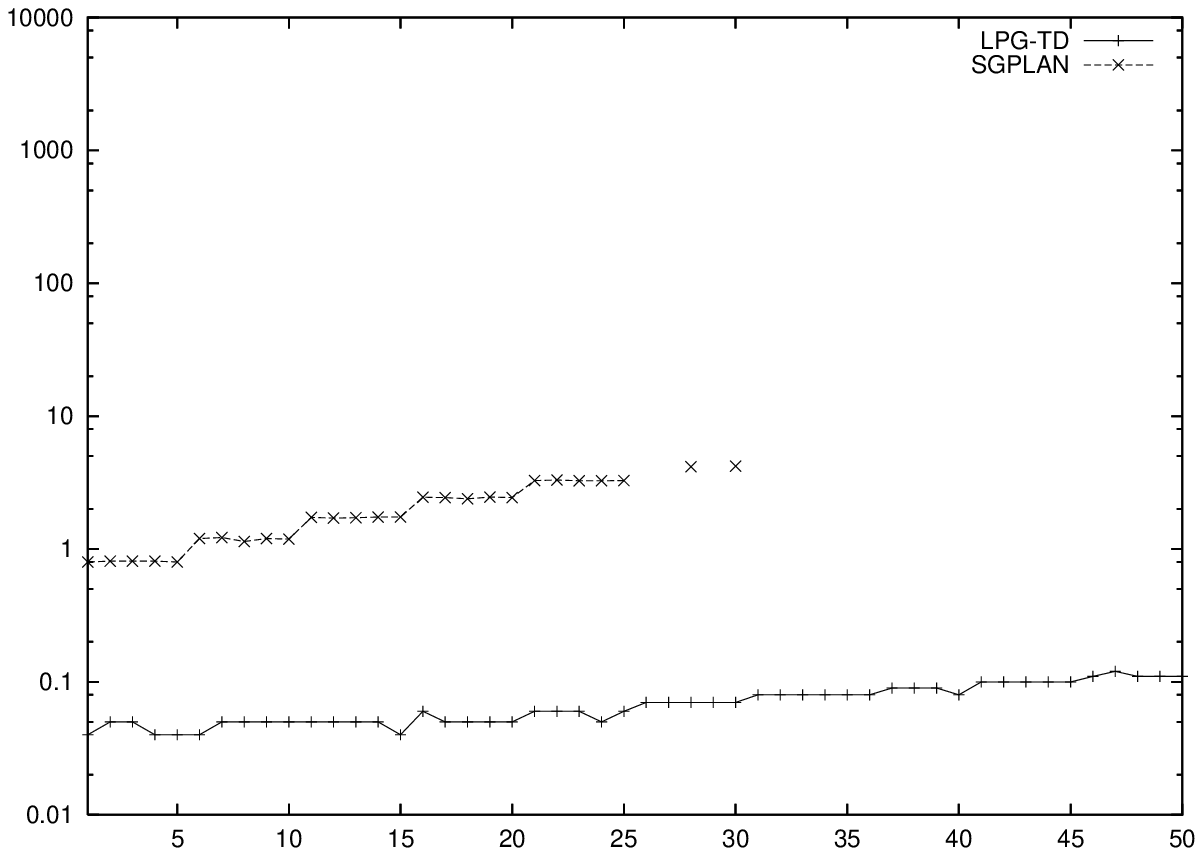} \\
(a) & (b)
\end{tabular}
\vspace{-0.0cm}
\end{center}
\caption{UMTS-timewindows (a); UMTS-flaw-timewindows (b).}
\label{umts:tw}
\end{figure}

Figure~\ref{umts:tw} shows the results in the two domain versions with
explicitly encoded time windows. No optimal planners could participate
since none of them could handle timed initial literals. In the version
without the flaw, again LPG-TD and SGPlan need only split seconds. The
only other competitor, Tilsapa, needs more runtime but also scales up
well. When introducing the flaw, the only competitors are SGPlan and
LPG-TD.  SGPlan's runtime performance becomes a lot worse, while
LPG-TD remains completely unaffected. Regarding plan quality, the
number of actions in SGPlan's plans is still $\lfloor x/5 \rfloor*8$
in all cases.  LPG-TD.speed and LPG-TD.quality return plans with
identical makespan in all cases. In the non-flawed version,
UMTS-timewindows, the makespan ratio Tilsapa vs LPG-TD is $[1.00 (1.20)
1.37]$.

We do not show runtime graphs for the domain versions with compiled
time windows. In UMTS-timewindows-compiled, SGPlan and CRIKEY
participated. Both scale up well and solve all instances, but SGPlan
needs only split seconds, while CRIKEY needs up to more than 100
seconds per instance in the larger cases. Both planners try to
minimize the number of actions, and both need exactly $\lfloor x/5
\rfloor*8+5$ actions for each instance number $x$ -- namely, the
optimal number $\lfloor x/5 \rfloor*8$ of actions as before, plus $5$
artificial actions encoding the time windows. In
UMTS-flaw-timewindows-compiled, the sole participating planner was
SGPlan. It solved only the smaller half of the instances, finding
plans of length $\lfloor x/5 \rfloor*8+6$ -- i.e., using one
unnecessary action in each instance (the action concerns an
application that does not need to be set up).

In domain version UMTS, we awarded 1st places to SGPlan, LPG-TD, and
HSP$_a^*$; we awarded 2nd places to CRIKEY and TP4. In domain version
UMTS-flaw, we awarded a 1st place to LPG-TD, and 2nd places to SGPlan,
HSP$_a^*$, and TP4. In version UMTS-timewindows, we awarded 1st places
to SGPlan and LPG-TD, and a 2nd place to Tilsapa. In
UMTS-flaw-timewindows, we awarded a 1st place to LPG-TD. In
UMTS-timewindows-compiled, we awarded a 1st place to SGPlan.

\section{IPC-4 Awards}
\label{awards}

The numbers of 1st and 2nd places achieved by the planners are shown
in Tables~\ref{awards-satisficing} and~\ref{awards-optimal}; planners
that never came in 1st or 2nd place are left out of the tables. 
Since many of the planners (6 of the satisficing planners, and 4 of
the optimal planners) only dealt with the purely propositional domain
versions (i.e., STRIPS or ADL), we counted the performance in these
domains separately.

\begin{table}[htb]
\begin{tabular}{lrrrrrrrr}
&
SGPlan &
LPG-TD &
FD &
FDD &
Macro-FF & 
YAHSP & 
CRIKEY &
Tilsapa \\
Propositional 
& 3 / 6
&1 / 6
&5 / 2
&6 / 3
&3 / 0
&4 / 1
&0 / 0 
&0 / 0\\
Temp/Metric
&13 / 2
&7 / 7
&
&
&
&
& 0 / 1
& 0 / 1\\ \hline

Total Count
&16 / 8
&8 / 13
&5 / 2
&6 / 3
&3 / 0
&4 / 1
&0 / 1
&0 / 1
\end{tabular}
\caption{\label{awards-satisficing}Summary of results: satisficing planners, number 
of 1st places / number of 2nd places.}
\end{table}

\begin{table}[htb]
\begin{tabular}{lrrrrrrrr}
&CPT 
&TP-4
&HSP$_a^*$
&SATPLAN
&Optiplan
&Semsyn
&BFHSP \\
Propositional 
&0 / 1
&0 / 0
&0 / 0
&5 / 1
&0 / 4
&0 / 2
&0 / 2 \\

Temp/Metric 
&3 / 0
&0 / 5
&1 / 2
&&&& \\ \hline

Total Count 
&3 / 1
&0 / 5
&1 / 2
&5 / 1
&0 / 4
&0 / 2
&0 / 2
\end{tabular}
\caption{\label{awards-optimal}Summary of results: optimal planners, number 
of 1st places / number of 2nd places.}
\end{table}

For the satisficing planners, based on our observations we decided to
award separate prizes for performance in the pure STRIPS and ADL
domains. For the optimal planners this seemed not appropriate due to,
first, the small number of planners competing in the temporal/metric
domains, and, second, the smaller overall number of competing systems
-- giving 4 prizes to 7 systems seemed too much.  Overall, the awards
made in the deterministic part of IPC-4 are the following:

\begin{itemize}
\item 1st Prize, Satisficing Propositional Track -- Fast (Diagonally) Downward, by Malte Helmert and Silvia Richter
\item 2nd Prize, Satisficing Propositional Track -- YAHSP, by Vincent Vidal
\item 2nd Prize, Satisficing Propositional Track -- SGPlan, by Yixin Chen, Chih-Wei Hsu and Benjamin W. Wah
\item 1st Prize, Satisficing Metric Temporal Track -- SGPlan, by Yixin Chen, Chih-Wei Hsu and Benjamin W. Wah
\item 2nd Prize, Satisficing Metric Temporal Track -- LPG-TD, by
  Alfonso Gerevini, Alessandro Saetti, Ivan Serina, and Paolo
  Toninelli
\item 1st Prize, Optimal Track -- SATPLAN, by Henry Kautz, David
  Roznyai, Farhad Teydaye-Saheli, Shane Neth, and Michael Lindmark
\item 2nd Prize, Optimal Track -- CPT, by Vincent Vidal and Hector Geffner 
\end{itemize}

We would like to re-iterate that the awarding of prizes is, and has to
be, a very sketchy ``summary'' of the results of a complex event such
as IPC-4. A few bits of information are just not sufficient to
summarize thousands of data points. Many of the decisions we took in
the awarding of the prizes, i.e. in the judgement of scaling behavior,
were very close. This holds especially true for most of the runtime
graphs concerning optimal planners, and for some of the runtime graphs
concerning satisficing propositional planners. What we think is best,
and what we encourage everybody to do, is to have a closer look at the
results plots for themselves. As said before, the full plots are
available in an online appendix.

%The above left aside, our congratulations to the awardees! And many
%thanks again to all the competitors, to the people who helped us
%creating the domains, and last but not least to the organizing
%committee who helped us set up the language PDDL2.2. All these people
%contributed to make IPC-4 such an exciting and fruitful event!!

\section{Conclusion}
\label{conclusion}

All in all, our feeling as the organizers is that the deterministic
part of IPC-4 was a great success, and made several valuable
contributions to the field. To mention the, from our perspective, two
most prominent points: the event provided the community with a set of
interesting new benchmarks, and made visible yet another major step
forward in the scalability of satisficing planning systems. The latter
was made possible by novel heuristics and domain analysis techniques.

There is a wide variety of questions to be addressed in the context of
the future of the (deterministic part of) the IPC. Let us discuss just
a few that we feel are important. Regarding benchmark domains, as said
we invested significant effort into the IPC-4 benchmarks, and we would
definitely recommend to re-use some of them in future IPC editions.
While some domain versions were solved relatively easily by certain
groups of planners, many of them still constitute major challenges.
Examples are Pipesworld with tankage restrictions, Optical-Telegraph,
large PSR instances, and UMTS for optimal planners.  That said, most
of the IPC-3 domains are also still challenging and should be re-used.
It is probably more useful for the community to consolidate
performance on the existing set of benchmarks, rather than to increase
the benchmark database at large pace. For one thing, to measure
progress between IPC editions, re-used benchmark domains (and
instances), like Satellite and Settlers in the case of IPC-4, are more
useful.\footnote{In the field addressing the SAT problem, particularly
  in the respective competition events, progress is also often
  measured simply in terms of the size (number of variables and
  clauses) of the formulas that could be tackled successfully, making
  just a few distinctions about the origin of the formulas (like,
  randomly generated or from an application). In the context of the
  IPC, one can do similar things by measuring parameters such as,
  e.g., the number of ground actions tackled successfully. However,
  given the large differences between the individual domains used in
  the IPC, more distinctions must be made. A detailed investigation of
  the effect of several parameters, in the IPC-4 domains, is given by
  \citeA{Benchmarks}.} More generally, the benchmark set is already
large; if there is a too large set, then a situation may arise where
authors select rather disjoint subsets in their individual
experiments.

Regarding PDDL extensions, the formalisms for derived predicates and
timed initial literals that we introduced are only first steps into
the respective directions. The PDDL2.2 derived predicates formalism is
restrictive in that it allows no negative interactions between derived
predicates; one could easily allow negative interactions that do not
lead to cycles and thus to ambiguous semantics
\cite{thiebaux:etal:ijcai-03,thiebaux:etal:ai-05}.  One could also
imagine derivation rules for values of more general data types than
predicates/Booleans, particularly for numeric variables. As for timed
initial literals, obviously they encode only a very restrictive subset
of the wide variety of forms that exogenous events can take. Apart
from exogenous events on numeric values, such events may in reality be
continuous processes conditioned on the world state, rather than
finitely many discrete time instants known beforehand. The PDDL2.2
action model itself is, of course, still restrictive in its
postulation of discrete variable value updates.  Still we believe that
the IPC should not let go off the very simple PDDL subsets such as
STRIPS, to support accessibility of the competition. As noted in
Section~\ref{systems:overview}, most systems are still not able to
handle all the language features introduced with IPC-3 and IPC-4.
Also, we believe that a STRIPS track, and more generally domain
versions formulated in simple language subsets, is important to
encourage basic algorithms research. A new idea is easier to try in a
simple language. To avoid misunderstandings: the language features
introduced with PDDL2.1 and PDDL2.2 already have a significant basis
of acceptance in implemented systems, and should definitely be kept
for future editions of the IPC.

In the context of basic research, it should be noted that the
satisficing track of IPC-4 was almost entirely populated by planners
of the relaxed-plan based heuristic search type. This demonstrates the
danger of the competition to concentrate research too much around a
few successful methods. Still, two of the most remarkable planners in
that track, Fast-Downward and SGPlan, use a significantly different
heuristic and new domain analysis techniques as the basis of their
success, respectively.

Putting the optimal planners in a separate track serves to maintain
this, very different, type of planning algorithms. We recommend to
keep this distinction in the future IPC events.

About the hand-tailored track, we have to say that it seems unclear if
and how that could be brought back into the focus. Maybe a more
suitable form of such an event would be an online (at the hosting
conference) programming (i.e., planner tailoring) competition. This
would, of course, imply a much smaller format for the event than the
format the IPC for automated planners has grown to already. But an
online hand-tailored competition would have the advantage of better
visibility of programming efforts, and of not taking up prohibitively
much time of the system developers.

In the context of competition size, last not least there are a few
words to be said about the role and responsibilities of the
organizers.  The IPC has grown too large to be handled, in all its
aspects, by just two persons. It may be worth thinking about
distributing the organization workload among a larger group of people.
One approach that might make sense would be to let different people
organize the tracks concerning the different PDDL subsets. Another
approach would be to let different people handle the language
definition, the benchmark preparation, and the results collection,
respectively. It would probably be a good idea to establish an IPC
council that, unlike the organizing committees in the past, would
persist across individual competitions, and whose role would be to
actively set up and support the organizing teams.

\acks{We would like to thank the IPC-4 organizing committee, namely
  Fahiem Bacchus, Drew McDermott, Maria Fox, Derek Long, Jussi
  Rintanen, David Smith, Sylvie Thiebaux, and Daniel Weld for their
  help in taking the decision about the language for the deterministic
  part of IPC-4, and in ironing out the details about syntax and
  semantics. We especially thank Maria Fox and Derek Long for giving
  us the latex sources of their PDDL2.1 article, and for discussing
  the modifications of this document needed to introduce the semantics
  of derived predicates and timed initial literals. We are indebted to
  Roman Englert, Frederico Liporace, Sylvie Thiebaux, and Sebastian
  Tr\"ug, who helped in the creation of the benchmark domains.  We
  wish to say a big thank you to all the participating teams for their
  efforts. There is significant bravery in the submission of a
  planning system to a competition, where the choice and design of the
  benchmark problems is up to the competition organizers, not to the
  individuals. We thank the LPG team for investing the extra effort of
  running the IPC-3 LPG version on the IPC-4 benchmarks, and we thank
  Shahid Jabbar for proofreading this text. We thank Subbarao
  Kambhampati for pointing out that the name ``classical part'' is
  ambiguous, and suggesting to use ``deterministic part'' instead.
  Last not least, we thank the anonymous reviewers, and Maria Fox in
  her role as the responsible JAIR editor, for their comments; they
  helped to improve the paper. At the time of organizing the
  competition, J\"org Hoffmann was supported by the DFG (Deutsche
  Forschungsgemeinschaft), in the research project ``HEUPLAN II''.
  Stefan Edelkamp was and is supported by the DFG in the research
  project ``Heuristic Search''.}

\begin{appendix}

\section{BNF Description of PDDL2.2}
\label{bnf}

\setcounter{tocdepth}{2}

This appendix contains a complete BNF specification of the PDDL2.2
language. For readability, we mark with $(***)$ the points in the BNF
where, in comparison to PDDL2.1, the new language constructs of
PDDL2.2 are inserted.

\subsection{Domains}

Domains are defined exactly as in PDDL2.2, except that we now also
allow to define rules for derived predicates at the points where
operators (actions) are allowed.

\begin{nopagebreak}\begin{tabtt}
\bump \= $\langle$domain$\rangle$ \bump\bump\=::= (def\=ine (domain $\langle$name$\rangle$)\+\+\+\\
                        {[}$\langle$require-def$\rangle$]  \\
                        {[}$\langle$types-def$\rangle$]\req{:typing} \\
                        {[}$\langle$constants-def$\rangle$] \\
                       {[}$\langle$predicates-def$\rangle$] \\
                        {[}$\langle$functions-def$\rangle$]\req{:fluents} \\
                        $\langle$structure-def$\rangle$\zom) \-\-\\
  $\langle$require-def$\rangle$  \> ::= (:requirements $\langle$require-key$\rangle$\oom) \\
  $\langle$require-key$\rangle$  \> ::= {\em See Section~\ref{sec-requirements}} \\
  $\langle$types-def$\rangle$    \> ::= (:types $\langle$typed list (name)$\rangle$) \\
  $\langle$constants-def$\rangle$ \>::= (:constants $\langle$typed list (name)$\rangle$) \\
  $\langle$predicates-def$\rangle$ \> ::= (:predicates $\langle$atomic formula skeleton$\rangle$\oom) \\
  $\langle$atomic formula skeleton$\rangle$ \\
                \> ::= ($\langle$predicate$\rangle$ $\langle$typed list (variable)$\rangle$) \\
  $\langle$predicate$\rangle$ \> ::= $\langle$name$\rangle$ \\
  $\langle$variable$\rangle$ \> ::= ?$\langle$name$\rangle$ \\
  $\langle$atomic function skeleton$\rangle$ \\
                \> ::= ($\langle$function-symbol$\rangle$ $\langle$typed list (variable)$\rangle$) \\
  $\langle$function-symbol$\rangle$ \> ::= $\langle$name$\rangle$ \\
  $\langle$functions-def$\rangle$ \> ::=\req{:fluents} (:fun\=ctions $\langle$function typed list \\
  \> \>                                 (atomic function skeleton)$\rangle$) \\
  $\langle$structure-def$\rangle$ \>::= $\langle$action-def$\rangle$\\
  $\langle$structure-def$\rangle$ \>::=\req{:durative-actions} $\langle$durative-action-def$\rangle$\\
\hspace{-1.2cm} $(***)$  $\langle$structure-def$\rangle$ \>::=\req{:derived-predicates} $\langle$derived-def$\rangle$\\
\= $\langle$typed list ($x$)$\rangle$ \=::= $x$\zom\+\\
  $\langle$typed list ($x$)$\rangle$ \>::=\req{:typing} $x$\oom - $\langle$type$\rangle$ $\langle$typed list($x$)$\rangle$\\
  $\langle$primitive-type$\rangle$\> ::= $\langle$name$\rangle$ \\
  $\langle$type$\rangle$ \> ::= (either $\langle$primitive-type$\rangle$\oom) \\
  $\langle$type$\rangle$ \> ::= $\langle$primitive-type$\rangle$ \\
\=  $\langle$function typed list ($x$)$\rangle$ \=::= $x$\zom\+\\
  $\langle$function typed list ($x$)$\rangle$ \>::=\req{:typing} $x$\oom - $\langle$\=function type$\rangle$ \\
  \> \> $\langle$function typed list($x$)$\rangle$\\
  $\langle$function type$\rangle$ \> ::= number \\
\end{tabtt}\end{nopagebreak}

\subsection{Actions}

The BNF for an action definition is the same as in PDDL2.2.

\begin{nopagebreak}\begin{tabtt}
\bump\=  $\langle$action-def$\rangle$ \bump \= ::= (:act\=ion $\langle$action-symbol$\rangle$\+\+\+\\
                        :parameters  ( $\langle$typed list (variable)$\rangle$ ) \\
                        $\langle$action-def body$\rangle$) \-\-\\
   $\langle$action-symbol$\rangle$ \> ::= $\langle$name$\rangle$ \\
   $\langle$action-def body$\rangle$ \>::= 
\=
                        {[}:precondition $\langle$GD$\rangle$] \+\+\\
                       {[}:effect       $\langle$effect$\rangle$]  \-\-\\
\= $\langle$GD$\rangle$ \hspace{3cm} \= ::= () \+\\
  $\langle$GD$\rangle$ \>::= $\langle$atomic formula(term)$\rangle$ \\
  $\langle$GD$\rangle$ \>::=\req{:negative-preconditions} $\langle$literal(term)$\rangle$\\
  $\langle$GD$\rangle$ \>::= (and $\langle$GD$\rangle$\zom) \\
  $\langle$GD$\rangle$ \> ::=\req{:disjunctive-preconditions} (or  $\langle$GD$\rangle$\zom) \\
  $\langle$GD$\rangle$\> ::=\req{:disjunctive-preconditions} (not $\langle$GD$\rangle$) \\
  $\langle$GD$\rangle$\> ::=\req{:disjunctive-preconditions} (imply $\langle$GD$\rangle$ $\langle$GD$\rangle$) \\
\smallskip
  $\langle$GD$\rangle$\> ::=\=\req{:existential-preconditions}  \+\+\\
             (exists ($\langle$typed list(variable)$\rangle$\zom) $\langle$GD$\rangle$ )\-\-\\ 
\smallskip
  $\langle$GD$\rangle$\> ::=\=\req{:universal-preconditions} \+\+\\
            (forall ($\langle$typed list(variable)$\rangle$\zom) $\langle$GD$\rangle$ ) \-\-\\ 
  $\langle$GD$\rangle$\> ::=\=\req{:fluents} $\langle$f-comp$\rangle$ \\
  $\langle$f-comp$\rangle$\> ::= ($\langle$binary-comp$\rangle$ $\langle$f-exp$\rangle$ $\langle$f-exp$\rangle$) \\
  $\langle$literal($t$)$\rangle$ \>::= $\langle$atomic formula($t$)$\rangle$ \\
  $\langle$literal($t$)$\rangle$ \>::= (not $\langle$atomic formula($t$)$\rangle$) \\
  $\langle$atomic formula($t$)$\rangle$ \>::= ($\langle$predicate$\rangle$ $t$\zom) \\
  $\langle$term$\rangle$ \> ::= $\langle$name$\rangle$ \\
  $\langle$term$\rangle$ \> ::= $\langle$variable$\rangle$ \\
  $\langle$f-exp$\rangle$ \> ::= $\langle$number$\rangle$ \\
  $\langle$f-exp$\rangle$ \> ::= ($\langle$binary-op$\rangle$ $\langle$f-exp$\rangle$ $\langle$f-exp$\rangle$) \\
  $\langle$f-exp$\rangle$ \> ::= (- $\langle$f-exp$\rangle$)\\
  $\langle$f-exp$\rangle$ \> ::= $\langle$f-head$\rangle$ \\
  $\langle$f-head$\rangle$ \> ::= ($\langle$function-symbol$\rangle$ $\langle$term$\rangle$\zom) \\
  $\langle$f-head$\rangle$ \> ::= $\langle$function-symbol$\rangle$ \\
  $\langle$binary-op$\rangle$ \> ::= $+$ \\
  $\langle$binary-op$\rangle$ \> ::= $-$ \\
  $\langle$binary-op$\rangle$ \> ::= $*$ \\
  $\langle$binary-op$\rangle$ \> ::= $/$ \\
  $\langle$binary-comp$\rangle$ \> ::= $>$ \\
  $\langle$binary-comp$\rangle$ \> ::= $<$ \\
  $\langle$binary-comp$\rangle$ \> ::= $=$ \\
  $\langle$binary-comp$\rangle$ \> ::= $>=$ \\
  $\langle$binary-comp$\rangle$ \> ::= $<=$ \\
  $\langle$number$\rangle$ \> ::= {\em Any numeric literal} \\
 \> $\quad$ {\em (integers and floats of form $n.n$).}\\
\=   $\langle$effect$\rangle$ \bump\= ::= () \+\\
  $\langle$effect$\rangle$ \> ::= (and $\langle$c-effect$\rangle$\zom)\\
  $\langle$effect$\rangle$ \> ::= $\langle$c-effect$\rangle$ \\
  $\langle$c-effect$\rangle$ \> ::=\req{:conditional-effects} (forall ($\langle$variable$\rangle$\zom) $\langle$effect$\rangle$) \\
  $\langle$c-effect$\rangle$ \>::=\req{:conditional-effects} (when $\langle$GD$\rangle$ $\langle$cond-effect$\rangle$) \\
  $\langle$c-effect$\rangle$ \>::= $\langle$p-effect$\rangle$ \\
  $\langle$p-effect$\rangle$ \> ::= ($\langle$assign-op$\rangle$ $\langle$f-head$\rangle$ $\langle$f-exp$\rangle$) \\
  $\langle$p-effect$\rangle$ \> ::= (not $\langle$atomic formula(term)$\rangle$) \\
  $\langle$p-effect$\rangle$ \>::= $\langle$atomic formula(term)$\rangle$  \\
  $\langle$p-effect$\rangle$ \>::=\req{:fluents}($\langle$assign-op$\rangle$ $\langle$f-head$\rangle$ $\langle$f-exp$\rangle$) \\  
  $\langle$cond-effect$\rangle$ \>::= (and $\langle$p-effect$\rangle$\zom)\\
  $\langle$cond-effect$\rangle$ \>::= $\langle$p-effect$\rangle$\\
  $\langle$assign-op$\rangle$ \>::= assign \\
  $\langle$assign-op$\rangle$ \>::= scale-up \\
  $\langle$assign-op$\rangle$ \>::= scale-down \\
  $\langle$assign-op$\rangle$ \>::= increase \\
  $\langle$assign-op$\rangle$ \>::= decrease \\
\end{tabtt}\end{nopagebreak}

\subsection{Durative Actions}

Durative actions are the same as in PDDL2.2, except that we restrict
ourselves to level 3 actions, where the duration is given as the fixed
value of a numeric expression (rather than as the possible values
defined by a set of constraints). This slightly simplifies the BNF.

\begin{nopagebreak}\begin{tabtt}
\bump\=  $\langle$durative-action-def$\rangle$ \= ::= (:dur\=ative-action $\langle$da-symbol$\rangle$\+\+\+\\
                        :parameters  ( $\langle$typed list (variable)$\rangle$ ) \\
                        $\langle$da-def body$\rangle$) \-\-\\
   $\langle$da-symbol$\rangle$ \> ::= $\langle$name$\rangle$ \\
   $\langle$da-def body$\rangle$ \>::= 
\=
                                :duration (= ?duration $\langle$f-exp$\rangle$) \+\+\\
                        :condition $\langle$da-GD$\rangle$\\
                        :effect $\langle$da-effect$\rangle$\-\-\\
\= $\langle$da-GD$\rangle$ \hspace{3cm} \= ::= () \+\\
  $\langle$da-GD$\rangle$ \>::= $\langle$timed-GD$\rangle$ \\
  $\langle$da-GD$\rangle$ \>::= (and $\langle$timed-GD$\rangle$\oom) \\
  $\langle$timed-GD$\rangle$ \>::= (at $\langle$time-specifier$\rangle$ $\langle$GD$\rangle$)\\
  $\langle$timed-GD$\rangle$ \>::= (over $\langle$interval$\rangle$ $\langle$GD$\rangle$)\\
  $\langle$time-specifier$\rangle$ \>::= start\\
  $\langle$time-specifier$\rangle$ \>::= end\\
  $\langle$interval$\rangle$ \>::= all\\
\end{tabtt}\end{nopagebreak}

\subsection{Derived predicates}

As said, rules for derived predicates can be given in the domain
description at the points where actions are allowed. The BNF is:

\begin{nopagebreak}\begin{tabtt}
\bump\= \hspace{-1.0cm} $(***)$  $\langle$derived-def$\rangle$ \bump \= ::= (:der\=ived $\langle$typed list (variable)$\rangle$
$\langle$GD$\rangle$)\\ 
\end{tabtt}\end{nopagebreak}

Note that we allow the specification of types with the derived
predicate arguments. This might seem redundant as the predicate types
are already declared in the {\tt :predicates} field. Allowing to
specify types with the predicate (rule) ``parameters'' serves to give
the language a more unified look-and-feel, and one might use the
option to make the parameter ranges more restrictive. (Remember that
the specification of types is optional, not mandatory.)

Repeating what has been said in Section~\ref{pddl:derived:syntax},
this BNF is more generous than what is considered a well-formed domain
description in PDDL2.2. We call a predicate $P$ {\em derived} if there
is a rule that has a predicate $P$ in its head; otherwise we call $P$
{\em basic}. The restrictions we apply are:

\begin{enumerate}
\item The actions available to the planner do not affect the derived
  predicates: no derived predicate occurs on any of the effect lists
  of the domain actions.
\item If a rule defines that $P(\overline{x})$ can be derived from
  $\phi(\overline{x})$, then the variables in $\overline{x}$ are
  pairwise different (and, as the notation suggests, the free
  variables of $\phi(\overline{x})$ are exactly the variables in
  $\overline{x}$).
\item If a rule defines that $P(\overline{x})$ can be derived from
  $\phi$, then the Negation Normal Form (NNF) of $\phi(\overline{x})$
  does not contain any derived predicates in negated form.
\end{enumerate}

\subsection{Problems}

The only change made to PDDL2.1 in the problem description is that we
allow the specification of timed initial literals.

\begin{nopagebreak}\begin{tabtt}
\bump\=$\langle$problem$\rangle$ \bump\bump \=::= (def\=ine (problem $\langle$name$\rangle$) \+\+\+\\
                                      (:domain $\langle$name$\rangle$) \\
                                     {[}$\langle$require-def$\rangle$] \\
                                      {[}$\langle$object declaration$\rangle$ ] \\
                                                          $\langle$init$\rangle$ \\
                                      $\langle$goal$\rangle$ \\
                                      {[}$\langle$metric-spec$\rangle$]\\
 % I think that length is a mistake, but it is used, so maybe we should include it?
                                      {[}$\langle$length-spec$\rangle$ ])\-\-\\
   $\langle$object declaration$\rangle$ ::= (:objects $\langle$typed list (name)$\rangle$) \\
   $\langle$init$\rangle$ \> ::= (:init $\langle$init-el$\rangle$\zom) \\
   $\langle$init-el$\rangle$ \> ::= $\langle$literal(name)$\rangle$\\
   $\langle$init-el$\rangle$ \> ::=\req{:fluents} (= $\langle$f-head$\rangle$ $\langle$number$\rangle$)\\
 \hspace{-1.2cm} $(***)$   $\langle$init-el$\rangle$ \> ::=\req{:timed-initial-literals} (at $\langle$number$\rangle$ $\langle$literal(name)$\rangle$)\\
   $\langle$goal$\rangle$ \> ::= (:goal $\langle$GD$\rangle$) \\
   $\langle$metric-spec$\rangle$ \> ::= (:metric $\langle$optimization$\rangle$ $\langle$ground-f-exp$\rangle$)\\
   $\langle$optimization$\rangle$ \> ::= minimize\\
   $\langle$optimization$\rangle$ \> ::= maximize\\
   $\langle$ground-f-exp$\rangle$ \> ::= ($\langle$binary-op$\rangle$ $\langle$ground-f-exp$\rangle$ $\langle$ground-f-exp$\rangle$)\\
   $\langle$ground-f-exp$\rangle$ \> ::= (- $\langle$ground-f-exp$\rangle$)\\
   $\langle$ground-f-exp$\rangle$ \> ::= $\langle$number$\rangle$\\
   $\langle$ground-f-exp$\rangle$ \> ::= ($\langle$function-symbol$\rangle$ $\langle$name$\rangle$\zom)\\
   $\langle$ground-f-exp$\rangle$ \> ::= total-time\\
   $\langle$ground-f-exp$\rangle$ \> ::= $\langle$function-symbol$\rangle$\\
\end{tabtt}
\end{nopagebreak}

Repeating what has been said in Section~\ref{pddl:derived:syntax}, the
requirement flag for timed initial literals implies the requirement
flag for durational actions (see also Section~\ref{sec-requirements}),
i.e. the language construct is only available in PDDL2.2 level 3.
Also, the above BNF is more generous than what is considered a
well-formed problem description in PDDL2.2. The times {\tt
  $\langle$number$\rangle$} at which the timed literals occur are
restricted to be greater than $0$. If there are also derived
predicates in the domain, then the timed literals are restricted to
not influence any of these, i.e., like action effects they are only
allowed to affect the truth values of the basic (non-derived)
predicates (IPC-4 will not use both derived predicates and timed
initial literals within the same domain).

\subsection{Requirements}
\label{sec-requirements}

\begin{sloppypar}
  Here is a table of all requirements in PDDL2.2. Some requirements
  imply others; some are abbreviations for common sets of
  requirements.  If a domain stipulates no requirements, it is assumed
  to declare a requirement for {\tt :strips}.
\begin{center}
{\small \begin{tabular}{ll}
\it Requirement & \it Description \\
\tt :strips & Basic STRIPS-style adds and deletes \\
\tt :typing & Allow type names in declarations of variables \\
\tt :negative-preconditions & Allow {\tt not} in goal descriptions\\
\tt :disjunctive-preconditions & Allow {\tt or} in goal descriptions \\
\tt :equality & Support {\tt =} as built-in predicate \\
\tt :existential-preconditions & Allow  {\tt exists} in
goal descriptions \\
\tt :universal-preconditions & Allow {\tt forall}  in
goal descriptions \\
\tt :quantified-preconditions & = {\tt :existential-preconditions}  \\
  & + {\tt :universal-preconditions} \\
\tt :conditional-effects & Allow {\tt when} in action effects \\
\tt :fluents & Allow function definitions and use of effects using\\
 & assignment operators and arithmetic preconditions.\\
\tt :adl & = {\tt :strips} + {\tt :typing}\\
   &      + {\tt :negative-preconditions}\\
   &      + {\tt :disjunctive-preconditions} \\
   &      + {\tt :equality}  \\
   &      + {\tt :quantified-preconditions}  \\
&+ {\tt :conditional-effects} \\
\tt :durative-actions & Allows durative actions. \\
& Note that this does not imply {\tt :fluents}.\\
\tt :derived-predicates & Allows predicates whose truth value is\\
& defined by a formula\\
\tt :timed-initial-literals & Allows the initial state to specify
literals\\
& that will become true at a specified time point\\
& implies {\tt durative actions} (i.e. applicable only\\
& in PDDL2.2 level 3)
\end{tabular}}
\end{center}
\end{sloppypar}

%%% Local Variables: 
%%% mode: latex
%%% TeX-master: t
%%% End: 

%%% Local Variables: 
%%% mode: latex
%%% TeX-master: t
%%% End: 

\end{appendix}

\bibliographystyle{theapa}
\bibliography{./abbreviations,./biblio,./crossref}

\end{document}